\documentclass{article}
\usepackage{iclr2027_conference,times}
\usepackage{hyperref}

\usepackage{amsmath,amsfonts,bm}

\def\eqref#1{Eq.~\ref{#1}}

\def\1{\bm{1}}

\def\vzero{{\bm{0}}}

\def\vtheta{{\bm{\theta}}}

\def\vb{{\bm{b}}}

\def\vd{{\bm{d}}}

\def\vg{{\bm{g}}}
\def\vh{{\bm{h}}}

\def\vm{{\bm{m}}}

\def\vr{{\bm{r}}}

\def\vu{{\bm{u}}}
\def\vv{{\bm{v}}}

\def\vx{{\bm{x}}}
\def\vy{{\bm{y}}}

\def\mF{{\bm{F}}}

\def\mH{{\bm{H}}}
\def\mI{{\bm{I}}}
\def\mJ{{\bm{J}}}

\def\mV{{\bm{V}}}
\def\mW{{\bm{W}}}
\def\mX{{\bm{X}}}
\def\mY{{\bm{Y}}}
\def\mZ{{\bm{Z}}}

\DeclareMathAlphabet{\mathsfit}{\encodingdefault}{\sfdefault}{m}{sl}
\SetMathAlphabet{\mathsfit}{bold}{\encodingdefault}{\sfdefault}{bx}{n}

\newcommand{\E}{\mathbb{E}}

\newcommand{\R}{\mathbb{R}}

\usepackage{hyperref}
\usepackage{url}
\usepackage{algorithm}
\usepackage{algorithmic}

\usepackage{graphicx}
\usepackage{wrapfig}
\newcommand{\needlines}[1]{\par\begingroup\dimen0=#1\baselineskip\relax
  \ifdim\pagegoal=\maxdimen\else\ifdim\dimexpr\pagegoal-\pagetotal\relax<\dimen0 \newpage\fi\fi\endgroup}
\usepackage{xcolor}
\definecolor{citecol}{RGB}{20,50,120}
\definecolor{refcol}{RGB}{191,87,0}
\definecolor{stepcol}{RGB}{30,125,105}
\newcommand{\hncstep}[2]{\textcolor{stepcol}{(#1)~#2}}
\hypersetup{colorlinks=true, citecolor=citecol, linkcolor=refcol, urlcolor=citecol}

\newcommand{\GN}{\mathbf{G}}
\newcommand{\Vzero}{\mathcal{V}_0}

\makeatletter
\def\section{\@startsection{section}{1}{\z@}%
  {-0.8mm plus -0.3mm minus -0.2mm}{0.4mm plus 0.15mm minus 0.1mm}%
  {\large\sc\raggedright}}
\def\subsection{\@startsection{subsection}{2}{\z@}%
  {-0.6mm plus -0.3mm minus -0.2mm}{0.3mm plus 0.1mm}%
  {\normalsize\sc\raggedright}}
\def\subsubsection{\@startsection{subsubsection}{3}{\z@}%
  {-0.5mm plus -0.3mm minus -0.2mm}{0.2mm plus 0.1mm}%
  {\normalsize\sc\raggedright}}
\def\paragraph{\@startsection{paragraph}{4}{\z@}%
  {0.6mm plus 0.3mm minus 0.2mm}{-0.5em}{\normalsize\bf}}
\makeatother

\title{Traversing the solution space of neural networks with Hessian Null Space Continuation}

\author{%
Ann Huang$^{1,2,3}$ \quad Mitchell Ostrow$^{4}$ \quad Zhouyang Lu$^{5}$ \\
\textbf{William T. Redman$^{6*}$ \quad Leo Kozachkov$^{5*}$ \quad Kanaka Rajan$^{2,3*}$} \\
\textnormal{$^{1}$Harvard University \quad $^{2}$Harvard Medical School \quad $^{3}$Kempner Institute} \\
\textnormal{$^{4}$Massachusetts Institute of Technology\quad $^{5}$Brown University \quad $^{6}$Johns Hopkins University} \\
\textnormal{$^{*}$Co-senior authors} \\
\texttt{\hspace{0.5pt}annhuang@g.harvard.edu}}

\iclrfinalcopy

\begin{document}

\maketitle
\lhead{Preprint}
\vspace{-3.8mm}
\begin{abstract}
\vspace{-2mm}

On a single task, deep neural networks can learn a wide range of solutions, depending on their optimizer, training data, architecture, and hyperparameters. Many of these solutions are surprisingly \textit{mode-connected}: rather than isolated points in the weight space, they are connected by low loss regions. Despite this observation, the diversity of solutions in terms of their internal computation in these regions has not been characterized. A parallel line of work has identified the \textit{degeneracy of neural representations}: many neural network solutions exist with similar training loss yet distinct internal structures. However, it is unclear how these diverse solutions are related in weight space. Here, we unify these subfields and demonstrate for the first time that there exist many different internal mechanisms within a local mode connected region in weight space. To do so, we introduce Hessian Null Space Continuation (HNC), a scalable method that uses local curvature information to traverse regions of weight space that preserve network function. 
HNC can additionally be steered toward solutions with specified properties. 
In RNNs trained on a memory task, HNC drives the networks to learn drastically different representations and dynamics, even with maintained behavior. In ImageNet-trained Vision Transformers, HNC finds alternative representations which differ more from the original network than any independently trained models of different architectures and training objectives.
In reinforcement-learning agents, HNC uncovers a distinct behavioral strategy at comparable return in a navigation task, and exposes a reward-hacking strategy in an AI Safety Gridworld environment. Finally, HNC provides local geometric information about the solution distribution, showing how model size and task complexity shape its dimension and functional sensitivity. Together, our results show that a surprisingly large amount of representational diversity exists near a single trained solution, which is unseen by standard gradient-based optimization techniques. Our domain-agnostic method, HNC, can identify and quantify this diversity, opening new possibilities for mechanistic understanding of solution spaces and providing a principled basis for model merging, editing, and fine-tuning. Project page and code available at\\
\href{https://ann-huang-0.github.io/Hessian-null-space-continuation/}{\textcolor{citecol}{\nolinkurl{ann-huang-0.github.io/Hessian-null-space-continuation}}}.

\end{abstract}

\vspace{-2mm}
\section{Introduction}
\vspace{-1mm}

Overparameterized neural networks can admit many local minima with similarly low
loss. A major line of work in machine learning therefore asks how the optimizer, training data,
architecture, and task shape which local minimum is reached by training. Investigation into this has shed
fundamental light on the implicit bias of stochastic gradient descent towards flatter
minima \citep{kleinberg2018alternative, feng2021inverse} and the converging representations
between networks as models get larger and tasks get more complex \citep{huh2024platonic}.
\vspace{-1mm}

A classic picture of the loss landscape depicts local minima as isolated basins that are well
separated in parameter space \citep{li2018visualizing}. This intuition, however, is grounded
in low dimensions and convex optimization, and becomes unreliable in the extremely high-dimensional, nonconvex landscapes of modern over-parameterized networks \citep{sagun2016eigenvalues, sagun2017empirical, gurari2018gradient}. Studies of local loss geometry have revealed many nearly flat directions around trained, low-loss solutions \citep{sagun2016eigenvalues, sagun2017empirical,chaudhari2017entropy, ghorbani2019investigation}. The mode connectivity literature has also shown
that independently trained networks are in fact often connected in weight space: nonlinear
paths of non-increasing loss provably exist between minima of fully connected neural networks
\citep{freeman2017, draxler2018essentially, garipov2018loss}, and some minima are even connected
by linear paths with negligible loss barriers (linear mode connectivity, \cite{frankle2020linear}). 
These results
suggest that apparently distinct minima may lie within a common connected region of low loss in weight space. However, this literature has largely characterized that region through
connectivity and loss alone, rather than the internal computations that networks within it implement. \citet{lubana2023} took an important step in this direction, showing
that linear mode connectivity between two networks can designate whether they share similar input attribution functions, and that nonlinear low-loss paths can connect even dissimilar minima. Yet these
analyses studied pairs of  networks identified via training, rather
than systematically examining the low-loss region surrounding a solution. It remains unknown whether the connected
low-loss region around a single trained solution contains genuinely distinct representations
and mechanisms, or only minor modifications of the same representation and mechanism.

Meanwhile, an emerging line of work suggests that a single task can admit a family of different
solutions \citep{goldman_memory_2009,huang2025measuring, damour_underspecication_nodate, turner_charting_nodate,
lappalainen_connectome-constrained_2024, kurtkaya_dynamical_2025, murray_phase_2025, clark2026structure, zhong2023clock, saxe2022neural,ostrow_metric_2026},
characterized by distinct out-of-distribution (OOD) generalization behavior \citep{huang2025measuring,
damour_underspecication_nodate,ostrow_metric_2026} and distinct geometric and dynamical properties of the internal
representations \citep{huang2025measuring, turner_charting_nodate, kurtkaya_dynamical_2025,
murray_phase_2025, clark2026structure,ostrow_metric_2026}. Such solution degeneracy is further supported by work showing that
function and representations can doubly dissociate \citep{braun2025solutions}, even in
nonlinear neural networks \citep{theiss_parameter_2026, theiss_representation_2026}. Beyond
standard solutions reached by gradient descent during training, recent work has deliberately sought to find alternative solutions to the same task by training with regularization
that either penalizes similarity to a standard solution \citep{qian2026discovering,
ostrow_metric_2026} or makes the desired representational similarity score a hard constraint
\citep{braun2025solutions, theiss_representation_2026}, or by following eigenvectors of
the Hessian from a saddle point toward distinct minima \citep{parkerholder2020ridge}. Relatedly, \citet{gan2026thickets}
showed that the local neighborhood of a single pretrained model can contain diverse
experts with improved performance on specific tasks, suggesting that meaningfully different
solutions may be accessible nearby in weight space. However, the relationship
among these alternative solutions in parameter space remains poorly understood.
In particular, \textbf{how much representational diversity exists within the connected,
function-preserving region surrounding a single trained solution,
and how can that region be traversed constructively?}

\begin{wrapfigure}{r}{0.6\textwidth}
  \centering
  \vspace{-6mm}
  \includegraphics[width=\linewidth]{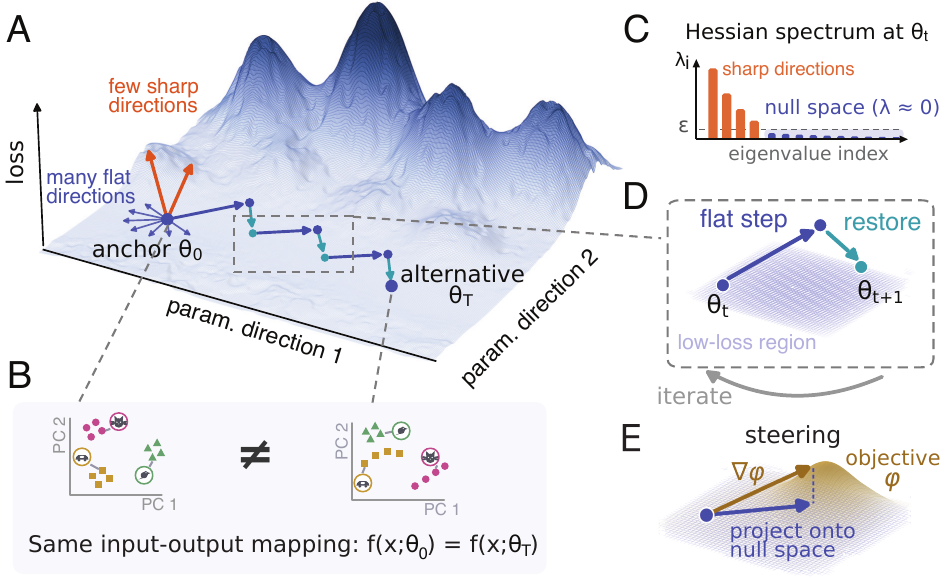}
  \vspace{-7mm}
  \caption{\small \textbf{Hessian Null Space Continuation (HNC) overview.} 
\textbf{(A)} Around a trained network $\vtheta_0$ (the anchor), the function-matching loss
\eqref{eq:funcloss} is sharp along a few directions and flat along many. HNC
follows the flat directions to an alternative network $\vtheta_T$ within the connected low-loss region.
\textbf{(B)} The two networks share the same input-output mapping, but their hidden-state representations of the same inputs can differ.
\textbf{(C)} The Hessian spectrum at $\vtheta_t$ separates a few sharp directions
from a high-dimensional null space with eigenvalues below a threshold $\epsilon$.
\textbf{(D)} Each iteration takes a step within the null space (flat step), then
re-minimizes the function-matching loss (restore), yielding $\vtheta_{t+1}$.
\textbf{(E)} The flat step can be steered to optimize a differentiable objective $\varphi$ by following its gradient projected onto the null space.
 }
  \label{fig:overview}
  \vspace{-2mm}
\end{wrapfigure}

The set of solutions a given neural network model learns is constrained not only by the task constraints \citep{cao2024explanatory,yamins2026contravariance}, but also the biases induced by optimizers. For example, stochastic gradient descent is known to induce low-rank, low-norm solutions \citep{soudry2018implicit,gunasekar2017implicit,arora2019implicit,woodworth2020kernel,huh2023lowrank,vardi2023implicit}. Characterizing the broader set of solutions that a model \textit{admits}
enables dissociation from optimizer constraints and therefore unbiased analysis of shared structure across all solutions via task demands. 
On the practical side, if representationally distinct solutions are reachable
from a learned solution through local, function-preserving weight updates, this exposes degrees
of freedom that could be exploited for fine-tuning \citep{aghajanyan_intrinsic_2021, hu_lora_2021,
gan2026thickets, qiu_evolution_2025, liang2026blessing}, model merging \citep{wortsman_model_2022,
ilharco_editing_2022}, and post-hoc model editing \citep{meng_locating_2022, mitchell_fast_2022}.

To address these questions, we developed Hessian Null Space Continuation (HNC), a generic, model-agnostic method for identifying diverse solutions in neural networks. Starting
from one base trained network (which we call the anchor), HNC  explores a connected set of
function-preserving alternatives using local curvature information. It can additionally be steered by
the gradient of any differentiable function to find solutions with specific properties.
In RNNs trained on a memory task, HNC transforms the canonical fixed-point
solution of \cite{sussillo2013opening} into alternatives with qualitatively different geometry and
dynamics. In ImageNet-trained Vision Transformers (ViTs),  HNC finds nearby representations which differ more from the original network than any independently trained models of different architectures and training objectives, and even untrained models. This reveals substantial representational freedom inside
neural networks underexplored by gradient training. We further extend HNC to reinforcement learning (RL), where it identifies policies that achieve comparable return through distinct behavioral strategies, including reward-hacking solutions.
Together, our results show that an unexpectedly large amount of representational diversity (both in quantity and quality) exists near a single trained solution, which is unseen by standard optimization techniques. Our domain general method, HNC, can identify and quantify this diversity.   

\section{Method}
\label{sec:method}

\subsection{Loss landscape curvature reveals function-preserving directions}

Starting from a trained network $\vtheta_0$, the anchor, we seek
networks with different internal representations but matched
behavior. We operationalize this through the function-matching loss
\vspace{-1mm}
\begin{equation}
\mathcal{L}(\vtheta)
=
\tfrac12
\E_{\vx\sim\mathcal{D}}
\left[
\left\lVert
f_{\vtheta}(\vx)-f_{\vtheta_0}(\vx)
\right\rVert^2
\right],
\label{eq:funcloss}
\end{equation}
which measures output drift from the anchor over an input distribution
$\mathcal{D}$. In practice, we estimate this expectation using a fixed
set $\mX$ sampled from $\mathcal{D}$, which we call the probe set. 
$\mathcal{L}$ is nonnegative and $\mathcal{L}(\vtheta_0)=0$, so the anchor is a global minimizer and $\nabla\mathcal{L}(\vtheta_0)=\vzero$.
A small weight perturbation $\bm{\delta}$ thus changes the loss only at second order:
\begin{equation}
\mathcal{L}(\vtheta_0+\bm{\delta})
=
\tfrac12
\bm{\delta}^{\top}\mH\bm{\delta}
+
O(\lVert\bm{\delta}\rVert^3),
\qquad
\mH
=
\nabla_{\vtheta}^2\mathcal{L}(\vtheta_0).
\label{eq:quadloss}
\end{equation}
The eigenvalues of $\mH$ therefore measure the network's local sensitivity to parameter change along different parameter directions. 
Large eigenvalues identify sharp directions along which the network outputs change rapidly, whereas near-zero eigenvalues identify flat directions with little output drift. We define their span,
\(
\Vzero
=
\mathrm{span}\bigl\{
\vv_i:\lambda_i\leq\epsilon
\bigr\} \), 
as the approximate null space. In trained networks, the Hessian typically has a few large eigenvalues and many near-zero eigenvalues, so this approximate null space often spans a substantial fraction of the parameter dimensions \citep{sagun2017empirical, gurari2018gradient, liang2026blessing}.
Locally, $\Vzero$ consists of directions
along which HNC can move the weights while approximately preserving the
network function, allowing it to search for solutions with different internal representations. We discuss our choice of $\epsilon$ later in Section \ref{sec:nullspace}.

\subsection{Hessian Null Space Continuation}

A single step within $\Vzero$ explores only the immediate neighborhood of the anchor.
Exploring the function-preserving parameter set more extensively therefore requires
a multi-step procedure. 
However, because the Hessian provides only a second-order
approximation to the function-matching loss in \eqref{eq:funcloss}, even a step along an
eigenvector with zero eigenvalue can change the loss through higher-order terms neglected
by \eqref{eq:quadloss}. Moreover, the Hessian is a local approximation and its null space
may cease to align with the flat directions away from $\theta_0$.

HNC addresses both points by alternating between a null space step and a
function restoration step (Fig. \ref{fig:overview}). At each iteration $t$, we \hncstep{1}{take a small step
within the current null space $\Vzero$}, and then \hncstep{2}{restore function matching}
by taking a few gradient-descent steps on the loss in \eqref{eq:funcloss}:
\begin{equation}
\tilde{\vtheta}_{t} \;=\; \vtheta_t + \eta\,\vd_t, \qquad
\vtheta_{t+1} \;=\; \tilde{\vtheta}_{t} - \alpha\,\nabla\mathcal{L}\bigl(\tilde{\vtheta}_{t}\bigr),
\label{eq:predcorr}
\end{equation}
where $\vd_t\in\Vzero$ is the step direction sampled uniformly from $\Vzero$ , and $\eta$, $\alpha$ are the step sizes. For
clarity, \eqref{eq:predcorr} shows a single restoration gradient step, and in practice we
apply $m$ such steps. The restoration step corrects the higher-order drift introduced by the
preceding null space step. When the loss after a function restoration step exceeds a threshold,
$\mathcal{L}(\vtheta_{t+1})>\tau$, this indicates that our linear approximation by the Hessian has become a poor fit of the local loss landscape. In these instances, we \hncstep{3}{relinearize} by recomputing the
Hessian null space at the current weights. We repeat this procedure until reaching either a prespecified number of steps or an
alternative solution with the desired properties. In this way, HNC can move far from the anchor,
while keeping $\mathcal{L}$ close to zero throughout. At a high level, HNC shares conceptual
similarity with the classical predict-correct structure of the \emph{numerical continuation}
method, which traces a curve by taking a small step along its local tangent and
then correcting back onto the curve \citep{keller1977numerical,allgower2003introduction}.

\vspace{-1mm}
\subsubsection{Forming the null space}
\label{sec:nullspace}
Because the function-matching loss has its exact minimum at $\vtheta_0$, its Hessian there is
positive semidefinite, with all eigenvalues nonnegative (Appendix~\ref{app:gn}). The null space
$\Vzero$ is therefore the span of the eigenvectors with the numerically smallest eigenvalues.
For a network with $P$ parameters, $\mH$ is a $P\times P$ matrix and is too large to store or
eigendecompose directly. 
We compute the $k$ smallest eigenpairs using the locally optimal
block preconditioned conjugate gradient method
(LOBPCG; Appendix~\ref{app:lobpcg}; \citealp{knyazev2001toward}), which uses Hessian-vector
products computed through automatic differentiation without
explicitly forming the Hessian \citep{pearlmutter1994fast}.
Time and memory then scale as $\mathcal{O}(kP)$ rather than $\mathcal{O}(P^3)$ and $\mathcal{O}(P^2)$.
A direction counts as flat when $\lambda_i \le \mu_{\mathrm{rel}}\,\lambda_{1}$, where $\lambda_{1}$ is the largest Hessian eigenvalue at the current linearization point and $\mu_{\mathrm{rel}}\in[10^{-7},10^{-3}]$ is a relative threshold chosen per experiment. Because the threshold is relative, the null space is re-computed at every linearization. Sweeps over $\mu_{\mathrm{rel}}$ are presented in Appendix~\ref{app:flat-threshold} and Appendix~\ref{app:ablation}.

\vspace{-1mm}
\subsubsection{Steering}
\vspace{-1mm}

HNC can be steered to optimize a differentiable objective $\varphi$ while preserving the network's outputs, through projecting its gradient $\nabla\varphi$ onto the Hessian null space. To retain $\nabla\varphi$ along flat parameter directions and suppress those along sharp directions, we use a \emph{soft projection}. Specifically, we rescale the gradient component along each Hessian eigenvector $\vv_i$ by $1/(1+\lambda_i/\mu)$, where $\lambda_i$ is its eigenvalue and $\mu>0$ is a damping parameter: components along flat directions ($\lambda_i\ll\mu$) pass through, while those along sharp directions ($\lambda_i\gg\mu$) are suppressed. Since the gradient component along the eigenvector $\vv_i$ is $(\vv_i^\top\nabla\varphi)\vv_i$, rescaling each component and summing over parameter directions therefore gives
\(
\vd=\sum_i \frac{\vv_i^\top\nabla\varphi}{1+\lambda_i/\mu}\,\vv_i
  =(\mI+\mH/\mu)^{-1}\nabla\varphi.
\)
This matrix form allows us to bypass the need to explicitly construct the eigenbasis of the Hessian.
We set the damping factor to the flatness threshold of Sec.~\ref{sec:nullspace}, $\mu=\mu_{\mathrm{rel}}\lambda_1$, so that a single hyperparameter $\mu_{\mathrm{rel}}$ defines both which directions count as flat and which gradient directions are retained through soft projection. 

\section{HNC drives RNNs to learn drastically different dynamics in a memory task}
\label{sec:rnn}

\begin{figure}[t]
\centering
\vspace{-7mm}
\includegraphics[width=\textwidth]{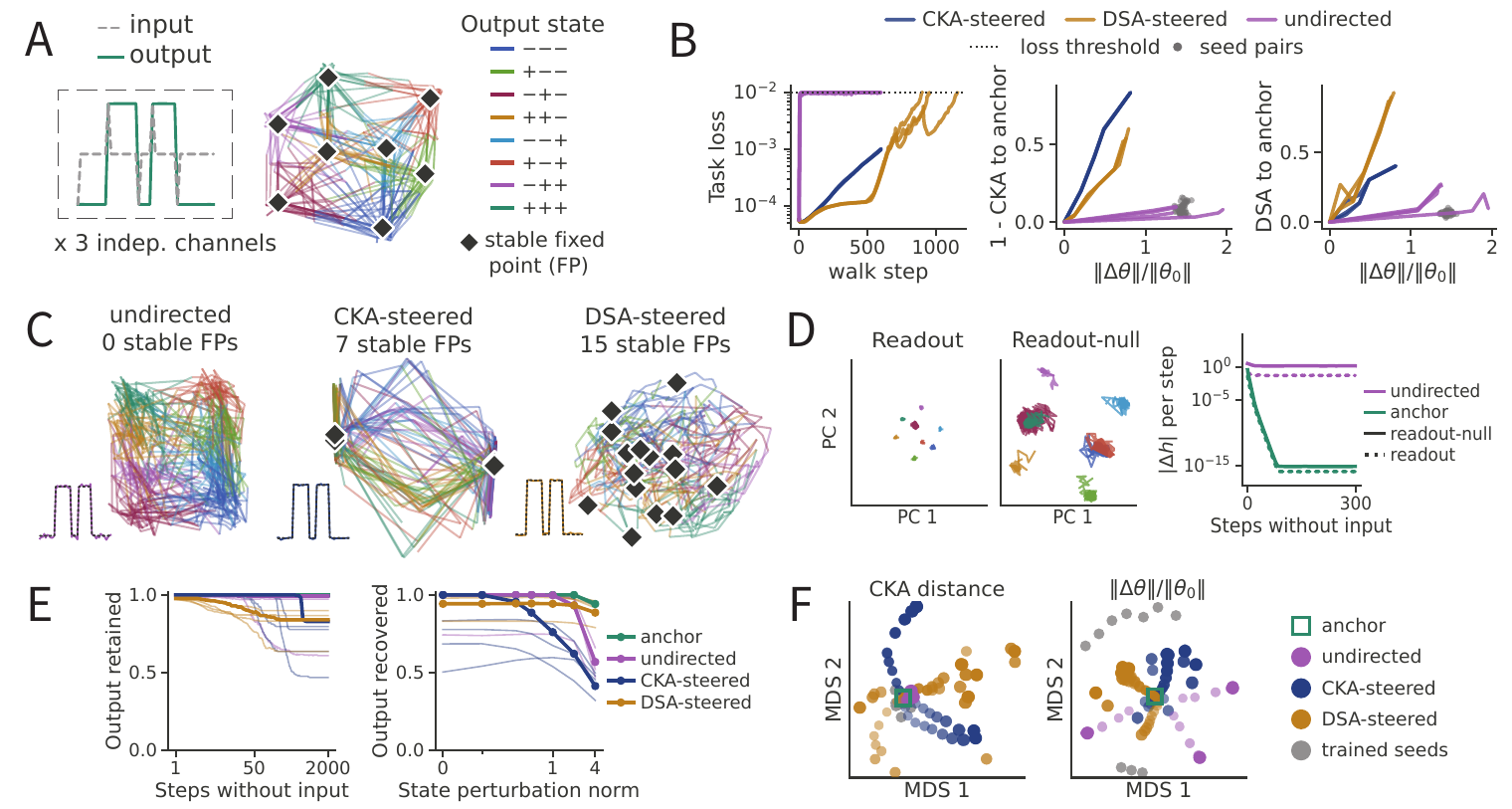}
\vspace{-6mm}
\caption{\small \textbf{HNC drives RNNs to learn drastically different dynamics in a memory task.}
\textbf{(A)} The 3-bit flip-flop task and the anchor's solution. Left: one input channel and the corresponding output, where the output holds the last nonzero input.
Right: hidden-state trajectories of the trained anchor projected onto their top three principal components,
colored by the output state, with the eight stable fixed points (black
diamonds) at the vertices of a cube.
\textbf{(B)} Undirected, CKA-steered, and DSA-steered HNC walks from the same anchor. Left: task loss plotted against the walk step. Middle and right: representational ($1 - \mathrm{CKA}$) and dynamical (DSA) distance to the anchor against relative weight movement. 
\textbf{(C)} Hidden-state trajectory projected onto their top three principles components, and output traces from networks reached during the HNC walks. 
\textbf{(D)} Left and middle: hidden states of the undirected endpoint during input-free memory periods, projected onto the top two PCs of the readout and readout-null subspaces, colored by output state as in (A). Right: per-step hidden-state movement in the readout (dotted) and readout-null (solid) subspaces. \textbf{(E)} Fraction of output retained over increasing steps without input (left) and recovered after hidden-state perturbations (right). Thick lines: networks shown in (C); thin lines: endpoints of other independent HNC walks and the other trained anchors. \textbf{(F)} MDS embeddings of 131 networks (the anchor, ten independently trained seeds, and all checkpoints from three HNC walks of each type), using pairwise representational distance ($1 - \mathrm{CKA}$, left) or relative weight distance (right). Marker size and opacity increase along each walk.
}
\label{fig:manifold}
\vspace{-6mm}
\end{figure}

We first applied HNC to small, interpretable RNNs trained on a memory task. RNN computations can be analyzed as a dynamical system: hidden-state trajectories and stable states reveal how networks store and update information \citep{sussillo2013opening, mante2013context}. We use the 3-Bit Flip-Flop (3BFF) task from computational neuroscience \citep{sussillo2013opening}, in which networks maintain three binary memories, each storing the last nonzero input on one input-output channel (Fig.~\ref{fig:manifold}A, left). We trained 64-unit \textit{tanh} RNNs using backpropagation through time (BPTT) and the Adam optimizer to minimize the mean-squared error between target and network outputs. After training, the networks represented the $2^3=8$ memory states as eight stable fixed points, which are hidden-states that remain unchanged when the input is turned off. These fixed points formed the vertices of a three-dimensional cube in activation space (Fig.~\ref{fig:manifold}A, right). The Hessian spectrum falls from a few sharp directions into a broad near-zero bulk, with a participation ratio of about 23 across the five anchors (Appendix~\ref{app:flat-threshold}, Fig.~\ref{fig:3bff_gn_spectrum}). 

Next, we aim to characterize both alternative 3BFF solutions accessible through undirected null space exploration, and those that differ maximally from the anchor. For the latter, we steer
toward maximizing one of two dissimilarity metrics: the CKA distance, defined as $1-\mathrm{CKA}$, which measures
how much the representational geometry (the kernel of the hidden activations) has changed
\citep{kornblith2019similarity}, or the DSA distance, which measures how much the recurrent dynamics
have changed up to invertible transformations \citep{ostrow2023beyond,ostrow_metric_2026}. We applied HNC to five trained networks (the ``anchors") with adaptive step size and periodic relinearization (see Appendix~\ref{app:walk-settings} for details). All HNC walks maintain task loss below $0.01$, while steered walks steadily increase representational and dynamical distances from their respective anchors (Fig.~\ref{fig:manifold}B). While undirected walks reach representational and dynamical distances comparable to those between independently trained networks, steered walks consistently achieve  greater divergence with less weight movement. During undirected exploration, the eight fixed points disappeared, replaced by a continuous manifold with eight distinct regions, one for each memory state (Fig. \ref{fig:manifold}C). Meanwhile, the CKA- and DSA-steered HNC walks altered the number of fixed points within the network and produced visually distinct state space structure as shown in Fig. \ref{fig:manifold}C. 

To understand how the alternative networks maintain memory, we examined the network reached during the undirected exploration. During input-free memory periods, its hidden-states form eight compact points in the readout subspace but continue moving within eight distinct clusters in the readout-null subspace (Fig.~\ref{fig:manifold}D, left and middle). Further quantifying the speed of hidden-state movement, the alternative solution's hidden-state continues to drift without input, mainly in the readout null space (Fig.~\ref{fig:manifold}D, right). Thus, this network maintains memory within distinct regions of state-space rather than at fixed points, preserving stable outputs despite ongoing internal dynamics. 
We further assessed the alternative solution's memory retention and robustness to perturbations, showing that the CKA-steered endpoint is less robust to hidden-state perturbations, whereas the DSA-steered endpoint shows memory drift during prolonged memory periods  (Fig. \ref{fig:manifold}E). This highlights that even networks with matching in-distribution input-output mappings can behave differently under extended or perturbed conditions. 
Finally, to visualize the structure of the solution set reached by HNC, we computed multidimensional scaling (MDS) embeddings of 131 networks, based on pairwise representational distance ($1-\mathrm{CKA}$) or relative weight distance (Fig. \ref{fig:manifold}F). These networks comprise the anchor, all checkpoints from three HNC walks of each type (undirected, CKA-steered, DSA-steered), and ten independently trained seeds. All networks solve the task with accuracy above $0.999$. In representation space, the independently trained and undirected-walk networks sit close to the anchor, while the steered walks extend far beyond them into different parts of the solution space. In weight space, by contrast, the steered walks stay closer to the anchor than both the undirected walks and the spread among independent seeds, demonstrating a clear decoupling of representational from weight-space distance. The solutions do not collapse onto a single tight cluster in representation space, highlighting the diversity of solutions accessible by HNC within the connected low-loss region around one trained network.

\vspace{-1mm}
\section{HNC reveals underexplored degrees of freedom in ViTs}
\vspace{-1mm}
\label{sec:prh}

The Platonic Representation Hypothesis (PRH) \citep{huh2024platonic} argues that representations
become increasingly similar across models as models scale and tasks get harder. \cite{huh2024platonic} attribute this to a convergence to the
shared underlying ``platonic'' representation of reality that models approach. Here, we ask how much representational freedom remains within a trained model when its function is held approximately fixed. 
To test this, we take a ViT-S/16 (22M parameters, \cite{dosovitskiy2021vit}) pretrained on ImageNet classification and search for maximally different representations at preserved input-output mapping. Concretely, we applied HNC to the pretrained ViT and steered to minimize the max-over-layers CKA similarity between the alternative representation and the anchor. In Appendix~\ref{app:prh_knn}, steering against the max-over-layers mutual k-nearest-neighbor (kNN) score, the metric used in \cite{huh2024platonic}, gives the same qualitative results.

\begin{figure}[H]
\centering
\vspace{-4mm}
\includegraphics[width=\textwidth]{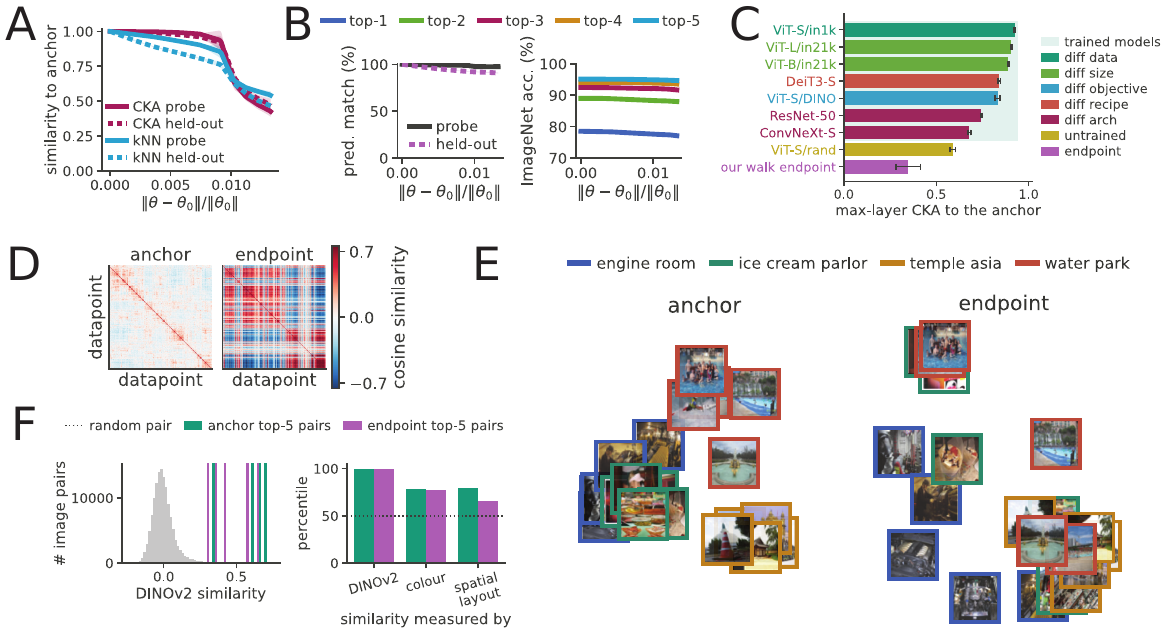}
\vspace{-8mm}
\caption{\small \textbf{HNC reveals representational freedom underexplored by gradient descent in ViTs.}
\textbf{(A)} Max-over-layers CKA and $k$-nearest-neighbor (kNN) to the anchor along the HNC walk, both on images it was steered on (the probe set) and on images never seen during HNC (the held-out set).
\textbf{(B)} Left: agreement between the anchor and HNC endpoint's predictions on Places-365 images. Right: top-1 to top-5 ImageNet accuracy. 
\textbf{(C)} Max-over-layers CKA to the anchor, between our HNC endpoint, an untrained ViT, and seven trained models. In (A-C), lines show the mean and shaded bands indicate $\pm 1$ s.d.\ across five random probe sets.
\textbf{(D)} Pairwise cosine similarities between penultimate-layer representations of probe set images, for the anchor and one HNC endpoint.
\textbf{(E)} Multidimensional scaling (MDS) embedding of the similarities in (D) for 20 images from 4 classes. 
\textbf{(F)} For both the anchor and HNC endpoint, we take the five image pairs represented as most similar, and score each pair's similarity on three attributes: semantic content (DINOv2 feature similarity), color, and spatial layout. Scores are percentiles among all pairs of probe images, where higher means the network's closest pairs share that attribute more than chance. Dashed line is the level of a random pair. 
}
\label{fig:prh}
\vspace{-5mm}
\end{figure}

Starting from the pretrained ViT, we evaluated the anchor on 512 images from the Places-365 validation set \citep{zhou2017places}, the dataset on which the PRH was evaluated \citep{huh2024platonic}. The function-matching loss in \eqref{eq:funcloss} is evaluated over the anchor and alternative network's logits on these same images. We record the CLS-token features after every transformer block together with the final pre-logits features, and steer the walk to reduce the highest CKA between any layer of the alternative network and any layer of the anchor. Because the hard maximum is not differentiable, we steer using a smooth softmax CKA surrogate while reporting the hard maximum CKA throughout (Appendix \ref{app:softmaxcka}). During HNC, the representational similarity between the anchor and the alternative network falls steadily, both on the images the walk was steered on (the probe set) and on images never seen during HNC (the held-out set) (Fig.~\ref{fig:prh}A). The max-over-layers kNN score also falls to a comparable extent even though the walk never steers it directly (Fig.~\ref{fig:prh}A), indicating that HNC reorganizes its image representations in general rather than exploiting a particular metric on a particular image set. Throughout the walk, the alternative networks keep the anchor's top-1 predictions on the probe images. When evaluated on ImageNet, which was never used to constrain the HNC, the top-1 accuracy drops by less than 1\% and top-5 accuracy remains nearly unchanged (Fig.~\ref{fig:prh}B). 

After applying HNC, the network’s weights have changed by only 1.3\% in norm, yet its resulting representations have become less similar to the anchor's than every trained vision model we compare against, spanning different sizes, architectures, training objectives, and training data (Fig.~\ref{fig:prh}C). More remarkably, it is even less similar than comparing a randomly initialized ViT to the anchor (Fig.~\ref{fig:prh}C). The pairwise cosine similarities
between image representations on the probe set change drastically from the anchor to the HNC
endpoint (Fig.~\ref{fig:prh}D). A multidimensional scaling (MDS) embedding of 20 images from four classes further reveals that HNC reorganizes the neighborhood structure both within and across classes (Fig.~\ref{fig:prh}E).
To identify which aspects of visual similarity change, we select the five most similar image pairs in the anchor and HNC endpoint. We then score each pair's similarity based on either their semantic content (DINOv2), color composition, or spatial layout (Appendix~\ref{app:visualsim}). Like the anchor, the HNC endpoint's most similar image pairs remain close in semantic content and color. However, they have lower spatial-layout similarity, approaching the random-pair baseline (Fig.~\ref{fig:prh}F), indicating that the endpoint can represent images with different spatial layouts as similar. Overall, HNC reveals substantial representational freedom within a single model that is underexplored by standard gradient-based training, even in ViTs trained on challenging tasks. This suggests that rather than interpreting cross-model convergence as evidence for a unique underlying representation fully specified by the task, the observed similarity may arise due to an optimization bias that samples a narrow subset of solutions. 

\vspace{-1mm}
\section{HNC finds behaviorally distinct, high-reward policies in reinforcement learning}

\begin{figure}[H]
\centering
\vspace{-2mm}
\includegraphics[width=\textwidth]{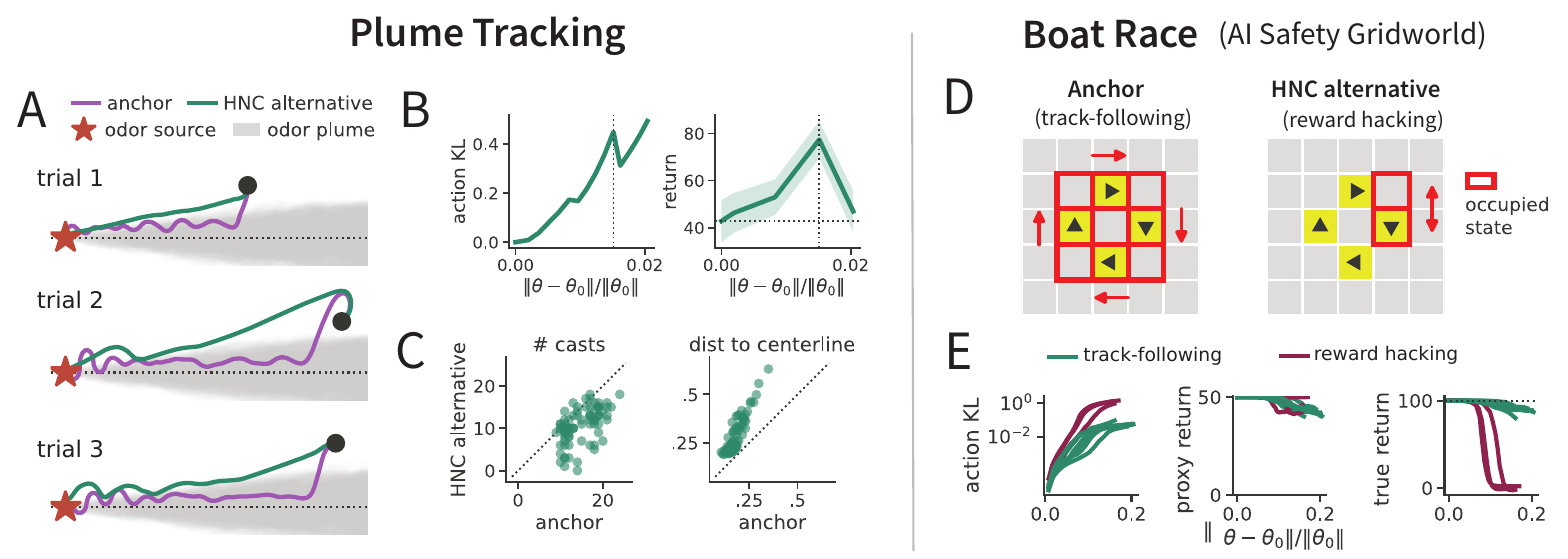}
\vspace{-8mm}
  \caption{\small \textbf{HNC finds behaviorally distinct, high-reward policies in RL.}
  \textbf{(A)} \textit{Plume Tracking} task where the agents navigate to the source of an odor plume, which travels away from the source under the dynamics of the wind in the environment. Anchor and alternative policy's navigation trajectory are shown for three trials. 
  \textbf{(B)}  Action divergence and return plotted against relative weight change during HNC (dotted vertical line: the policy that we analyze in (A) and (C)).
  \textbf{(C)} Number of casts, defined as the number of times the agent sweeps across the wind and reverses its heading direction, 
and mean distance from the plume centerline for 240 shared initial conditions. 
 \textbf{(D)} \textit{Boat race} task. The proxy reward gives $+3$ for entering an arrow tile clockwise; the true return  measures net clockwise progress around the track but is hidden to the agent during standard training. The anchor follows the track as intended, while HNC exposes a reward-hacking policy that steps on and off a single arrow tile, collecting the proxy reward without making clockwise progress.
\textbf{(E)} Action divergence, proxy return, and true return plotted against relative weight change during HNC for ten seeds.}
  \label{fig:rl}
  \vspace{-5mm}
\end{figure}

Unlike supervised learning, where the target output for each data sample is precisely
defined, in reinforcement learning (RL), the scalar reward rarely fully specifies the
optimal sequences of actions over an episode. This underspecification can admit
many behaviorally distinct policies with similar reward level, even unintended strategies that 
exploit the reward function rather than accomplish the intended task, a phenomenon known as reward hacking. 
We therefore apply HNC to search for behaviorally distinct solutions that
achieve comparable return to a trained policy.

Let $\pi_{\theta_0}$ be the policy anchor. We seek weights
$\theta$ whose actions differ from the anchor's while the return is unchanged.
The true reward function is usually unknown, but from an agent's perspective,
the return can be approximated by the off-policy importance sampling surrogate. Given a buffer of state-action pairs $\{(s_i,a_i)\}_{i=1}^{N}$ collected from rollouts of the anchor policy, we aim to preserve the surrogate reward
\(
  \phi_R(\theta) \;=\; \frac{1}{N}\sum_{i}
  \frac{\pi_\theta(a_i\mid s_i)}{\pi_{\theta_0}(a_i\mid s_i)}\,A^0_i \),
where $\pi_{\theta_0}(a_i\mid s_i)$ denotes the
probability of taking action $a_i$ in state $s_i$ and 
$A^0_i$ is the advantage of sample $i$ estimated from the buffer rollouts (Appendix~\ref{app:rl-adv}). We find
maximally different policies by steering the null space walk to maximize the behavioral divergence, defined by the KL divergence between the action distributions of the anchor and the alternative policy on the buffer states: 
\(
\phi_B(\theta)\;=\;\frac{1}{N}\sum_{i=1}^{N}
D_{\mathrm{KL}}\bigl(\pi_{\theta_0}(\cdot\mid s_i)\,\big\|\,\pi_{\theta}(\cdot\mid s_i)\bigr)
\). 
We periodically recollect the state--action buffer from the current policy to account for the changing state distribution, and recompute the surrogate reward and its null space along the walk. Details for continuous and discrete action spaces are provided in Appendix~\ref{app:phib}.

We applied HNC on Plume Tracking \citep{singh2023emergent} and AI Safety Gridworld \citep{leike2017ai} environment; additional MuJoCo results are provided in Appendix~\ref{app:rl_ant}. On the \textit{Plume Tracking} task \citep{singh2023emergent} (Fig.~\ref{fig:rl}A), RNNs simulating artificial flies are trained by deep RL to navigate to the source of a turbulent odor plume in a windy 2D
arena. Trained artificial flies resemble real flies by surging upwind when they detect odor and casting (sweeping back and forth across the wind) after losing it \citep{singh2023emergent} (Fig.~\ref{fig:rl}A). Over the HNC, the action divergence between anchor and the alternative network steadily increases, while the episode return
and the success rate (Appendix~\ref{app:plume_behavior}) are maintained, even increased early in the walk (Fig.~\ref{fig:rl}B). The alternative policy solves the
task in a visibly different way: instead of surging straight into the plume, the agent
slides to the edge of the plume and smoothly tracks it (Fig.~\ref{fig:rl}A). Behavioral quantification confirms
that the alternative policy casts less and on average stays farther from the centerline
of the plume (Fig.~\ref{fig:rl}C; details in Appendix~\ref{app:plume_behavior}). Strikingly, when evaluated
on out-of-distribution conditions with sparse odor or switching wind direction, the
alternative policy outperforms the anchor (Appendix~\ref{app:plume_ood}).

We next ask whether HNC can expose reward-hacking policies near a well-behaved solution in the AI Safety Gridworlds environment \citep{leike2017ai}.
In the \emph{boat race} task, an agent earns a proxy reward of $+3$ each time it enters an arrow tile clockwise (Fig.~\ref{fig:rl}D). The true return, which the agent never sees, instead measures net clockwise progress. The proxy therefore admits a known exploit: stepping on and off a single arrow tile collects the proxy reward repeatedly without net progress. We train MLP policies with PPO on the true return, which yields track-following behavior in all $10$ seeds (Appendix~\ref{app:boat}).
Starting from these anchors, we use HNC to search for behaviorally distinct policies while preserving the proxy return. Behavioral divergence increases steadily along every null space walk, and the endpoints separate into two clusters (Fig.~\ref{fig:rl}E). The three most divergent endpoints adopt the reward-hacking strategy, where their net clockwise progress falls to zero. Figure \ref{fig:rl}D contrasts the state occupancy of an anchor and its HNC endpoint, showing a transition from track-following to oscillation around a single arrow. Overall, these results demonstrate that HNC can find behaviorally distinct yet reward-matched policies nearby in weight space in RL-trained networks, and can expose the underspecification in reward design that admits reward hacking solutions.

\vspace{-2mm}
\section{HNC measures loss landscape geometry}
\vspace{-1mm}
\label{sec:geometry}

\begin{wrapfigure}{r}{0.45\textwidth}
\centering
\vspace{-4mm}
\includegraphics[width=\linewidth]{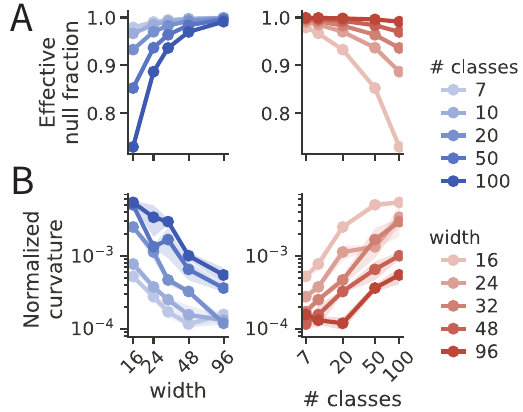}
\vspace{-6mm}
\caption{\small
\textbf{Model size and task difficulty shape local flatness and curvature.}
\textbf{(A)} Effective null fraction as a function of model width and task difficulty (number of image classes).
\textbf{(B)} Normalized curvature along HNC-stepped directions. Curvature is summarized by the median over steps within each walk. Shading indicates standard error across 3 seeds.}
\label{fig:scaling}
\vspace{-4mm}
\end{wrapfigure}

Beyond finding alternative solutions, HNC provides two complementary measures of loss-landscape geometry: the Hessian characterizes the local geometry at an anchor, while the null space walk tracks how that geometry changes along a trajectory. We use these measurements to ask how model size and task complexity shape the solution set. Previous work has addressed this question through loss barriers between independently trained networks \citep{garipov2018loss,draxler2018essentially,frankle2020linear,entezari2021role}, or through stylized theoretical models \citep{cooper2018loss,simsek2021geometry}. HNC instead enables direct measurements from a single trained network. Here, we train CNNs on CIFAR-100 subsets while varying network width and the number of classes, our proxy for task complexity (Appendix~\ref{app:cnn-setup}).

We first estimate the \textbf{effective null fraction}, the fraction of parameter directions along which a small step leaves the outputs approximately unchanged. We set the spectral cutoff using a common function-drift tolerance, where a step of $1\%$ of the weight norm along a flat direction raises the function-matching loss in \eqref{eq:funcloss} by at most $\epsilon = 0.05$ (Appendix~\ref{app:landscape}). This ties the cutoff to an interpretable output-drift tolerance, allowing comparisons across conditions with different overall Hessian scales. The trends below hold across different magnitudes of $\epsilon$ (Appendix~\ref{app:drift-sweep}). The effective null fraction grows with width and shrinks with the number of classes (Fig.~\ref{fig:scaling}A). Nevertheless, even the smallest network on the hardest task retains a large effective null fraction. We next ask whether smaller networks and harder tasks instead increase the functional cost of moving along these directions.

We therefore measured the \textbf{normalized curvature} along the HNC walk, $d^\top H d/(nC)$, where $d$ is a unit-norm step direction, $n$ is the number of probe inputs, and $C$ is the number of output classes. This quantity measures how sensitive the outputs are to small weight changes along $d$. To avoid constraining the output sensitivity, we replaced the output-drift cutoff with a common relative flatness threshold, $\mu=10^{-6}\lambda_{\max}$. Normalized curvature generally increases with task complexity and decreases with width (Fig.~\ref{fig:scaling}B), and these trends persist across the relative thresholds tested (Appendix~\ref{app:curv-threshold}). Although approximately flat directions remain abundant across the sweep, the directions HNC follows become more sensitive to weight changes on harder tasks. This indicates that the landscape around a local minimum stiffens and increases in curvature for harder tasks, thereby providing a potential mechanism for prior work that suggests that the loss landscape fragments into separate basins as tasks get harder  \citep{frankle2020linear,entezari2021role,simsek2021geometry}.

\vspace{-1.5mm}
\section{Discussion}
\vspace{-1mm}

We introduced Hessian Null Space Continuation (HNC), a scalable, domain-agnostic method for traversing the connected solution space around a trained network while preserving its input-output mapping. Across RNNs, Vision Transformers, and reinforcement-learning policies, HNC uncovered solutions with substantially different representations, dynamics, and behavioral strategies. It also provided local geometric measurements of the solution set, revealing how model size and task complexity shape its dimension and functional sensitivity. 
\vspace{-0.4mm}

Our geometric results in Section \ref{sec:geometry} complement \citet{huang2025measuring}, which found that harder tasks reduce variability in neural dynamics across independently trained RNNs. They also found that inter-model variability decreases with model size, whereas our results show that larger models have larger null spaces on the same task. Together, these findings suggest that larger models admit more alternative solutions within a local region of weight space, even as training converges to a more narrow subset of them. This is consistent with the stronger simplicity bias of large models proposed to explain representational convergence across models \citep{huh2024platonic}. 
Prior work has shown that parameter symmetries and reparameterizations can also reorganize representations while preserving function \citep{theiss_parameter_2026, theiss_representation_2026}. However, the diverse solutions found by HNC go beyond reparameterizations of a trained solution: they involve qualitative changes in the network's dynamical regime, altered out-of-distribution generalization, and qualitatively distinct behavioral strategies. Moreover, whereas the parameter symmetries studied by \citet{theiss_parameter_2026} involve discrete operations that may change the size of the network, such as neuron duplication, HNC makes continuous, small changes to the weights of a trained network to reach distinct solutions. 

One limitation of HNC is that function preservation is approximate and enforced on a finite probe set. Small function-matching loss therefore does not guarantee output agreement on unseen inputs, particularly under distribution shift. The diversity accessible to HNC depends on both probe coverage and the allowed output drift, making held-out functional evaluation important when interpreting the resulting alternatives (Appendix \ref{app:3bff-ood}, \ref{app:prh_cka_function}, \ref{app:prh_knn_function}). HNC is also a local search algorithm, making it agnostic to the existence of multiple solution basins. However, we found that alternating with a global method such as that in \cite{ostrow_metric_2026} enables broader search (Appendix \ref{app:search}).

Beyond the input-output mapping, the preserved quantity can be any differentiable function of the network. Replacing the function-matching loss with the task loss would instead allow HNC to explore the broader set of task-compatible networks (Appendix \ref{app:func-vs-task}). The steering objective is equally flexible and could encode performance on a second task to obtain multitask networks or alignment with neural data (Appendix~\ref{app:brainscore}). The same Hessian spectrum also provides a complementary prescription when the goal is to change the network's function: flat directions preserve the function, whereas sharp directions are most effective for changing it \citep{gan2026thickets, liang2026blessing}. 

More broadly, HNC allows us to constructively explore the solution space and compare the particular solution reached by training with the function-preserving alternatives around it. This comparison helps distinguish properties demanded by the task from those specific to an individual training run \citep{damour_underspecication_nodate, fisher_all_2018}. In this respect, HNC enables empirical studies of the optimizer's implicit bias and complements approaches that estimate implicit regularization from discrepancies between weight updates and loss gradients \citep{rudoler2026estimating} or from the displacement of Hessian eigenvectors during training \citep{marjankowska2026characterizing}. Our method could also be further scaled to large language models, for which second-order information has already been used to improve optimization \citep{martens2015optimizing, george2018fast, liu2024sophia, grosse2023studying}. Finally, the function-preserving solutions exposed by HNC provide a natural geometric setting for model merging, fine-tuning, and editing \citep{wortsman_model_2022, ilharco_editing_2022, ainsworth2023git, meng_locating_2022, mitchell_fast_2022, gan2026thickets}, as well as for auditing reward specifications by uncovering behaviorally distinct or reward-hacking policies \citep{amodei2016concrete,leike2017ai,skalse2022defining}.

\section*{Acknowledgments}
Funded by NIH (RF1DA056403 to K.R.), 
James S. McDonnell Foundation (220020466 to K.R.), 
Simons Foundation (Pilot Extension-00003332-02 to K.R.), 
McKnight Endowment Fund (K.R.), 
CIFAR Azrieli Global Scholar Program (K.R.), 
NSF (2046583 to K.R.), 
Harvard Medical School Neurobiology Lefler Small Grant Award (K.R.), 
Harvard Medical School Dean's Innovation Award (K.R.), and Army Research Office (W911NF-26-1-A201 to W.T.R.)
A.H is supported by the Kempner Graduate Fellowship. M.O. is funded by the NSF GRFP.

A.H. wishes to give special thanks to Binxu Wang, Flavio Martinelli, Billy Qian, Tatiana Engel, Daniel Yamins, Satpreet Singh, and all members of the Rajan Lab for helpful discussions. W.T.R. thanks Yannis Kevrekidis for inspiring interest in different solutions.

\clearpage
\bibliography{iclr2027_conference}
\bibliographystyle{iclr2027_conference}

\appendix
\newcommand{\appsec}[1]{\item[] \hyperref[#1]{\textbf{\ref{#1}\quad\nameref{#1}}}\ \dotfill\ \pageref{#1}}
\newcommand{\appsub}[1]{\item[] \hyperref[#1]{\ref{#1}\quad\nameref{#1}}\ \dotfill\ \pageref{#1}}
\section*{Appendix Outline}
\begin{itemize}\setlength{\itemsep}{0pt}\setlength{\parskip}{0pt}\setlength{\topsep}{1pt}\setlength{\partopsep}{0pt}
  \item[] \textbf{Method}\vspace{1pt}
  \appsec{app:subroutines}
  \begin{itemize}\setlength{\itemsep}{0pt}\setlength{\parskip}{0pt}\setlength{\topsep}{1pt}\setlength{\partopsep}{0pt}
    \appsub{app:complexity}
    \appsub{app:lobpcg}
    \appsub{app:solver}
  \end{itemize}
  \appsec{app:gn}
  \appsec{app:ablation}
  \begin{itemize}\setlength{\itemsep}{0pt}\setlength{\parskip}{0pt}\setlength{\topsep}{1pt}\setlength{\partopsep}{0pt}
    \appsub{app:abl-corrector}
    \appsub{app:abl-null}
    \appsub{app:abl-relin}
    \appsub{app:abl-penalty}
  \end{itemize}
  \appsec{app:func-vs-task}
  \appsec{app:search}
  \vspace{\baselineskip}
  \item[] \textbf{Experiments}\vspace{1pt}
  \appsec{app:rnn}
  \begin{itemize}\setlength{\itemsep}{0pt}\setlength{\parskip}{0pt}\setlength{\topsep}{1pt}\setlength{\partopsep}{0pt}
    \appsub{app:3bff-setup}
    \appsub{app:flat-threshold}
    \appsub{app:walk-settings}
    \appsub{app:3bff-gallery}
    \appsub{app:3bff-mds}
    \appsub{app:sparsity}
    \appsub{app:sgd_stable}
    \appsub{app:3bff-ood}
  \end{itemize}
  \appsec{app:prh}
  \begin{itemize}\setlength{\itemsep}{0pt}\setlength{\parskip}{0pt}\setlength{\topsep}{1pt}\setlength{\partopsep}{0pt}
    \appsub{app:softmaxcka}
    \appsub{app:visualsim}
    \appsub{app:prh_knn}
    \appsub{app:vit-walk-settings}
    \appsub{app:prh_cka_function}
    \appsub{app:prh_knn_function}
    \appsub{app:prh_loci}
    \appsub{app:prh_vits_vitb}
    \appsub{app:probesize}
    \appsub{app:prh_mds}
  \end{itemize}
  \appsec{app:brainscore}
  \appsec{app:rl}
  \begin{itemize}\setlength{\itemsep}{0pt}\setlength{\parskip}{0pt}\setlength{\topsep}{1pt}\setlength{\partopsep}{0pt}
    \appsub{app:rl-adv}
    \appsub{app:phib}
    \appsub{app:rl-walk-settings}
    \appsub{app:plume_behavior}
    \appsub{app:plume_ood}
    \appsub{app:boat}
    \appsub{app:rl_ant}
  \end{itemize}
  \appsec{app:landscape}
  \begin{itemize}\setlength{\itemsep}{0pt}\setlength{\parskip}{0pt}\setlength{\topsep}{1pt}\setlength{\partopsep}{0pt}
    \appsub{app:cnn-setup}
    \appsub{app:drift-sweep}
    \appsub{app:curv-threshold}
    \appsub{app:repdim}
  \end{itemize}
  \vspace{2pt}
\end{itemize}
\clearpage
\renewcommand{\needlines}[1]{\par\vskip0pt\begingroup\dimen0=#1\baselineskip\relax
  \ifdim\pagegoal=\maxdimen\else\ifdim\dimexpr\pagegoal-\pagetotal\relax<\dimen0 \newpage\fi\fi\endgroup}






\section{Hessian Null-Space Continuation pseudocode}
\label{app:subroutines}

After the restore steps, HNC accepts a step only if the loss stays below a ceiling $\tau$. The
step-size rule and relinearization schedule are described per experiment
(Appendices~\ref{app:walk-settings}, \ref{app:vit-walk-settings} and~\ref{app:rl-walk-settings}).
Following the discussion in Appendix~\ref{app:gn}, Algorithms~\ref{alg:nullspace} and~\ref{alg:stepdir}
use the Gauss--Newton operator $\GN=\mJ^\top\mJ$ in place of the Hessian $\mH$, because it equals
$\mH$ at the anchor, stays positive semidefinite along the walk, and is cheaper to apply.

\begin{algorithm}[H]
\caption{Hessian null-space continuation (HNC)}
\label{alg:hnc}
\begin{algorithmic}[1]
\REQUIRE trained anchor $\vtheta_0$, probe set $\mX$; null-space step size $\eta$, restore step size $\alpha$
and number of restore steps $m$, continuation steps $T$, null threshold $\epsilon$, block size $k$, loss
ceiling $\tau$; \emph{optional} potential $\varphi$ with damping $\mu$
\STATE $\vtheta \gets \vtheta_0$;\quad $\mathcal{L}$ as in \eqref{eq:funcloss};\quad
$\Vzero\gets\textsc{NullSpace}(\vtheta,\epsilon,k)$
\COMMENT{the initial null-space; Alg.~\ref{alg:nullspace}}
\FOR{$t=1,\dots,T$}
  \STATE $\vd\gets\textsc{StepDir}(\vtheta,\Vzero,\varphi,\mu)$
\COMMENT{step direction; Alg.~\ref{alg:stepdir}}
  \STATE $\tilde{\vtheta}\gets\vtheta+\eta\,\vd/\lVert\vd\rVert$
\COMMENT{null-space step}
  \FOR{$j=1,\dots,m$}
    \STATE $\tilde{\vtheta}\gets\tilde{\vtheta}-\alpha\,\nabla\mathcal{L}(\tilde{\vtheta})$
\COMMENT{project back onto $\mathcal{L}\!\approx\!0$}
  \ENDFOR
  \IF{$\mathcal{L}(\tilde{\vtheta})>\tau$}
    \STATE shrink $\eta$;\quad $\Vzero\gets\textsc{NullSpace}(\vtheta,\epsilon,k)$
\COMMENT{reject: keep $\vtheta$, relinearize}
  \ELSE
    \STATE $\vtheta\gets\tilde{\vtheta}$
\COMMENT{accept}
  \ENDIF
  \STATE relinearize periodically, $\Vzero\gets\textsc{NullSpace}(\vtheta,\epsilon,k)$
\COMMENT{relinearization schedule}
  \STATE record $\vtheta_t\gets\vtheta$
\ENDFOR
\RETURN a trajectory of function-preserving solutions $\{\vtheta_t\}_{t=1}^{T}$
\end{algorithmic}
\end{algorithm}

\begin{algorithm}[H]
\caption{\textsc{NullSpace}$(\vtheta,\epsilon,k)$: near-null (flat) subspace of the loss at $\vtheta$}
\label{alg:nullspace}
\begin{algorithmic}[1]
\REQUIRE weights $\vtheta$, null threshold $\epsilon$, block size $k$
\STATE define the matrix-free Gauss--Newton operator $\GN\!:\vv\mapsto\mJ^\top(\mJ\vv)$ at $\vtheta$
\STATE $\{(\lambda_i,\vv_i)\}_{i=1}^{k}\gets\textsc{LOBPCG}(\GN,k)$ \COMMENT{bottom $k$ eigenpairs of the spectrum}
\RETURN $\Vzero=\mathrm{span}\{\vv_i:\lambda_i<\epsilon\}$
\end{algorithmic}
\end{algorithm}

\begin{algorithm}[H]
\caption{\textsc{StepDir}$(\vtheta,\Vzero,\varphi,\mu)$: null-space step direction inside the flat subspace}
\label{alg:stepdir}
\begin{algorithmic}[1]
\REQUIRE weights $\vtheta$, flat subspace $\Vzero$; \emph{optional} potential $\varphi$ with damping $\mu$
\IF{potential $\varphi$ is given}
  \RETURN $(\mI+\GN/\mu)^{-1}\nabla\varphi(\vtheta)$ \COMMENT{\emph{steering}}
\ELSE
  \RETURN any $\vd\in\Vzero$ \COMMENT{undirected exploration}
\ENDIF
\end{algorithmic}
\end{algorithm}

\subsection{Computational cost}
\label{app:complexity}

HNC is matrix-free and accesses the Hessian $\mH$ only through Hessian--vector products $\mH\vv$, which automatic differentiation computes without explicitly forming $\mH$ \citep{pearlmutter1994fast}. Each product costs approximately one gradient evaluation and requires memory linear in the number of parameters $P$.
A dense approach instead must store the $P\times P$ Hessian, requiring $\mathcal{O}(P^2)$ memory, and its eigendecomposition costs $\mathcal{O}(P^3)$ time. In contrast, LOBPCG applies $\mH$ to a block of $k$ vectors and stores only this block. For fixed $k$ and iteration count, its time and memory scale as $\mathcal{O}(kP)$. The dense approach is feasible, and faster due to lower overhead, for the $4{,}611$-parameter 3BFF RNN, where we use it to validate the iterative solver. For the 22-million-parameter ViT-S, however, the dense Hessian alone would require approximately $2$~PB of memory, whereas the matrix-free approach remains practical. Table~\ref{tab:complexity} summarizes the computational costs.

\clearpage
\begin{table}[h]
\centering
\caption{Cost of a dense eigendecomposition of the Hessian against the matrix-free LOBPCG
route used by HNC. $P$ is the number of parameters and $k$ the block size. RNN timings are for the 3BFF anchor of Section~\ref{sec:rnn} on one
CPU node with $8$ threads; ViT-S timings are from the walks of Section~\ref{sec:prh} on one NVIDIA H200 with 512 probe images.}
\label{tab:complexity}
\vspace{0.3cm}
\setlength{\arrayrulewidth}{1pt}
\begin{tabular}{@{}lp{4.6cm}p{4.6cm}@{}}
\hline
 & Dense eigendecomposition & Matrix-free LOBPCG \\[0.1cm]
\hline
\noalign{\global\arrayrulewidth=0.4pt \vskip 0.1cm}
Time & $\mathcal{O}(P^3)$ & $\mathcal{O}(kP)$\\
Memory & $\mathcal{O}(P^2)$ & $\mathcal{O}(kP)$ \\[0.05cm]
\hline
\noalign{\vskip 0.05cm}
Memory, 3BFF RNN ($P=4611$) & $85$~MB & $1.2$~MB ($k=64$) \\
Time, 3BFF RNN & $10$~s & $39$~s ($k=64$) \\[0.05cm]
\hline
\noalign{\vskip 0.05cm}
Memory, ViT-S ($P=22$M) & $1.9$~PB & $5.6$~GB ($k=64$) \\
Time, ViT-S & $2\times10^{7}$ products to assemble ($\sim$months of GPU time), factorization infeasible & $\approx 1$~s per product; one continuation step $\approx 1$~min \\[0.05cm]
\hline
\end{tabular}
\end{table}

\subsection{Finding the flattest directions by LOBPCG}
\label{app:lobpcg}

Section~\ref{sec:nullspace} needs the $k$ eigenvectors of $\mH$ with the smallest eigenvalues,
which are the directions with flattest curvature. LOBPCG \citep{knyazev2001toward} finds them by
minimizing the Rayleigh quotient
\begin{equation}
\rho(\vv)\;=\;\frac{\vv^\top\mH\vv}{\vv^\top\vv},
\label{eq:rayleigh}
\end{equation}
which measures the curvature of the function-matching loss along a parameter direction $\vv$.
Its minimum over all $\vv$ is the smallest eigenvalue, attained at the corresponding
eigenvector. LOBPCG minimizes \eqref{eq:rayleigh} jointly over a block of $k$ mutually
orthogonal directions, which at convergence span the bottom of the spectrum. Each iteration
updates the block using the current directions, the residuals $\mH\vv-\rho(\vv)\vv$, and the
previous iteration's directions, which is the ``locally optimal'' part of the method and plays
the same role as the momentum term in conjugate gradients. The block is accessed only through the products $\mH\vv$, so the solver never explicitly forms $\mH$. We
compute these products by automatic differentiation \citep{pearlmutter1994fast}, at a cost of
roughly one gradient evaluation each (Appendix~\ref{app:complexity}). 

\subsection{Solving the soft projection by conjugate gradients}
\label{app:solver}
The steered direction $\vd=(\mI+\GN/\mu)^{-1}\nabla\varphi$ of Algorithm~\ref{alg:stepdir}
is the solution of the linear system $(\mI+\GN/\mu)\,\vd=\nabla\varphi$. The matrix
$\mI+\GN/\mu$ is symmetric positive definite, so we solve the system with the conjugate
gradient method (CG; \citealp{hestenes1952methods}), which touches $\GN$ only through products
$\GN\vv$. Each product is one forward-mode Jacobian--vector product followed by one
reverse-mode product, $\mJ^\top(\mJ\vv)$, and costs about as much as one gradient
evaluation, so a solve with $n_{\mathrm{CG}}$ iterations costs about $n_{\mathrm{CG}}$
gradient evaluations and never forms $\GN$. We run a fixed budget of $n_{\mathrm{CG}}$
iterations and exit early once the relative residual falls below $10^{-8}$. Along a walk the
solution changes slowly from step to step, so the solve is warm-started at the previous step's
direction. The exception is the 3BFF walks of Fig.~\ref{fig:manifold}, which start every
solve from zero with a larger iteration budget (Tables~\ref{tab:cka-settings} and
\ref{tab:dsa-settings}). Because the operator is re-formed at the current weights at every
step, the steered walks need no separate relinearization.

\section{Gauss--Newton approximation of the Hessian at loss minima}
\label{app:gn}

Our methods reads the flat directions of the loss from a curvature operator that is cheaper
than the full Hessian, the Gauss-Newton operator $\GN=\mJ^\top\mJ$ which provides close approximation
of the full Hessian around loss minimizers.

Write the residual $\vr(\vtheta)=f_\vtheta(\mX)-f_{\vtheta_0}(\mX)$, so that
$\mathcal{L}(\vtheta)=\tfrac12\lVert\vr(\vtheta)\rVert^2$, and let
$\mJ=\partial\vr/\partial\vtheta=\partial f_\vtheta(\mX)/\partial\vtheta$ be its Jacobian (the anchor
outputs $f_{\vtheta_0}(\mX)$ are constant). The gradient is $\nabla\mathcal{L}=\mJ^\top\vr$, and
differentiating once more gives
\begin{equation}
\nabla^2\mathcal{L}(\vtheta) \;=\; \underbrace{\mJ^\top\mJ}_{\GN}
\;+\; \sum_{k} r_k(\vtheta)\,\nabla^2 r_k(\vtheta).
\label{eq:hessian-split}
\end{equation}
The first term is the Gauss--Newton operator; the second is a residual-weighted sum of per-output
curvatures. By construction the anchor is a loss minimizer with near-zero residual,
$\vr(\vtheta_0)=f_{\vtheta_0}(\mX)-f_{\vtheta_0}(\mX)=\vzero$, so the second term vanishes and
\begin{equation}
\nabla^2\mathcal{L}(\vtheta_0)\;=\;\mJ^\top\mJ\;=\;\GN.
\label{eq:hessian-eq-gn}
\end{equation}

Because $\GN=\mJ^\top\mJ$ is symmetric positive semidefinite, its
eigenvalues are real and nonnegative, so the flat (near-null) subspace is unambiguously the bottom of
its spectrum, with no spurious negative-curvature directions to disentangle. And $\GN$ never has to be
formed: a Gauss--Newton--vector product is one Jacobian-vector product followed by one
vector-Jacobian product, which is what keeps the null-space computations
\autoref{alg:hnc} affordable at scale. 

\section{Ablation study on HNC algorithm}
\label{app:ablation}

We ablate the main components of HNC using RNNs trained on the 3BFF task (Appendix~H.1). Unless stated otherwise, we steer the walks to minimize CKA similarity to the anchor. The ablations address three questions: (1) Do the function-restoring steps correct the higher-order drift accumulated during the null-space walk? (2) Does projecting the steering gradient onto the null-space improve function preservation? (3) Because the null-space is only a local approximation, does relinearization reduce function drift as the walk moves away from the anchor, and how does its frequency affect the outcome?

Finally, we compare HNC with a direct penalty-based alternative that jointly optimizes the steering objective and task loss:
\begin{equation}
\max_{\vtheta}\;\varphi(\vtheta)-\beta\mathcal{L}_{\mathrm{task}}(\vtheta),
\label{eq:penalty}
\end{equation}
where $\beta$ controls the trade-off between divergence and task performance. This comparison tests whether explicitly following the evolving null-space allows HNC to find more dissimilar solutions at a matched task-loss.

\subsection{Function-restoring step}
\label{app:abl-corrector}

\begin{figure}[H]
\centering
\includegraphics[width=\textwidth]{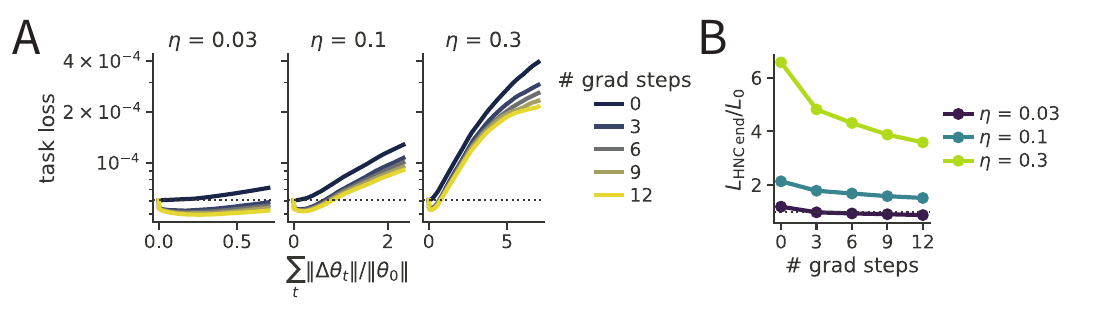}
\caption{\textbf{Ablation experiment on the function-restoring step.} Undirected null-space walks at three
null-space step sizes $\eta$, with $m\in\{0,3,6,9,12\}$ restore steps. \textbf{(A)} The task
loss against the normalized weight travel during HNC walks, where the dotted line is the anchor's loss. \textbf{(B)} HNC endpoint's
task loss relative to the anchor's against $m$. Larger null-space steps drift faster, and the restore step
corrects for part of the drift rather than eliminating it.}
\label{fig:abl-corrector}
\end{figure}

After each null-space step, HNC takes $m$ gradient steps on the function-matching loss to pull the network
back toward the low-loss region. \autoref{fig:abl-corrector} varies $m$ on an undirected walk at three
null-space step sizes $\eta$. More function-restoring steps reduce the accumulated function drift
and therefore lower the task loss at every step size, with the largest effect at the largest
step. At the smallest step size the benefit saturates after a few restore steps, and the walk
ends at or below the anchor's loss. At $\eta=0.3$, twelve restore steps lower the endpoint task
loss from $6.6$ times the anchor's loss to $3.6$ times, with diminishing returns from each
additional step. The function-restoring step therefore does not eliminate the drift induced by large
null-space steps, and taking smaller steps within the null-space protects the function more
effectively than correcting afterwards. The function-restoring step is best thought of as a
safeguard that limits drift rather than a projection that removes it, which is why our HNC walks are combined with a small step size, an adaptive step schedule, and a loss ceiling.

\subsection{Null-space projection of the steering gradient}
\label{app:abl-null}

\begin{figure}[H]
\centering
\vspace{-4mm}
\includegraphics[width=0.6\textwidth]{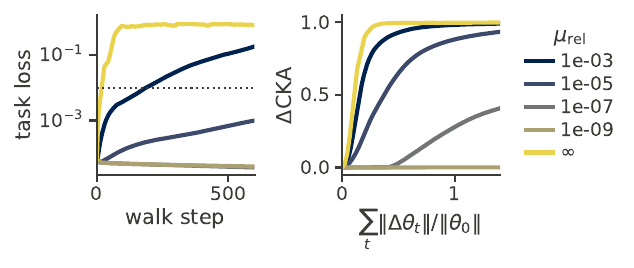}
\caption{\textbf{Ablation experiment on the null-space projection of the steering gradient.} CKA-steered walks from the same 3BFF anchor network at varying magnitudes of damping. Here, the
maximum loss threshold is removed so that we can compare the task loss reached during HNC under
different damping. A larger damping factor allows the steering gradient along sharper directions
to pass through, therefore leading to faster representational divergence at the price of higher
task loss. In contrast, with a highly strict damping factor ($\mu_{\mathrm{rel}}=10^{-9}$),
neither the task loss nor the representation changes at all.}
\vspace{-4mm}
\label{fig:abl-proj}
\end{figure}

When the walk is steered by a differentiable objective $\varphi$, we project its gradient onto
the null-space before taking the step. The soft projection is
\begin{equation}
\vd \;=\; (\mI + \mH/\mu)^{-1}\,\nabla\varphi(\vtheta),
\label{eq:softproj}
\end{equation}
where $\mu>0$ is a damping factor. Along each eigenvector $\vv_i$ of the Hessian, this
projection rescales the gradient component by $1/(1+\lambda_i/\mu)$, where $\lambda_i$ is the
corresponding eigenvalue. The factor is close to one for flat directions ($\lambda_i\ll\mu$)
and close to zero for sharp directions ($\lambda_i\gg\mu$), so the projection keeps the
gradient along flat directions and removes it along sharp ones. A larger $\mu$ therefore
retains the gradient along sharper directions.
\autoref{fig:abl-proj} shows the task loss and the
representational divergence $\Delta\mathrm{CKA}$ along HNC for several values of
$\mu_{\mathrm{rel}}$, where $\mu_{\mathrm{rel}}=\infty$ denotes no projection at all.

A larger damping factor lets the walk diverge faster, but the task loss grows with it. At
$\mu_{\mathrm{rel}}=10^{-7}$ the task loss stays close to the anchor's while
$\Delta\mathrm{CKA}$ rises steadily. Increasing $\mu_{\mathrm{rel}}$ to $10^{-5}$ and
$10^{-3}$ speeds up the divergence, at the cost of a clearly higher task loss. With no
projection at all ($\mu_{\mathrm{rel}}=\infty$), the walk follows the raw CKA gradient: the
steering objective rises fastest of all, but the task loss rises just as rapidly and the
function breaks within a few dozen steps, even with the restore step still in place. At the
other extreme, the smallest damping factor ($10^{-9}$) leaves both the task loss and the
representation unchanged. The
magnitude of $\mu$ therefore trades how far the steering objective moves against how much function drifts.

\subsection{Relinearization}
\label{app:abl-relin}

\begin{figure}[H]
\centering
\includegraphics[width=\textwidth]{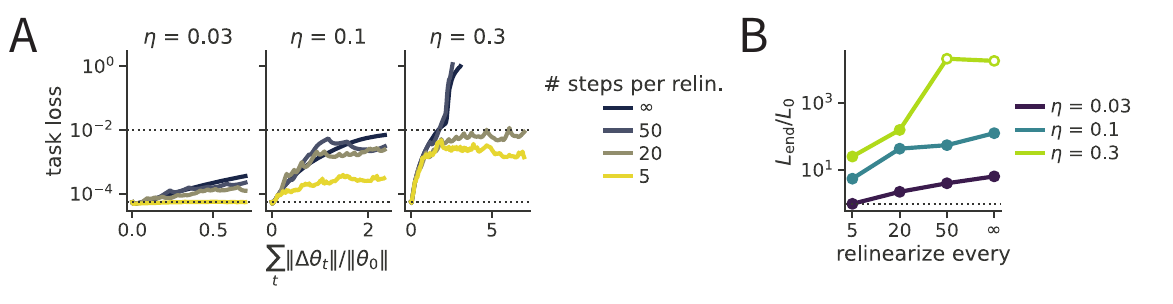}
\vspace{-4mm}
\caption{\textbf{Ablation experiment on the relinearization step.} \textbf{(A)} Constraint task loss against the normalized weight travel; dotted lines mark the anchor's
loss and the walk's usual ceiling. \textbf{(B)} Endpoint task loss relative to the anchor's against different
relinearization interval. Open markers mark walks whose loss blew up. }
\label{fig:abl-relin}
\end{figure}

The null-space is computed at a given weight configuration, so as the walk moves away from
those weights, the curvature along the null-space directions can change and the directions can
stop being flat. Relinearization recomputes the null-space at the current weights. To isolate
its effect, the walks in \autoref{fig:abl-relin} use a fixed null-space step size and
relinearize on a fixed schedule: never, or every 50, 20, or 5 steps.

More frequent relinearization always lowers the task loss relative to less frequent or no
relinearization. When the step size is small, frequent relinearization matters less, but for a
large step size it is critical to keep the task loss from blowing up (\autoref{fig:abl-relin}).

\subsection{Comparison with penalized optimization}
\label{app:abl-penalty}
\begin{figure}[H]
\centering
\vspace{-4mm}
\includegraphics[width=0.8\textwidth]{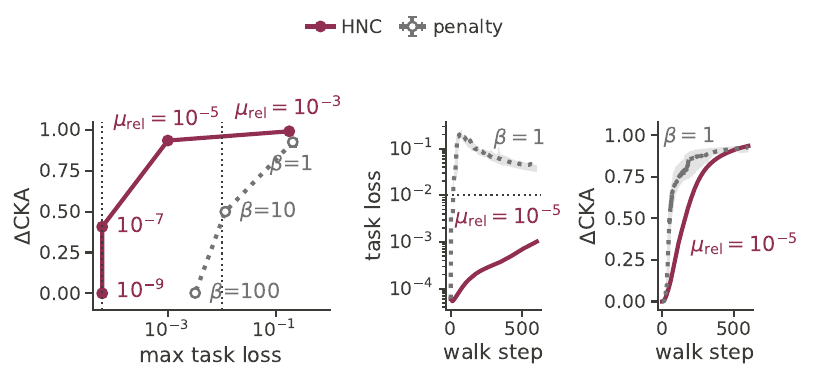}
\caption{\textbf{HNC against training with a similarity penalty.} \textbf{Left:} the highest
task loss reached during 600 steps against the divergence from the anchor,
$\Delta\mathrm{CKA}=1-\mathrm{CKA}$, for HNC at four flatness thresholds $\mu_{rel}$ and for penalized training
at three penalty weights $\beta$ (five seeds each; markers show the median, bars the range).
Dotted lines mark the anchor's loss and the loss ceiling. \textbf{Middle, right:} task loss and
divergence over the same 600 steps for one HNC walk ($\mu_{\mathrm{rel}}=10^{-5}$) and one
penalized run ($\beta=1$) that end at similar divergence. At every level of task loss, HNC
reaches more divergence. The penalized run crosses the loss ceiling within its first twenty
steps, gains nearly all of its divergence during that excursion, and only then returns toward
low loss, whereas HNC never leaves the low-loss region.}
\label{fig:abl-penalty}
\end{figure}

The most straightforward way to find a dissimilar solution is to train on the task loss with a penalty on
similarity to the anchor, as in \eqref{eq:penalty}. This needs no curvature information: the
network follows the combined gradient with steps of the same length as the HNC walk, and the
weight $\beta$ sets how strongly similarity is penalized. We compare this against HNC over the
same 600 steps, running HNC at four damping factors and the penalized training at three values
of $\beta$ (\autoref{fig:abl-penalty}).

At every level of task loss, HNC reaches a larger divergence from the anchor than penalized
training. The two methods also arrive at their divergence in different ways. Penalized training
crosses the loss ceiling within its first 20 steps, gains nearly all of its divergence
while the task loss is high, and only then returns toward low loss. In comparison, HNC never leaves the
low-loss region and gains its divergence steadily. 

\section{Preserving function versus task loss in HNC}
\label{app:func-vs-task}

HNC holds the function-matching loss of \eqref{eq:funcloss} approximately fixed rather than the task loss on
which the network was trained. Preserving the task loss instead would let HNC explore the broader
set of task-compatible networks, so here we discuss why we choose to preserve the function-matching loss instead in the paper. 

\paragraph{The anchor is an exact minimum of the function-matching loss but only an approximate
minimum of the task loss.}
A flat direction of the Hessian keeps a loss constant only if the gradient of that loss is also
zero. Training stops when the task-loss gradient $\vg$ is small, not when it vanishes, so a step
$\eta\vv$ along a flat direction still raises the task loss by $\eta\,\vg^\top\vv$. The null-space step
removes the second-order term but still carries the first-order term. Because the gradient term is linear in the step size while the curvature term is quadratic, it is the larger of the two at the small step sizes that HNC takes. Walking a flat subspace of the task loss therefore can drift away from
the task the network was trained on, and the drift accumulates step after step. The
function-matching loss of \eqref{eq:funcloss} has zero gradient at the anchor by construction,
because the anchor reproduces its own outputs exactly. Its flat directions are therefore flat in both first and second order,
which is why HNC preserves it rather than the task loss.

The two losses also differ in the signs of their Hessian eigenvalues. The function-matching Hessian at the anchor is positive semidefinite (Appendix~\ref{app:gn}), so its eigenvalues are all nonnegative. Its algebraically smallest eigenvalues are therefore also those closest to zero and identify the flattest directions. On the other hand, the task-loss Hessian contains an additional term:
\begin{equation}
\nabla^2\mathcal{L}_{\mathrm{task}}
= \mJ^\top\bigl(\nabla_f^2\ell\bigr)\mJ
+ \sum_k \frac{\partial\ell}{\partial f_k}\,\nabla^2 f_k .
\label{eq:task-hessian}
\end{equation}
The output Hessians $\nabla^2 f_k$ have no sign constraint, so the additional term can outweigh the nonnegative Gauss–Newton term in some directions, giving the task-loss Hessian both positive and negative eigenvalues and making it indefinite. On 3BFF, we find several hundred negative eigenvalues at every trained anchor. Selecting the algebraically smallest eigenvalues therefore favors negative-curvature directions over nearly flat directions, whose eigenvalues are closest to zero.

\begin{figure}[H]
\centering
\includegraphics[width=0.6\linewidth]{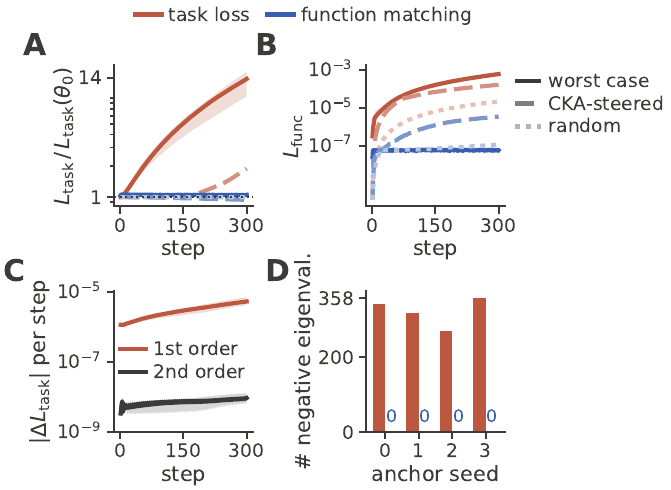}
\caption{\textbf{Task-loss and function-matching walks along the worst-case flat direction on 3BFF.}
\textbf{A, B}: task loss relative to its anchor value, and function-matching loss to the outputs of
the anchor, over $300$ null-space steps from four anchors (median, band from minimum to maximum).
Solid lines follow the flat direction along which the task loss rises fastest; the paler dashed
lines follow the CKA-steered heading and a random flat direction from one anchor. \textbf{(C)}: change
of the task loss per step along the task-loss walk, split into the first-order term set by the
gradient and the second-order term set by the curvature. \textbf{(D)}: number of negative Hessian
eigenvalues at each anchor for the two losses. The task-loss walk drifts because the gradient, not
the curvature, moves the loss, whereas the function-matching walk holds both losses in place.}
\label{fig:func-vs-task-3bff}
\end{figure}

We measured these effects on four independently trained 3BFF anchors (seeds $0$ to $3$ of
Appendix~\ref{app:3bff-setup}), using a dense Hessian and eigendecomposition in double precision
so that the sign of every eigenvalue is exact. Two walks start from each anchor and are identical
except for the loss that defines the flat subspace and is restored after every null-space step:
the task MSE or the function-matching loss. Both walks take $300$ null-space steps of the same
size along the flat direction in which the task loss rises fastest, the worst case for a
task-loss walk, with no loss ceiling so that the drift is visible
(Figure~\ref{fig:func-vs-task-3bff}). At every anchor the task-loss gradient is nonzero, between
$2\times10^{-3}$ and $1\times10^{-2}$ in norm, and a small part of it lies inside the flat subspace.
The task-loss Hessian has several hundred negative eigenvalues out of $4{,}611$ parameters, whereas
the Hessian of the function-matching loss has none. Along the walk, the first-order
drift of the task loss per step exceeds its second-order term by two to three orders of magnitude,
and the task-loss walk multiplies the task loss roughly tenfold. The function-matching walk keeps
the outputs pinned to the anchor and, as a by-product, holds the task loss within ten percent of
its anchor value.

On 3BFF the two losses are nevertheless close in practice, because the task target is the full
output trajectory and the trained anchor reaches it up to a small residual, so the task MSE is
nearly a function-matching loss to a fixed target. 

\paragraph{The full output constrains the network more tightly than a scalar loss.}
Both losses are enforced on a finite probe set, so the question is how well the constraint carries
over to inputs outside the probe. Many output vectors share the same loss value on an image, so a
network that keeps the anchor's loss on every probe image can still change its prediction on that
image, and it is unconstrained on unseen images. Matching the full output vector fixes one number
per class for every probe image rather than one number in total, so the probe constraint transfers
better beyond the probe.

We tested this on ViT-S with walks from the same anchor, steered by the same max-over-layers CKA
objective with the same solver settings as in Section~\ref{sec:prh}, on a probe of $512$ labeled
ImageNet validation images. The walks differ only in what they hold fixed: the full logits (the
function-matching loss), the per-image cross-entropy at its anchor value, or the mean cross-entropy
at its anchor value. The two cross-entropy variants are exact level sets and therefore share the
zero-gradient property of the previous paragraph. What they lack is constraint strength.

\begin{figure}[t]
\centering
\includegraphics[width=0.65\linewidth]{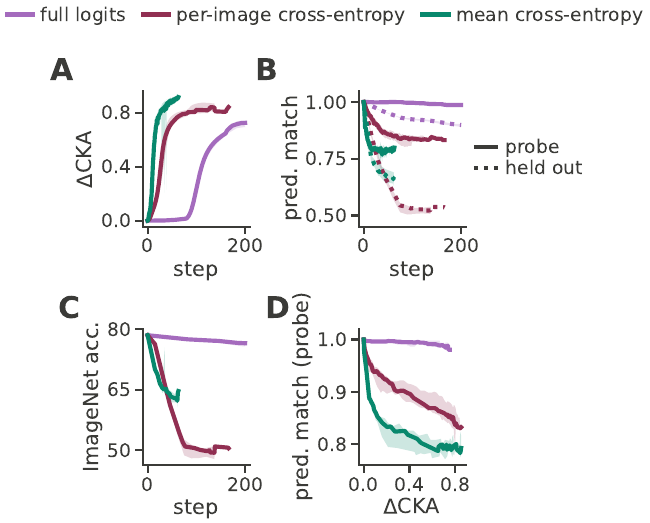}
\caption{\textbf{Constraint strength of three preserved quantities on ViT-S.}
CKA-steered walks from one anchor that hold fixed, on a probe of $512$ labeled ImageNet validation
images, the full logit vector, the per-image cross-entropy, or the mean cross-entropy (median over
probe seeds, band from minimum to maximum). \textbf{(A)}: representational divergence from the anchor
along the walk. \textbf{(B)}: fraction of images that keep the top-1 prediction of the anchor, on the
probe images (solid) and on $5{,}000$ held-out images (dotted). \textbf{(C)}: held-out top-1 accuracy.
\textbf{(D)}: probe prediction match against divergence, so the walks are compared at matched
representational change rather than at matched step. The fewer constraints a quantity imposes per
image, the more predictions change, on the probe and beyond it.}
\label{fig:func-vs-task-vit}
\end{figure}

In Figure~\ref{fig:func-vs-task-vit}, the logit walk keeps
almost every probe prediction and stays within a few points of the accuracy of the anchor, while
both cross-entropy walks lose predictions from the first steps and give up far more accuracy. They
lose predictions even on the probe images whose loss is pinned, because a walk can trade errors on
some probe images for gains on others at constant loss. The walks also diverge at very different
rates, so a given step means something different for each: the cross-entropy walks climb within a
few steps and the logit walk moves only later, which is why panel D compares them at matched
representational change instead. Off the probe the ordering of the two cross-entropy variants is
not stable, because the mean cross-entropy walk moves the network less per unit of representational
change and so carries fewer of its probe errors to unseen images.

To quantify the drift, Table~\ref{tab:constraint-count} counts the constraints each quantity
imposes and reports the probe prediction match at three values of divergence. The full logit vector
constrains a thousand numbers per image and holds almost every probe prediction at any divergence we
reach. The per-image cross-entropy imposes one constraint per image and the mean cross-entropy a
single constraint for the whole probe, and both lose probe predictions immediately and keep losing
them as the walk proceeds.

\begin{table}[H]
\centering
\caption{Preserved quantities on ViT-S, ordered by the number of constraints they impose on a probe of $512$
labeled ImageNet validation images over $1{,}000$ classes. Prediction match is the fraction of probe
images given the top-1 prediction of the anchor, at three values of representational divergence
(max over layers, mean and standard deviation over probe seeds).}
\label{tab:constraint-count}
\vspace{0.3cm}
\setlength{\arrayrulewidth}{1pt}
\begin{tabular}{@{}lcccc@{}}
\hline
& & \multicolumn{3}{c}{Pred. match (probe) with $\Delta\mathrm{CKA} =$} \\
\noalign{\vskip 0.05cm}
\cline{3-5}
\noalign{\vskip 0.05cm}
Preserved quantity & \# constraints & \quad$0.2$ & \quad$0.4$ & \quad$0.6$ \\[0.1cm]
\hline
\noalign{\global\arrayrulewidth=0.4pt \vskip 0.1cm}
Full logits & $512{,}000$ & \quad$0.996$ & \quad$0.994$ & \quad$0.993$ \\
Per-image cross-entropy & $512$ & \quad$0.916$ & \quad$0.894$ & \quad$0.875$ \\
Mean cross-entropy & $1$ & \quad$0.831$ & \quad$0.811$ & \quad$0.793$ \\
\hline
\end{tabular}
\end{table}

\paragraph{Function matching needs neither labels nor a task.}
The function-matching loss requires only the outputs of the anchor, so the same procedure runs on
unlabeled probes, on shifted inputs, on policies whose return is flat or non-differentiable
(Appendix~\ref{app:rl}), and on language models where the natural target is the next-token
distribution. Because any task loss is a function of the outputs, preserving the function preserves
every task loss at once, and the function-preserving set is a subset of the task-compatible set. 

Nonetheless, someone who wants the broader solution set for a given task can replace \eqref{eq:funcloss} by the task loss, as each caveat above can be mitigated. The first-order drift disappears if the task loss is
held at its anchor value through a squared residual, $\tfrac12\bigl(\mathcal{L}_{\mathrm{task}}(\vtheta)-\mathcal{L}_{\mathrm{task}}(\vtheta_0)\bigr)^2$,
which has zero gradient at the anchor by construction. Alternatively the null-space step can be projected orthogonal to the task-loss gradient, or the anchor can be trained closer to convergence before applying HNC. 
Additionally, the loose constraint can be tightened by constraining more probe inputs,  constraining the per-input losses
rather than their mean, or adding the decision on each input as a further target. Each of these
moves the constraint toward the full output, and in the limit recovers \eqref{eq:funcloss}.

\clearpage

\section{Combining HNC with global search for alternative solutions}
\label{app:search}

HNC starts from a single trained network and explores the connected,
function-preserving region around it, so it reaches solutions in the local region.
Retraining is the opposite kind of search: a fresh run from a new initialization can land
anywhere in weight space, but where it lands is not under our control, and independent runs tend
to converge on similar solutions. Here we give a proof of principle that the two can be
combined. Starting from the anchor, we alternate two stages. A DSA-steered HNC walk first moves
the current network along its loss level set while maximizing the summed dynamical distance to
every network found so far. A fresh network is then trained from scratch with the same summed
distance as a repulsive regularizer, which asks training to land far from all of them. Each
stage hands its endpoint to the next as the new anchor or the new set of networks to repel.

\begin{figure}[H]
\centering
\includegraphics[width=\textwidth]{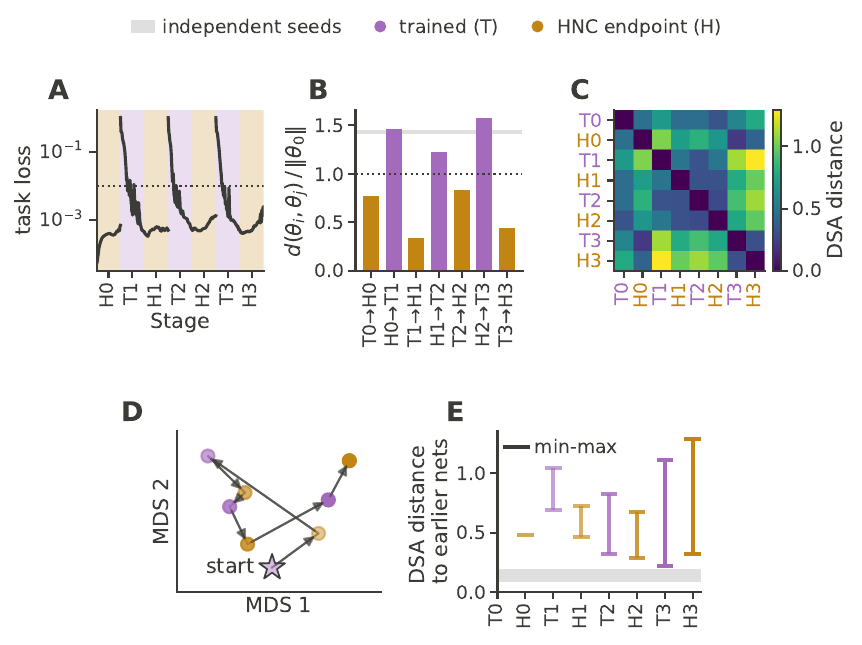}
\caption{\textbf{Alternating retraining with the DSA-steered HNC walk yields a chain of
dynamically distinct 3BFF solutions at preserved function.} Starting from the anchor $T_0$, each
round first runs a DSA-steered walk from the current trained network ($H_k$), then trains a fresh
network with a penalty on dynamical similarity to every network found so far ($T_{k+1}$).
\textbf{(A)} Held-out task loss along the chain (purple: retraining; orange: HNC walk; dotted line:
loss ceiling). Every trained network reaches the anchor's accuracy and every walk stays well below
the ceiling. \textbf{(B)} Weight distance between consecutive networks, in units of the anchor's
weight norm; the grey band is the range of distances between independently trained seeds.
Retraining lands as far away as an unrelated seed, whereas a walk moves less than one anchor norm.
\textbf{(C)} Pairwise DSA distance between the eight networks. \textbf{(D)} Two-dimensional embedding
of the same distances; the star is the anchor and the arrows follow the chain. \textbf{(E)} For each
new network, its DSA distance to the nearest and farthest earlier network, against the range between
independently trained seeds (grey band). Every network in the chain lies farther from all its
predecessors than two seeds lie from each other.}
\label{fig:chain_overview}
\end{figure}

\begin{figure}[H]
\centering
\includegraphics[width=\textwidth]{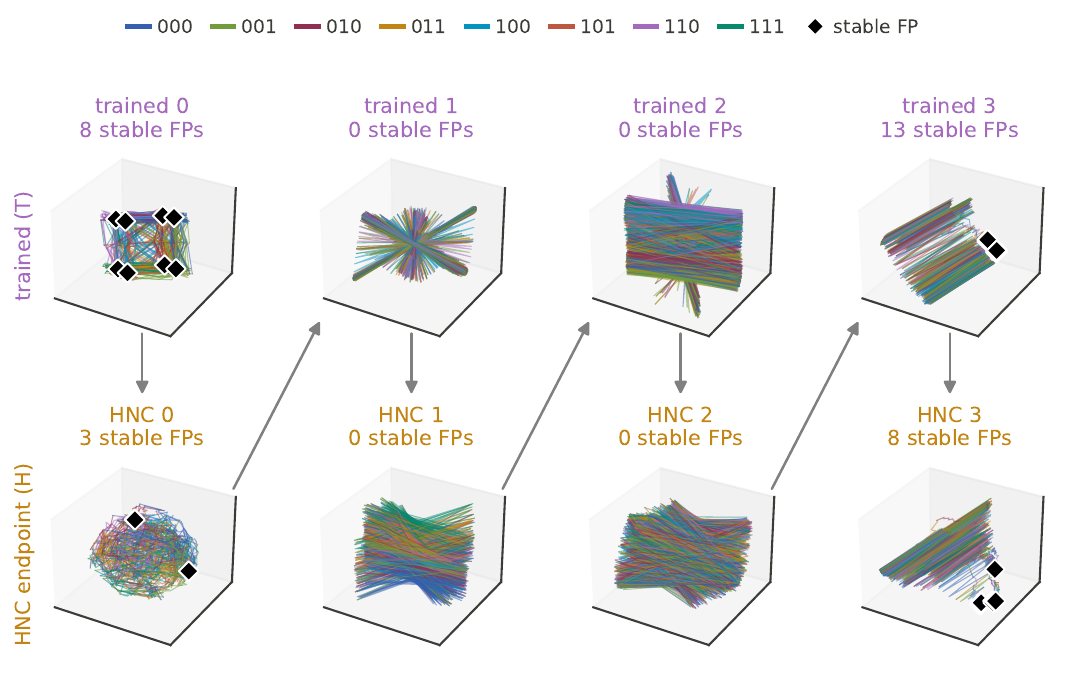}
\caption{\textbf{The chain visits qualitatively different dynamical mechanisms.} Hidden-state
trajectories of every network in the chain on the same trials, projected onto each network's top
three principal components and coloured by memory state, with stable fixed points as black diamonds.
Top row: trained networks; bottom row: the HNC endpoints reached from the network above; grey arrows
give the order of the chain. The anchor stores each memory state at its own fixed point. The first
walk keeps only a few fixed points and spreads the activity into a cloud, the two retrained networks
that follow hold the memory with no stable fixed point at all, and the last round returns to a
fixed-point solution with more fixed points than memory states. All eight networks solve the task
with bit accuracy above $0.999$.}
\label{fig:chain_gallery}
\end{figure}

\begin{figure}[H]
\centering
\includegraphics[width=\textwidth]{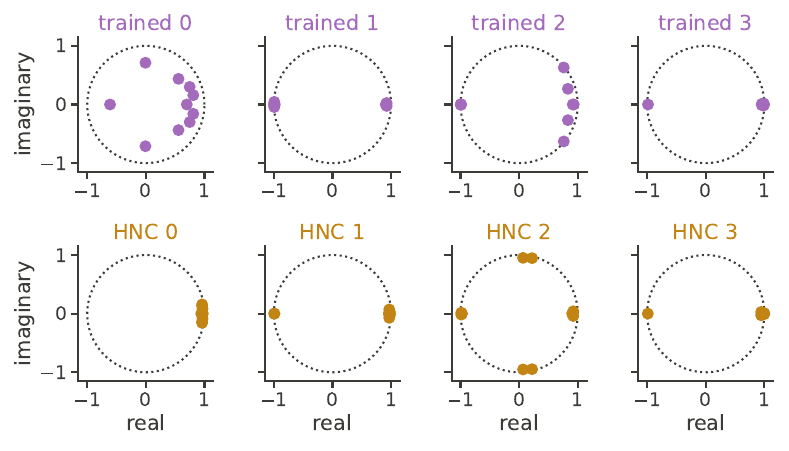}
\caption{\textbf{Eigenvalue spectra of the recurrent dynamics along the chain.} Eigenvalues of
a linear model fitted to each network's hidden-state trajectories, with the unit circle for
reference; the distance between these spectra is the quantity that both the retraining penalty and
the steered walk push apart. The anchor's eigenvalues fan out from the real axis. The networks
without fixed points concentrate their eigenvalues at $+1$ and $-1$, a slowly decaying mode and a
mode that alternates every step, and the third walk endpoint adds a pair of modes that repeat every
four steps.}
\label{fig:chain_spectra}
\end{figure}

The chain travels farther in weight space than either method alone, and every network in it
performs the task as well as the anchor (Figure~\ref{fig:chain_overview}). Each retraining moves
about as far as an unrelated seed, and each walk then moves a further fraction of an anchor
norm without leaving its loss valley. More importantly, the chain keeps finding new
mechanisms: attractor-based memories, memories held by ongoing drift with no stable fixed
point, and a solution with more attractors than memory states
(Figure~\ref{fig:chain_gallery}), with correspondingly different dynamical spectra
(Figure~\ref{fig:chain_spectra}). Every new member lies farther from all of its predecessors
than two independently trained seeds lie from each other, so alternating global and local
search covers more of the solution space than repeated retraining would.

There are other ways to give HNC a more global reach. For example, we can loosen the loss ceiling during the walk and filter the visited networks by task loss afterwards, so the walk could then cross shallow ridges and basin boundaries. The walk could also be restarted from a perturbed copy
of its endpoint, a small random kick followed by a few restore steps, to leave a
flat sheet whose boundary it has reached. The same regularizer could also repel the alternative networks from solutions found by different architectures or optimizers,
and the walk could be steered toward a target network rather than away from the anchor, as in
\autoref{app:prh_vits_vitb}, which would let it bridge two independently found solutions
through function-preserving intermediates. We leave these directions to future work.

\clearpage

\section{Details for the 3BFF RNN experiments}
\label{app:rnn}

\subsection{Architecture and training of the 3BFF RNN}
\label{app:3bff-setup}

\paragraph{Task} In the 3-bit flip-flop task, each RNN receives N=3 independent input channels taking values in
{-1, 0, +1}, which switch with probability $p_{\mathrm{flip}}=0.3$. The network has N=3 output channels that must
retain the most recent nonzero input on their respective channels. Trials are $T=100$ timesteps long, and each batch
contains $256$ trials. Since the three memories are binary and independent. The task has $2^3=8$
distinct memory states, which is the origin of the eight-fixed-point cube in Fig.~\ref{fig:manifold}.

\paragraph{Architecture} We use a vanilla $\tanh$ RNN with $N_h=64$ hidden units, whose update rule is
\begin{equation}
\vh_t=\tanh\!\bigl(\mW_{\mathrm{ih}}\vu_t+\vb_{\mathrm{ih}}+\mW_{\mathrm{hh}}\vh_{t-1}+\vb_{\mathrm{hh}}\bigr),
\qquad
\vy_t=\mW_{\mathrm{out}}\vh_t+\vb_{\mathrm{out}} .
\end{equation}
We use Kaiming uniform initialization for $\mW_{\mathrm{ih}}$ and $\mW_{\mathrm{out}}$, and orthogonal initialization for $\mW_{\mathrm{hh}}$. All biases are zero.
All biases are initialized to zero.

\paragraph{Training} Networks are trained by backpropagation through time over all $T=100$
timesteps on the Mean Squared Error (MSE) loss between the network's output and the target.
We use Adam with a constant learning rate of $10^{-3}$ with the gradient
clipped to a global norm of $1.0$. Training stops once the epoch loss falls below $5\times10^{-5}$ on two consecutive epochs.

\subsection{Size of the flat subspace under different thresholds}
\label{app:flat-threshold}

\begin{figure}[h]
\centering
\includegraphics[width=0.75\textwidth]{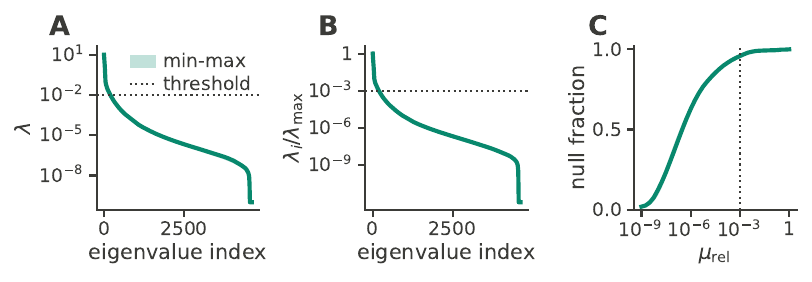}
\caption{\textbf{Hessian spectrum of the trained 3BFF RNNs.} \textbf{(A)} All $4611$
eigenvalues at the anchor, sorted. \textbf{(B)} The same eigenvalues divided by the largest one.
\textbf{(C)} Null fraction, the fraction of parameter directions with $\lambda_i<\mu_{\mathrm{rel}}\lambda_{\max}$,
as the threshold varies. Curves and bands are the median and range over five independently
trained anchors. Dotted lines mark the threshold of the undirected walks,
$\mu_{\mathrm{rel}}=10^{-3}$.}
\label{fig:3bff_gn_spectrum}
\end{figure}

The 3BFF RNN is small enough ($P=4611$) that we can form the full Hessian and
eigendecompose it directly, which gives the complete spectrum rather than the $k$ flat directions that
LOBPCG returns. Figure~\ref{fig:3bff_gn_spectrum} shows the spectra of the five trained networks.
A few directions carry almost all of the curvature: the effective dimensionality of the
spectrum, measured by its participation ratio, is $23.2\pm1.6$ out of $4611$ (mean $\pm$ s.d.\ over
five anchors). The remaining eigenvalues decay smoothly over nine orders of magnitude with no gap
between sharp and flat directions. The size of the
flat subspace therefore depends on where the threshold is placed
(Figure~\ref{fig:3bff_gn_spectrum}C, Table~\ref{tab:flat-threshold}). 

\begin{table}[h]
\centering
\caption{Dimension of the flat subspace of the 3BFF RNN ($P=4611$) as the relative flatness threshold
$\mu_{\mathrm{rel}}$ varies. A direction is counted as flat when its eigenvalue satisfies
$\lambda<\mu_{\mathrm{rel}}\lambda_1$, where $\lambda_1$ is the top eigenvalue of the Hessian (median $10.8$ across seeds).}
\label{tab:flat-threshold}
\vspace{0.3cm}
\setlength{\arrayrulewidth}{1pt}
\begin{tabular}{ccccc}
\hline
$\mu_{\mathrm{rel}}$ & absolute flatness threshold& median dimension & fraction of $P$ & $\max\Delta\mathcal{L}$ \\[0.1cm]
\hline
\noalign{\global\arrayrulewidth=0.4pt \vskip 0.1cm}
$1.0\times10^{-7}$ & $1.1\times10^{-6}$ & 1701 & 36.9\% & $5.4\times10^{-7}$ \\
$1.0\times10^{-6}$ & $1.1\times10^{-5}$ & 2896 & 62.8\% & $5.4\times10^{-6}$ \\
$1.0\times10^{-5}$ & $1.1\times10^{-4}$ & 3684 & 79.9\% & $5.4\times10^{-5}$ \\
$1.0\times10^{-4}$ & $1.1\times10^{-3}$ & 4145 & 89.9\% & $5.4\times10^{-4}$ \\
$1.0\times10^{-3}$ & $1.1\times10^{-2}$ & 4418 & 95.8\% & $5.4\times10^{-3}$ \\
$1.0\times10^{-2}$ & $1.1\times10^{-1}$ & 4565 & 99.0\% & $5.4\times10^{-2}$ \\
\hline
\end{tabular}
\end{table}

\subsection{Hyperparameters of the null-space walks}
\label{app:walk-settings}

Tables~\ref{tab:und-settings} to~\ref{tab:dsa-settings} list the settings of the walks in
Fig.~\ref{fig:manifold}. Every walk starts from one of five independently trained anchors
(seeds $0$ to $4$ of \autoref{app:3bff-setup}), with three replicates per arm and anchor. The
task MSE is measured on a fixed batch of $128$ trials. After each null-space step and its
restore steps, the step is accepted if this loss is below the loss ceiling and rejected
otherwise. A rejected step is undone and the step size halved; after an accepted step the step
size grows again. The undirected walks follow an explicit basis of the flattest eigenvectors,
which they recompute on a fixed schedule and after every rejection. The steered walks never
form a basis. They project the gradient of the steering objective with the soft projection of
Section~\ref{sec:method}, solved at the current weights at every step, so the flat directions
are refreshed automatically and no separate relinearization is needed.

\begin{table}[h]
\centering
\caption{Settings of the undirected walks in Fig.~\ref{fig:manifold}.}
\label{tab:und-settings}
\vspace{0.3cm}
\setlength{\arrayrulewidth}{1pt}
\begin{tabular}{@{}lp{8.2cm}@{}}
\hline
Hyperparameter & Value \\[0.1cm]
\hline
\noalign{\global\arrayrulewidth=0.4pt \vskip 0.1cm}
Flat directions & explicit basis of eigenvectors with $\lambda<10^{-3}\lambda_1$\\
Step direction & one random heading direction uniformly sampled from $\Vzero$, held fixed and re-projected onto the current basis at every relinearization\\
Restore steps & $12$ per null-space step \\
Null-space step size $\eta$ & starts at $1.0$, halved on rejection and grown $1.1\times$ on acceptance of a step;  lower bounded by $10^{-4}$ and upper bounded by $4.0$\\
Loss ceiling& $10^{-2}$ \\
Relinearization & every $20$ accepted steps or on rejection, at most $60$ times \\
Walk length & $600$ steps \\
\hline
\end{tabular}
\end{table}

\begin{table}[h]
\centering
\caption{Settings of the CKA-steered walks in Fig.~\ref{fig:manifold}.}
\label{tab:cka-settings}
\vspace{0.3cm}
\setlength{\arrayrulewidth}{1pt}
\begin{tabular}{@{}lp{8.2cm}@{}}
\hline
Hyperparameter & Value \\[0.1cm]
\hline
\noalign{\global\arrayrulewidth=0.4pt \vskip 0.1cm}
Flat directions & soft projection with damping $\mu=10^{-7}\lambda_1$\\
Step direction & gradient of the steering objective, after a seeded random first step \\
Steering objective & $1-\mathrm{CKA}$ between the hidden states of the current network and the anchor, on a probe of $32$ trials \\
Solver & plain CG, $200$ iterations, started from zero at every step (Appendix~\ref{app:solver}) \\
Restore steps & $12$ per null-space step \\
Null-space step size $\eta$ & starts at $0.03$; halved on rejection down to $5\times10^{-4}$ \\
Loss ceiling & $10^{-2}$ \\
Relinearization & at every step\\
Walk length & $600$ steps \\
\hline
\end{tabular}
\end{table}

\begin{table}[h]
\centering
\caption{Settings of the DSA-steered walks in Fig.~\ref{fig:manifold}.}
\label{tab:dsa-settings}
\vspace{0.3cm}
\setlength{\arrayrulewidth}{1pt}
\begin{tabular}{@{}lp{8.2cm}@{}}
\hline
Hyperparameter & Value \\[0.1cm]
\hline
\noalign{\global\arrayrulewidth=0.4pt \vskip 0.1cm}
Flat directions & soft projection with damping $\mu=10^{-7}\lambda_1$\\
Step direction & gradient of the steering objective, after a seeded random first step \\
Steering objective & DSA distance between the hidden-state dynamics of the current network and the anchor ($5$ delays, rank $10$), on a probe of $32$ trials \\
Solver & plain CG, $200$ iterations, started from zero at every step (Appendix~\ref{app:solver}) \\
Restore steps & $12$ per null-space step \\
Null-space step size $\eta$ & starts at $0.03$; halved on rejection down to $5\times10^{-4}$ \\
Loss ceiling & $10^{-2}$ \\
Relinearization & at every step\\
Walk length & $2{,}000$ steps \\
\hline
\end{tabular}
\end{table}

\subsection{A set of diverse 3BFF solutions}
\label{app:3bff-gallery}
Here, we show a set of diverse solutions that all complete the task with 100\% accuracy whose task MSE lies well below 0.05 over an example CKA-steered, DSA-steered, and undirected null-space walk.
\begin{figure}[H]
\centering
\includegraphics[width=\textwidth]{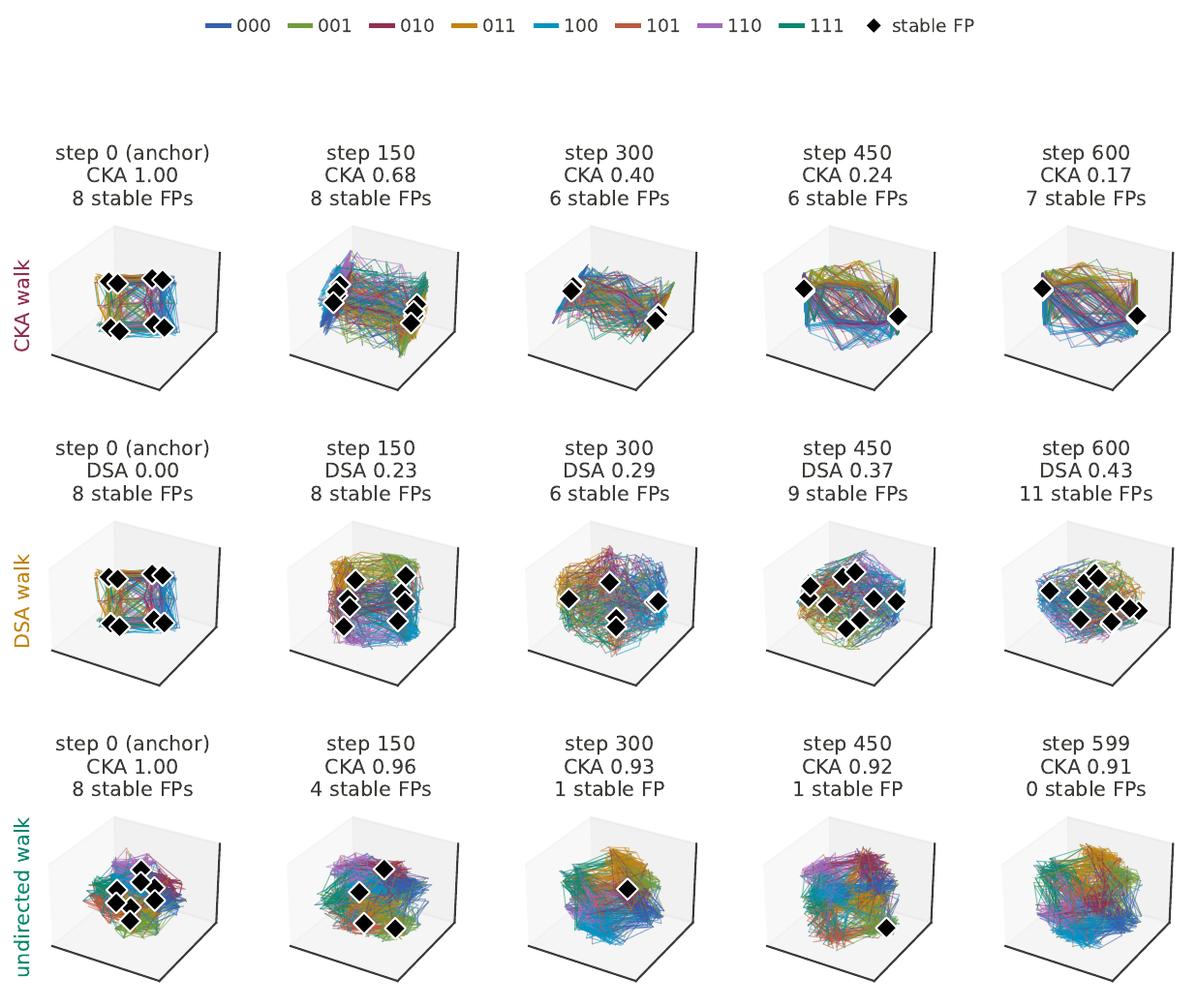}
\caption{\textbf{A set of diverse 3BFF solutions produced by the function-preserving null-space walks.}
Hidden-state trajectories on the probe trials, projected onto each snapshot's own top three
activity PCs and coloured by the current memory state, with the snapshot's stable zero-input
fixed points as black diamonds. }
\label{fig:3bff_gallery}
\end{figure}

\subsection{MDS embeddings of the solutions reachable from ten anchors}
\label{app:3bff-mds}
Fig.~\ref{fig:manifold}F embeds the solutions reachable from one anchor, here we aggregate all networks reached during the HNC walks starting from 10 anchors to highlight the diversity of solutions that a task admits. We trained ten anchors with the training procedure of Appendix~\ref{app:3bff-setup}. From each of the ten anchors, we first ran 3 undirected, DSA-steered, and CKA-steered HNC walks, resulting in a total of 9 HNC walks, and saving evenly spaced checkpoints per walk. 
From five of the anchors, we additionally ran walks steered away from an earlier HNC endpoint instead of the anchor, as in Appendix~\ref{app:search}. 

\begin{figure}[H]
\centering
\includegraphics[width=0.8\linewidth]{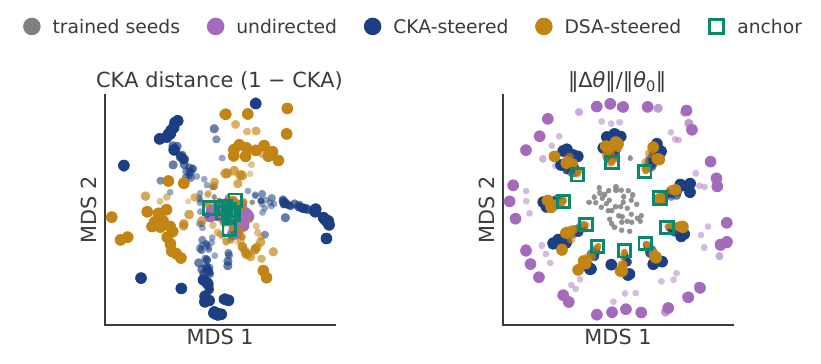}
\caption{\textbf{MDS embeddings of the solutions reachable from ten anchors, by representation and by weights.}
Metric MDS of 470 networks: ten independently trained anchors (hollow squares), four checkpoints from each of their undirected, CKA-steered, and DSA-steered walks, and forty further independently trained networks (grey). Walks steered away from an earlier endpoint rather than from the anchor share the colour of the metric they steer. Marker size and opacity grow along each walk. Left: representational distance ($1-\mathrm{CKA}$). Right: relative weight distance. In weights every anchor forms its own cluster with its walks around it. In representation all anchors and trained networks collapse onto one point, and the walks from every anchor spread far beyond it.}
\label{fig:3bff_mds_10anchors}
\end{figure}

For every pair of networks we computed the CKA distance between hidden states on a common set of probe trials and the Euclidean weight distance relative to the anchor norm, and embedded each distance matrix in two dimensions with metric MDS (Figure~\ref{fig:3bff_mds_10anchors}).

The structure of this solution clearly decouples at the representational level and at the weight level. In weight space the networks group by anchor. Each anchor sits at the centre of its own walks, and no walk comes closer to another anchor than to its own. In representation space, all ten anchors are close to each other, and the steered walks leave them in many directions, reaching distances several times the spread of independent training. The solutions reached by HNC spread across the embedding rather than collapsing into a tight cluster, so the 3BFF task admits far more solutions than independent training reveals.
\clearpage

\subsection{Function-preserving structured pruning}
\label{app:sparsity}

Beyond representational and dynamical dissimilarity, the steering potential can also target
the structure of the weights themselves. Here we steer the walk to maximize sparsity of the network weights
so to enable network pruning. In the RNN, a unit whose overall output weight magnitude is 
zero affects neither the recurrent dynamics nor the output, so it can be removed without affecting the network computation. 
Let unit $j$'s outgoing weight group be $\vg_j = [\,\mW_{\mathrm{hh}}[:,j]\,;\ \mW_{\mathrm{out}}[:,j]\,]$,
we want to steer the walk to maximize a differentiable count of the total number of removable units, 

\begin{equation}
\varphi_{\mathrm{prune}}(\vtheta) \;=\; \frac{1}{N_h}\sum_{j=1}^{N_h}
\exp\!\Bigl(-\tfrac{\lVert \vg_j\rVert^2}{2\sigma^2}\Bigr),
\end{equation}
where each unit contributes a Gaussian bump that equals one when its outgoing group is zero
and vanishes when the group is large. Because all outgoing weights of a unit share a single
bump, the gradient of $\varphi_{\mathrm{prune}}$ drives all output weights of a single unit toward zero together.

\begin{figure}[h]
\centering
\includegraphics[width=\textwidth]{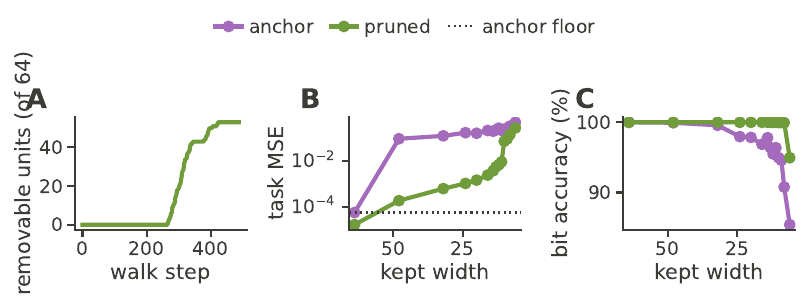}
\caption{\textbf{The prune-steered walk makes the network hard-prunable at preserved
function.} \textbf{(A)} Fraction of removable units (outgoing group norm below the dead
threshold) against the normalized weight movement, the walk arclength divided by the anchor
weight norm. \textbf{(B)} Hard-truncation test: the RNN is physically rebuilt keeping only
the units with the largest outgoing group norms; task MSE against the removed fraction, for
the anchor and the prune-steered endpoint. The dotted line marks the largest removed
fraction at which the pruned network is still bit-perfect ($53/64=0.83$, $11$ units kept).
\textbf{(C)} Bit accuracy of the same truncations: the pruned endpoint stays at $100\%$ down
to $11$ units, while the anchor is already imperfect at the first truncation and degrades
from there.}
\label{fig:prune}
\end{figure}

Starting from the seed-$0$ anchor, we apply HNC with an annealed $\sigma$ schedule, 
gradually decreasing $\sigma$ from the median norm of the anchor’s outgoing weight groups to one-tenth 
of that value. This causes the walk to initially shrink all outgoing weights uniformly, before increasingly 
targeting low-magnitude groups and driving them effectively to zero.
Throughout the walk, the fraction of removable units steadily increases (Figure~\ref{fig:prune}A). Physically
removing the units with near-zero outgoing weights prunes the $64$-unit RNN down to $11$
units, about $17\%$ of the original hidden size and $5\%$ of the original parameters. The
truncation leads to a small rise in the task MSE, yet the pruned network still performs the
3BFF task with $100\%$ accuracy (Figure~\ref{fig:prune}B,~C). This suggests that HNC can
directly find and exploit the unconstrained degrees of freedom in the network weights,
enabling network pruning.

\clearpage

\subsection{Steered solutions are SGD-stable}
\label{app:sgd_stable}
\begin{figure}[h]
\centering
\includegraphics[width=\textwidth]{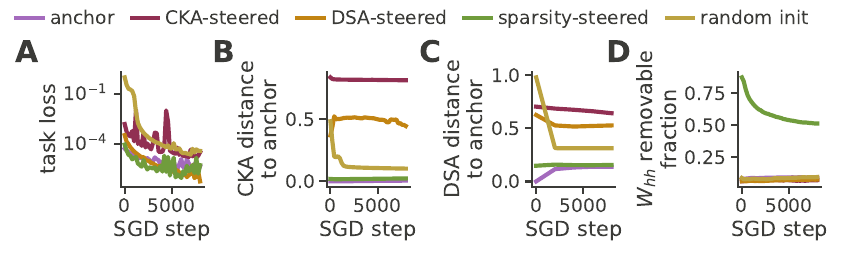}
\caption{\textbf{Resumed SGD maintains the steered divergence while finding a lower-loss solution.}
Plain SGD resumed from each steered 3BFF endpoint (and from the anchor and a random initialization),
tracking the task loss and the quantities the steering moved: representation (CKA distance to the
anchor), dynamics (DSA distance to the anchor), and structure (fraction of near-zero $W_{hh}$
columns).}
\label{fig:sgd_stable_traj}
\end{figure}

Next, we ask whether the solution originally found by gradient descent is privileged for the
3BFF task, such that retraining with SGD would converge back to it. We therefore resume SGD
on the 3BFF task for $8{,}000$ steps from each HNC endpoint and, as references, from the
anchor and from a fresh random initialization (Figure~\ref{fig:sgd_stable_traj}).

\begin{figure}[h]
\centering
\includegraphics[width=0.75\textwidth]{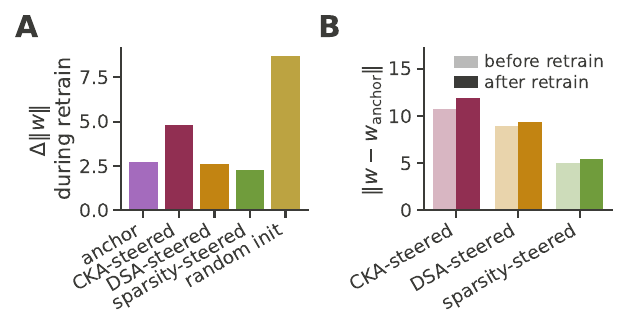}
\caption{\textbf{SGD stays local.} \textbf{(A)} How far the weights move during the resumed
training, from each start. \textbf{(B)} Distance of each steered solution to the anchor before
(light) and after (solid) retraining.}
\label{fig:sgd_basin}
\end{figure}

Upon retraining, the optimization settles into a solution in the vicinity of each steered
endpoint. The property that each walk was steered by (the CKA distance to the anchor, the
DSA distance to the anchor, or the fraction of removable $W_{hh}$ columns) stays where the
walk left it. Moreover, retraining moves the network weights only a small distance, far less
than the distance traveled from a random initialization. The retrained solutions come no closer to the anchor than the steered
endpoints they started from either, so retraining does not pull the endpoints back toward
the anchor in the weight space (Figure~\ref{fig:sgd_basin}).

\clearpage

\subsection{Out-of-distribution probes of the walk endpoints}
\label{app:3bff-ood}

\begin{figure}[h]
\centering
\includegraphics[width=0.65\textwidth]{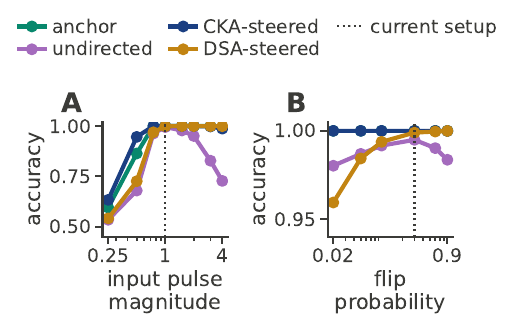}
\caption{\textbf{Out-of-distribution input probes of the walk endpoints.} Accuracy of the anchor
and the three walk endpoints when the input statistics differ from training. The dotted line marks
the trained value. \textbf{(A)} All input pulses scaled by a common magnitude. \textbf{(B)} Pulse
probability per step. Rare pulses leave long stretches without input within a trial.}
\label{fig:3bff_ood}
\end{figure}

The walk endpoints match the anchor on the trained input distribution, so we ask whether they
also match it on inputs that this distribution never contains. During training, each input bit
receives a unit pulse at every step with probability $0.3$. We therefore vary the pulse magnitude
and the pulse rate, and we evaluate the anchor and the three walk endpoints, all of which reach an
in-distribution accuracy of $0.99$ or above (Figure~\ref{fig:3bff_ood}). Memory retention and
recovery from hidden-state perturbations are shown in Fig.~\ref{fig:manifold}E. 

Weak pulses degrade the DSA-steered and undirected endpoints before the anchor. Strong pulses
degrade only the undirected endpoint, whose recurrent spectral radius has grown along the walk.
Rare pulses leave long stretches without input within a trial. There the DSA-steered endpoint
loses accuracy first, in line with its weaker memory retention in Fig.~\ref{fig:manifold}E.

\clearpage
\section{Additional experiments on ViT}
\label{app:prh}
\subsection{Smooth surrogate for the max-over-layers CKA}
\label{app:softmaxcka}

Section~\ref{sec:prh} steers the walk to lower the highest CKA between any layer of the anchor and
any layer of the current network. Let $c_{ij}(\vtheta)$ be the linear CKA between the CLS-token
features of layer $i$ of the anchor and layer $j$ of the current network on the $N=512$ probe
images, for $i,j=1,\dots,L$ with $L=13$ (one per transformer block plus the pre-logits features).
The hard maximum $\max_{i,j}c_{ij}$ is not smooth: its gradient comes from a single layer pair and
jumps whenever another pair takes over. The walk therefore ascends a smooth version,
\begin{equation}
\varphi_{\mathrm{CKA}}(\vtheta)\;=\;1-\frac{1}{\beta}\log\sum_{i,j=1}^{L}\exp\bigl(\beta\,c_{ij}(\vtheta)\bigr),
\qquad \beta=30 ,
\label{eq:softmaxcka}
\end{equation}
which replaces the maximum by a log-sum-exp. For large $\beta$ the log-sum-exp stays close to the
maximum. Its gradient is a weighted average of the gradients of all $c_{ij}$,
with most of the weight on the most similar layer pairs, so each step pushes down whichever pairs
are currently closest to the maximum. The $k$NN potential of \eqref{eq:softknn} uses the same
construction with the soft mutual-$k$NN alignment in place of $c_{ij}$. Throughout, we report the
exact hard maximum (Fig.~\ref{fig:prh}).

\subsection{Scoring the visual attributes of the most similar image pairs}
\label{app:visualsim}

Figure~\ref{fig:prh}E asks what the anchor and the endpoint have in common when they represent two
images similarly. For the anchor and the HNC endpoint network, we rank all possible pairs of probe images by the cosine similarity of
their penultimate-layer features and keep the five highest-ranked pairs. Every pair of probe images is then scored on three visual attributes that are independent of the class identity of each image:

\begin{itemize}\setlength{\itemsep}{0pt}
  \item \emph{Semantic content.} Cosine similarity between the class-token features of a DINOv2
  ViT-S/14 \citep{oquab2023dinov2}. This model is trained without labels, so it cannot inherit
  the anchor's supervision.
  \item \emph{Color.} Cosine similarity between $8\times8\times8$ HSV color histograms. Each
  pixel is described by three numbers: its hue, saturation, and value. Each of the three axes is divided into eight bins, giving $512$ bins in total, and we count how many of the image's pixels fall into each one and normalize the sum of counts to one. The histogram records only what colors an image contains and in what
  proportions, and is unchanged if the pixels are rearranged. 
  \item \emph{Spatial layout.} Correlation between the images downsampled to $32\times32$
  grayscale, which keeps the coarse arrangement of light and dark regions and discards color and
  fine detail.
\end{itemize}

The three scores are on different scales, so we report each one as a percentile within the
distribution of that score over all pairs of probe images. A pair drawn at random sits at the
$50$th percentile by construction, which sets the dashed reference line, and a high percentile
means the network's closest pairs share that attribute more than two arbitrary images do. The
plotted value is the mean percentile over the five pairs.

\subsection{Steering by mutual $k$-nearest neighbor}
\label{app:prh_knn}

\begin{figure}[h]
\centering
\includegraphics[width=\textwidth]{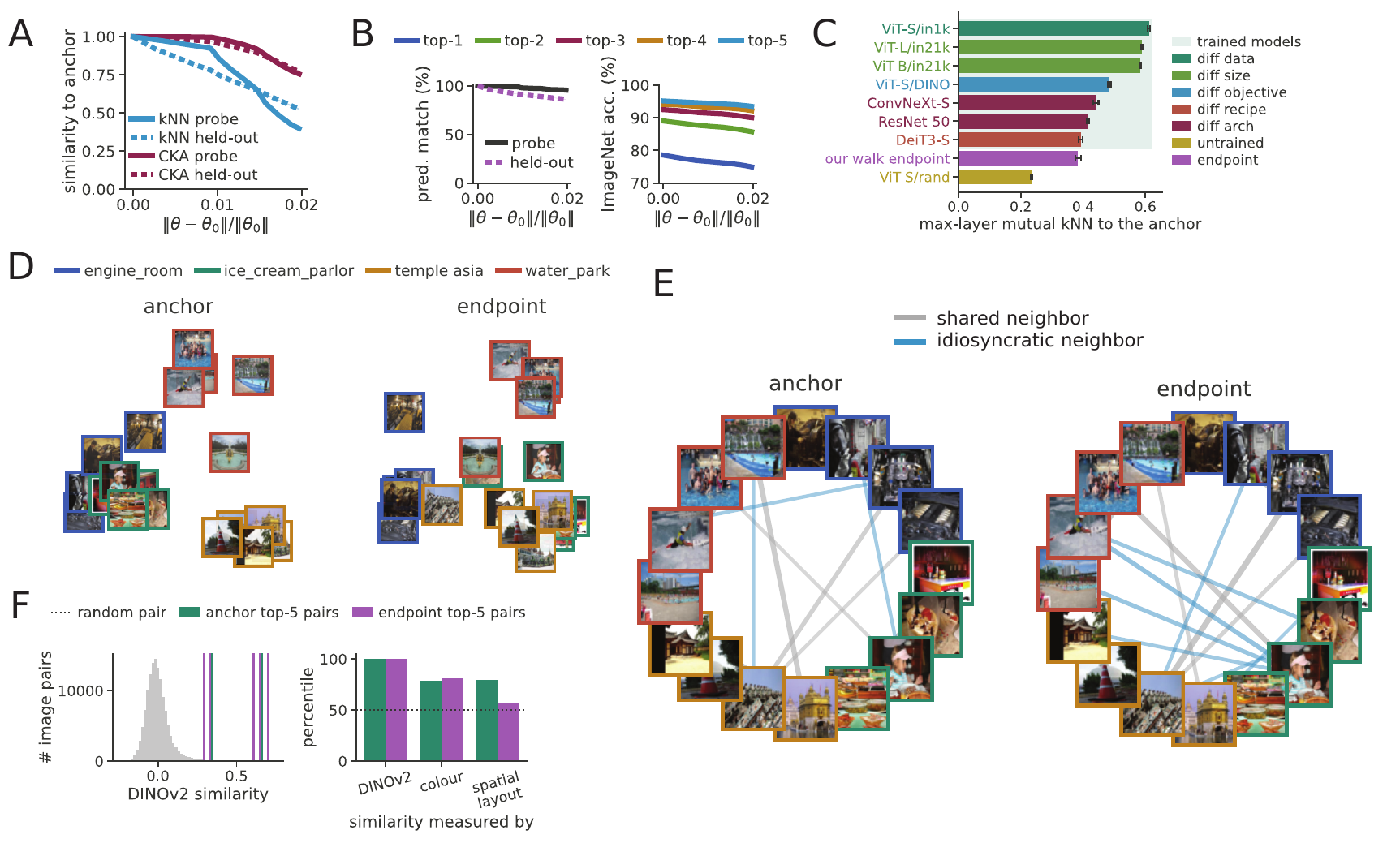}
\caption{\textbf{Steering the mutual $k$NN score finds alternative representations at preserved input-output mapping.}
\textbf{(A)} The max-over-layers mutual $k$NN similarity to the anchor falls along the walk, both on the probe images and on held-out images never seen during HNC. The max-over-layers CKA falls as well, even though the walk does not steer it.
\textbf{(B)} The top-$k$ predictions on Places-365 images and the top-$k$ accuracy on ImageNet are largely preserved along the walk.
\textbf{(C)} At the endpoint, the max-layer mutual $k$NN to the anchor sits below every independently trained model and above only an untrained initialization.
\textbf{(D)} MDS embedding of 20 images from four classes, for the anchor and the endpoint.
\textbf{(E)} Nearest-neighbor graph of the same images. Grey edges mark neighbors shared by the anchor and the endpoint; blue edges mark neighbors found in only one of the two networks.
\textbf{(F)} The most similar image pairs of the endpoint match those of the anchor in semantic content (DINOv2) and color, but their spatial-layout similarity falls toward chance.}
\label{fig:vit_knn}
\end{figure}

Section~\ref{sec:prh} steers the linear CKA. Here we repeat the experiment with the other
similarity score used by the PRH, the mutual $k$-nearest-neighbor alignment of
\citet{huh2024platonic} (Figure~\ref{fig:vit_knn}). The anchor is the same ImageNet-trained
ViT-S/16, the function is held fixed on $256$ Places-365 images, and the score follows the PRH
exactly: for each image, the fraction of its $k=10$ nearest neighbors that the two
representations share, averaged over images and taken at its maximum over all layer pairs.
Nearest-neighbor sets change discontinuously as the weights move, so they cannot be steered
directly. We therefore relax them: image $b$ counts as a neighbor of image $a$ with a weight that
decays smoothly with how far $b$ lies beyond the $k$-th nearest neighbor of $a$. Writing
$d^{(l)}_{ab}$ for the squared distance between the layer-$l$ features of $a$ and $b$, and
$r^{(l)}_{a}$ for the squared distance from $a$ to its $k$-th nearest neighbor, the soft
neighborhood at layer $l$ is $S^{(l)}_{ab}(\vtheta)=\sigma\big((r^{(l)}_{a}-d^{(l)}_{ab})/w_l\big)$,
where $\sigma$ is the logistic sigmoid, $r^{(l)}_{a}$ is held constant when differentiating, and
the width $w_l$ is set once from the anchor at $5\%$ of the median $r^{(l)}_{a}$ over the probe.
The steering potential is the soft dissimilarity
\begin{equation}
\varphi_{\mathrm{kNN}}(\vtheta)\;=\;1-\frac{1}{\beta}\log\sum_{i,j=1}^{L}\exp\big(\beta\,m_{ij}(\vtheta)\big),
\qquad
m_{ij}(\vtheta)\;=\;\frac{1}{Nk}\sum_{a\neq b}\bar S^{(i)}_{ab}\,S^{(j)}_{ab}(\vtheta),
\label{eq:softknn}
\end{equation}
where $m_{ij}$ measures how much of the anchor's neighborhood structure at layer $i$ (the fixed
set $\bar S^{(i)}$, computed at $\vtheta_0$) survives in layer $j$ of the current network,
$N=256$ is the probe size, $L=13$ counts the class-token layers (one per block plus the
pre-logits feature), and the log-sum-exp with $\beta=30$ is a smooth stand-in for the maximum,
as in \eqref{eq:softmaxcka}. Raising $\varphi_{\mathrm{kNN}}$ therefore pushes down the alignment
at whichever layer pair is currently the most similar. We report the exact hard score throughout.

\clearpage
\subsection{Hyperparameters of the ViT walks}
\label{app:vit-walk-settings}

\begin{table}[h]
\centering
\caption{Settings of the CKA-steered ViT-S walk in Fig.~\ref{fig:prh}.}
\label{tab:vit-settings}
\vspace{0.3cm}
\setlength{\arrayrulewidth}{1pt}
\begin{tabular}{@{}lp{8.2cm}@{}}
\hline
Hyperparameter & Value \\[0.1cm]
\hline
\noalign{\global\arrayrulewidth=0.4pt \vskip 0.1cm}
Anchor & ImageNet-trained ViT-S/16, $22$M parameters\\
Probe set & $512$ Places-365 images\\
Flat directions & soft projection with damping $\mu=10^{-7}\lambda_1$ \\
Solver & preconditioned conjugate gradient, $40$ iterations\\
Steering objective & $\varphi_{\mathrm{CKA}}$ of \eqref{eq:softmaxcka}, $\beta=30$ \\
First step & random direction, soft-projected \\
Restore steps & $6$ per null-space step \\
Null-space step size $\eta$ & starts at $0.05$; halved on rejection down to $10^{-3}$ \\
Loss ceiling & $10^{-2}$ \\
Relinearization & at every step \\
Walk length & up to $336$ steps\\
\hline
\end{tabular}
\end{table}

\begin{table}[h]
\centering
\caption{Settings of the mutual-$k$NN-steered ViT-S walk in Figure~\ref{fig:vit_knn}.}
\label{tab:vit-knn-settings}
\vspace{0.3cm}
\setlength{\arrayrulewidth}{1pt}
\begin{tabular}{@{}lp{8.2cm}@{}}
\hline
Hyperparameter & Value \\[0.1cm]
\hline
\noalign{\global\arrayrulewidth=0.4pt \vskip 0.1cm}
Anchor & ImageNet-trained ViT-S/16, $22$M parameters\\
Probe set & $512$ Places-365 images\\
Flat directions & soft projection with damping $\mu=10^{-7}\lambda_1$ \\
Solver & preconditioned conjugate gradient, $40$ iterations\\
Steering objective & $\varphi_{\mathrm{kNN}}$ of \eqref{eq:softknn}, $\beta=30$, $k=10$ \\
First step & random direction, soft-projected \\
Restore steps & $8$ per null-space step \\
Null-space step size $\eta$ & starts at $0.05$; halved on rejection down to $10^{-3}$ \\
Loss ceiling & $10^{-2}$ \\
Relinearization & at every step \\
Walk length & $500$ steps\\
\hline
\end{tabular}
\end{table}

\clearpage
\subsection{Functional preservation beyond the probe set for CKA-steered HNC}
\label{app:prh_cka_function}

The CKA-steered walks in Figure~\ref{fig:prh} constrain the function on only $512$ Places-365 images. Although the endpoints match the anchor on these probe images, this constraint does not guarantee that they will preserve its behavior on unseen inputs. We therefore ask how well the endpoints retain the anchor's function beyond the probe set, including under distribution shift.

We evaluate the anchor and the endpoints of the five CKA-steered walks on held-out ImageNet validation images and on three distribution-shifted test sets with the ImageNet label space: ImageNet-V2 \citep{recht2019imagenet}, ImageNet-Rendition \citep{hendrycks2021many}, and ImageNet-Sketch \citep{wang2019learning}. For each dataset, we report three complementary measures. Top-$k$ accuracy assesses whether the endpoints retain the anchor's overall task performance. Prediction match measures the fraction of images on which an endpoint makes exactly the same top-1 prediction as the anchor. Error consistency \citep{geirhos2020beyond} measures chance-corrected agreement in which images the two networks classify correctly; unlike prediction match, it does not require the networks to select the same class when both are wrong.

We calibrate these measures against two reference networks. ViT-S/in1k has the same architecture as the anchor but is trained only on ImageNet-1k, providing a reference for a separately trained model. The diff-seed networks provide a stricter comparison: they repeat the anchor's final ImageNet-1k fine-tuning stage from the same ImageNet-21k checkpoint, changing only the random seed.

\begin{figure}[h]
\centering
\includegraphics[width=0.7\textwidth]{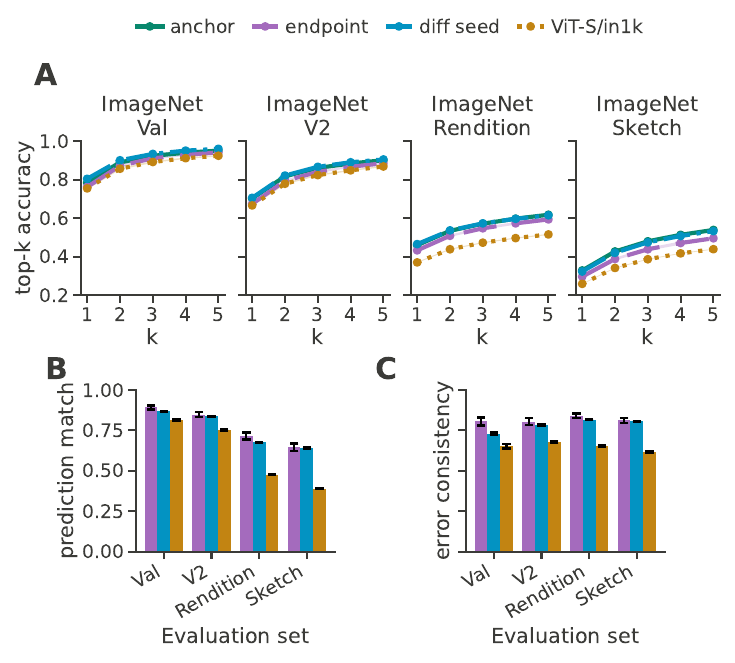}
\caption{\textbf{Accuracy, prediction match and error consistency of the CKA-steered endpoints on
four held-out test sets.} \textbf{(A)} Top-$k$ accuracy of the anchor, the walk endpoints, a diff-seed network (the anchor's
final fine-tuning stage repeated with a different seed) and an independently trained ViT-S/16 (ViT-S/in1k) on clean
ImageNet validation images and on ImageNet-V2, ImageNet-Rendition and ImageNet-Sketch. None of these
images constrained the walk. \textbf{(B)} Fraction of images on which a network makes the same top-1
prediction as the anchor. \textbf{(C)} Error consistency with the anchor (chance-corrected agreement on
which images are classified correctly). Endpoint curves and bars show the mean and s.d.\ over the five
walks of Figure~\ref{fig:prh}; diff-seed curves and bars the mean and s.d.\ over two seeds. The anchor and ViT-S/in1k are single networks; their bands and bars show the s.d.\ over
bootstrap resamples of the test images.}
\label{fig:cka_offprobe}
\end{figure}

The CKA-steered endpoints closely track the anchor's top-$k$ accuracy on all four datasets, with a small but consistent accuracy deficit across values of $k$ (Figure~\ref{fig:cka_offprobe}A). Their exact prediction agreement is highest on clean ImageNet and ImageNet-V2 and decreases on Rendition and Sketch (Figure~\ref{fig:cka_offprobe}B). This decline is expected in part because, when both networks are wrong, prediction match counts them as agreeing only if they select the same incorrect class. Error consistency remains near $0.8$ across all four datasets (Figure~\ref{fig:cka_offprobe}C), indicating that the endpoints continue to succeed and fail on largely the same images as the anchor even as exact prediction agreement decreases. Thus, their similarity to the anchor extends beyond aggregate accuracy to the image-level structure of their errors, although the agreement is not perfect.

The reference networks clarify the strength of this preservation. The endpoints exhibit higher prediction match and error consistency than ViT-S/in1k on every dataset, with the largest difference on Rendition and Sketch. More strikingly, their image-level agreement with the anchor is comparable to, and generally higher than, that of the diff-seed networks, despite their representations having moved much farther from the anchor. The CKA-steered walks therefore produce substantial representational change while retaining the anchor's off-probe behavior at least as well as repeating its final training stage. At the same time, the remaining accuracy loss and prediction disagreement show that this preservation is approximate rather than exact.

\needlines{22}
\begin{wrapfigure}{r}{0.5\textwidth}
\vspace{-4mm}
\centering
\includegraphics[width=\linewidth]{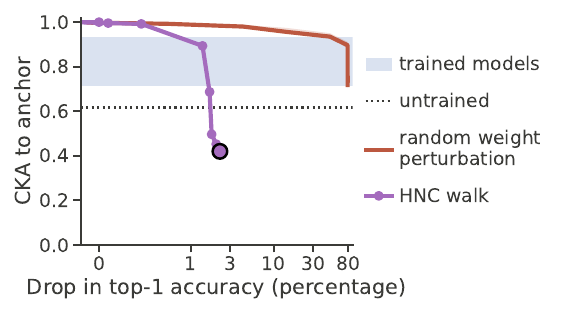}
\caption{\textbf{CKA to the anchor against accuracy lost, for random weight noise and for the HNC
walk.} Max-over-layers CKA against the drop in top-1 accuracy on held-out ImageNet validation images,
for Gaussian weight noise of increasing scale (three seeds) and for the walk checkpoints up to the
reported endpoint (ringed). Band: independently trained models; dotted: untrained.}
\label{fig:cka_damage}
\vspace{-3mm}
\end{wrapfigure}

A possible alternative explanation is that the low CKA simply reflects damage to the network associated with its modest accuracy loss. We test this by adding Gaussian noise to the anchor's weights at increasing scales and comparing the resulting accuracy loss and CKA with those observed along HNC (Figure~\ref{fig:cka_damage}). Random perturbations matched to the endpoint's accuracy loss leave CKA above $0.98$, and even perturbations that reduce accuracy to chance retain substantially higher CKA than the HNC endpoint. By contrast, the CKA-steered walk falls below the untrained-network reference after losing only less than 1\% accuracy. The representational displacement produced by HNC therefore cannot be explained as a generic consequence of degraded performance.

\clearpage

\subsection{Functional preservation beyond the probe set for mutual-$k$NN-steered HNC}
\label{app:prh_knn_function}

We next ask whether the representational result persists when HNC is steered with mutual-$k$NN similarity, which directly measures the preservation of local neighborhoods. Figure~\ref{fig:vit_knn} shows that mutual-$k$NN similarity falls on both probe and held-out images and, at the endpoint, lies below that of every independently trained model in the comparison. We evaluate the functional preservation of these endpoints using the same datasets, measures, and reference networks as for the CKA-steered walks.

\begin{figure}[h]
\centering
\includegraphics[width=0.7\textwidth]{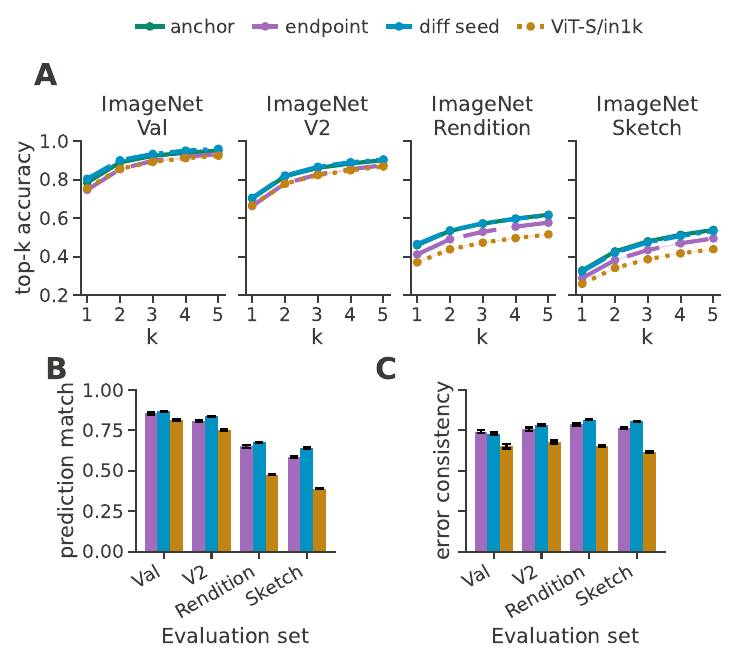}
\caption{\textbf{Accuracy, prediction match and error consistency of the $k$NN-steered endpoints on
four held-out test sets.} Same panels as Figure~\ref{fig:cka_offprobe}. \textbf{(A)} Top-$k$
accuracy of the anchor, the walk endpoints, the diff-seed network and ViT-S/in1k. \textbf{(B)} Fraction of images on
which a network makes the same top-1 prediction as the anchor. \textbf{(C)} Error consistency with the
anchor (chance-corrected). Endpoint curves and bars show the mean and s.d.\ over three $k$NN-steered walks with different
probe draws; diff-seed curves and bars the mean and s.d.\ over two seeds; anchor and ViT-S/in1k bands and bars
the s.d.\ over bootstrap resamples of the test images.}
\label{fig:knn_offprobe}
\end{figure}

The mutual-$k$NN-steered endpoints retain the overall shape of the anchor's top-$k$ accuracy curves, but their accuracy deficit is larger than for the CKA-steered endpoints (Figure~\ref{fig:knn_offprobe}A). Exact prediction agreement remains substantial, falling from approximately $0.85$ on clean ImageNet to approximately $0.6$ on ImageNet-Sketch (Figure~\ref{fig:knn_offprobe}B). Error consistency remains around $0.75$ across datasets (Figure~\ref{fig:knn_offprobe}C), showing that the endpoints still tend to make errors on the same images as the anchor even when they choose different output classes.

The endpoints agree with the anchor more strongly than ViT-S/in1k under both prediction match and error consistency. They nevertheless fall below the diff-seed networks on all four datasets, particularly under distribution shift. The mutual-$k$NN-steered walks therefore recover the central representational result using a complementary metric, but at this operating point they preserve the anchor's off-probe behavior less faithfully than either the diff-seed networks or the CKA-steered walks. Together, the two experiments show that HNC can move representations beyond the variation observed across trained models while retaining substantial task behavior, but that the degree of functional preservation depends on the walk configuration and steering objective.

This comparison reflects the walk configuration tested here as much as the choice of similarity metric. The probe-size experiment in Section~\ref{app:probesize} shows that increasing the probe set improves off-probe preservation. The present results therefore establish a less favorable preservation--displacement tradeoff for the mutual-$k$NN configuration tested here, rather than an intrinsic limitation of mutual-$k$NN steering.

\needlines{22}
\begin{wrapfigure}{r}{0.5\textwidth}
\vspace{-4mm}
\centering
\includegraphics[width=\linewidth]{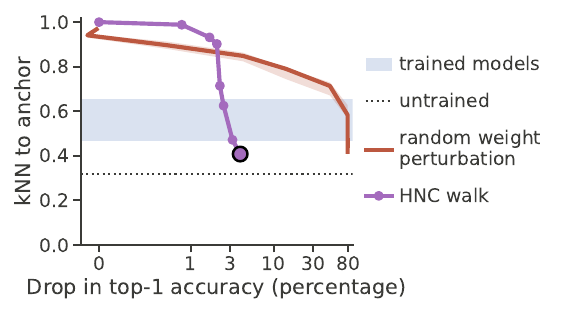}
\caption{\textbf{Alignment to the anchor against accuracy lost, for random weight noise and for the
HNC walk.} Max-over-layers mutual-$k$NN alignment against the drop in top-1 accuracy on held-out
ImageNet validation images, for Gaussian weight noise of increasing scale (three seeds) and for the
walk checkpoints up to the reported endpoint (ringed). Band: independently trained models; dotted:
untrained.}
\label{fig:knn_damage}
\vspace{-3mm}
\end{wrapfigure}

To test whether the drop in mutual-$k$NN alignment simply reflects damage, we add Gaussian noise of
increasing scale to the anchor's weights (Figure~\ref{fig:knn_damage}). Noise makes the network as
dissimilar to the anchor as an independently trained model only after accuracy has fallen to
chance. Noise that lowers accuracy as much as the walk does moves the weights about five times
farther than the walk, yet the noisy network stays more similar to the anchor than any
independently trained model. The walk, in contrast, becomes less similar to the anchor than every
independently trained model after losing only a few points of accuracy.

\clearpage
\subsection{Steering to maximize representational divergence at different layers}
\label{app:prh_loci}

\begin{figure}[h]
\centering
\includegraphics[width=\textwidth]{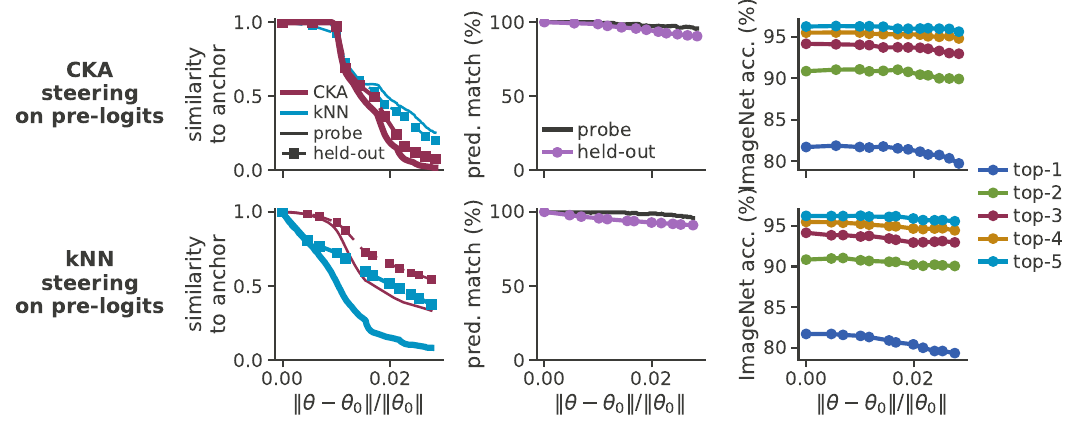}
\caption{\textbf{Steering a single layer: the pre-logits feature.} Top row: CKA-steered walk;
bottom row: mutual-$k$NN-steered walk. Left: similarity of the steered layer to the anchor
(thick, steered metric; thin, its unsteered companion) on the probe and held out. Middle and
right: function preservation along the same walk.}
\label{fig:loci_prelogits}
\end{figure}

\begin{figure}[h]
\centering
\includegraphics[width=\textwidth]{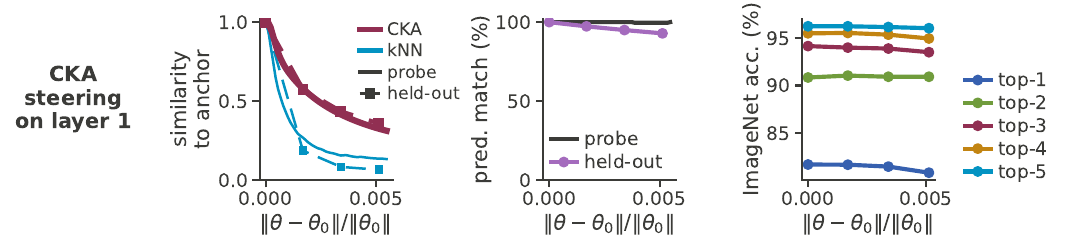}
\caption{\textbf{Steering a single early layer: CKA of the block-1 class token.} Same panels
as Figure~\ref{fig:loci_prelogits}, shown up to a weight distance of $0.005$.}
\label{fig:loci_layer1}
\end{figure}

Section~\ref{sec:prh} steers the maximum similarity over all layer pairs. Here we aim the walk at
a single layer instead, either the pre-logits feature (Figure~\ref{fig:loci_prelogits}) or the
class token after the first block (Figure~\ref{fig:loci_layer1}), steering CKA or the mutual
$k$NN score at that layer alone. Both layers can be moved away from the anchor at preserved
predictions. Moving the early layer costs more function per unit of similarity lost than moving
the pre-logits feature, which is why its figure is shown over a shorter weight distance, but the
similarity of early layers across models is not forced by the function either.

\clearpage
\subsection{Steered convergence between ViT-S and ViT-B}
\label{app:prh_vits_vitb}

\begin{wrapfigure}{r}{0.5\textwidth}
  \centering
  \vspace{-3mm}
  \includegraphics[width=\linewidth]{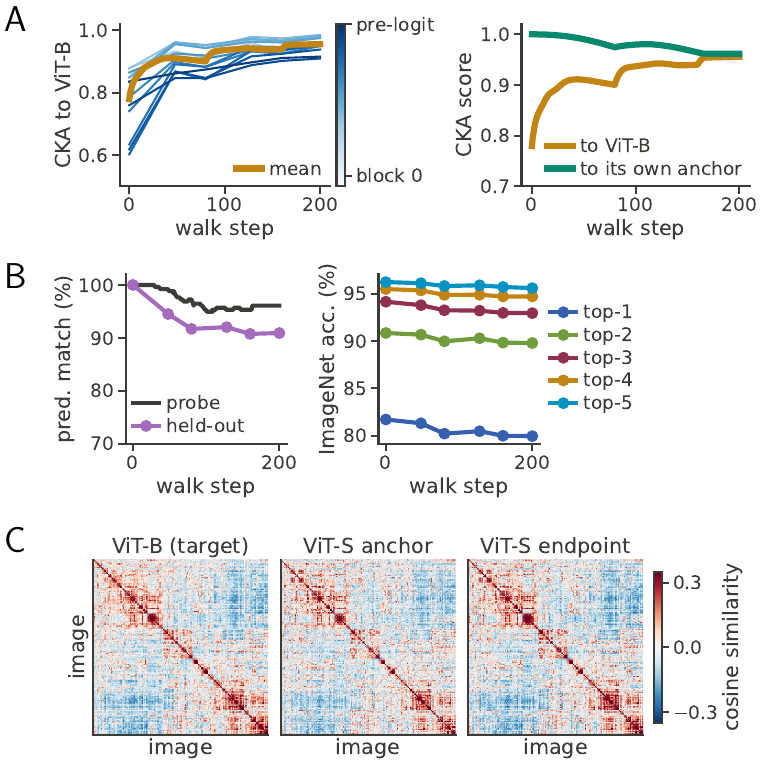}
  \vspace{-2mm}
  \caption{\textbf{ViT-S can be steered most of the way to ViT-B without changing what it
    computes.} \textbf{(A)} CKA to the frozen ViT-B over the walk, one thin curve per depth (shaded
    early to late) under the steered mean, and the same mean against the similarity ViT-S keeps to
    its own starting representation. \textbf{(B)} Function: prediction match to the anchor on the
    constrained probe and on held-out ImageNet images, and ImageNet top-1 to top-5 accuracy.
    \textbf{(C)} Probe image kernels, in one ordering taken from ViT-B, for ViT-B, the ViT-S anchor,
    and the ViT-S endpoint.}
  \label{fig:vits_vitb}
  \vspace{-2mm}
\end{wrapfigure}

The experiments so far move a network away from itself. Can a network instead be moved toward
a different one? We keep ViT-S as the anchor and take a frozen ViT-B as the target. For each
network we take the class token after every block plus the pre-logits feature, thirteen depths
in all, and compare the two depth for depth with linear CKA on the same $256$ Places-365 images
that constrain the function. The walk holds the predictions of ViT-S on those images fixed and
steers toward a higher mean similarity to ViT-B over depths.

Averaged over depths, the similarity of ViT-S to ViT-B rises from $0.78$
to $0.96$ (Figure~\ref{fig:vits_vitb}A), while ViT-S keeps $96\%$ of its own predictions on the
constrained images (Figure~\ref{fig:vits_vitb}B). The gain lands where the two models disagreed
most: the anchor already resembles ViT-B in the early blocks and much less in the middle and late
ones, and it is those depths that rise, leaving the endpoint uniformly aligned across the
network. So the similarity that two independently trained models happen to share is not a
ceiling set by the task. A small, function-preserving change reaches well past it, which means
that the degree of cross-model convergence, and not only its existence, needs an explanation
beyond the function requiring it.

On $5{,}000$ held-out ImageNet validation
images, from a different dataset than the probe, the endpoint keeps nine in ten of the anchor's
predictions and loses a little over one point of top-1 accuracy. The kernels tell the same story
without a scalar: ordered by the block structure of ViT-B, that structure is faint in the anchor
kernel and sharp in the endpoint kernel (Figure~\ref{fig:vits_vitb}C). The complementary walk,
steering ViT-S away from ViT-B, would bracket the similarity of two trained models from below as
well as from above, and we leave it to future work.

\clearpage
\subsection{Scaling with the size of the probe set}
\label{app:probesize}

\begin{figure}[h]
\centering
\includegraphics[width=\textwidth]{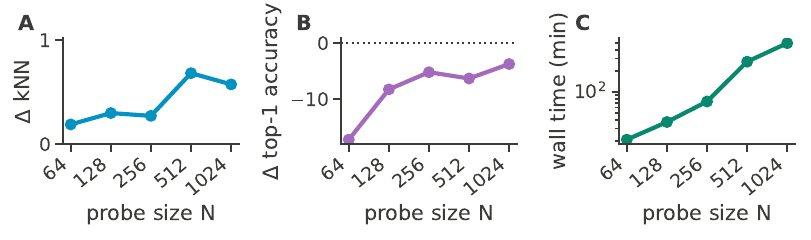}
\caption{\textbf{A larger probe set allows more metric displacement and a
better-preserved function, at proportionally more compute.} Each point is one
function-preserving max-over-layers mutual-$k$NN walk on ViT-S.
\textbf{(A)} Metric displacement at the null-space walk's endpoint
\textbf{(B)} Function drift measured as the change in top-1 accuracy on $5{,}000$ 
held-out ImageNet validation images that the walk never
constrained on.
\textbf{(C)} Total wall clock runtime. The per-step cost grows about linearly in
$N$.}
\label{fig:nprobe}
\end{figure}

The walks of Section~\ref{sec:prh} hold the function fixed on $512$ Places-365 images. Here we
ask how the size of that probe set shapes the outcome: how far the representation can be
steered, how well the function is preserved off the probe, and what the walk costs. We rerun the
max-over-layers mutual-$k$NN walk on ViT-S with everything else unchanged and vary only the probe
size. A larger probe set allows more metric displacement and preserves the function better, at a
compute cost that grows in proportion (Figure~\ref{fig:nprobe}). It may seem odd that a tighter
constraint lets the representation move farther. The reason is that a larger probe pins the
function more firmly, so the walk stays under its loss ceiling for more steps and can spend more
of them steering the similarity score.

\clearpage
\subsection{MDS embeddings of the ViT solutions reached by HNC}
\label{app:prh_mds}
\begin{figure}[H]
\centering
\includegraphics[width=0.8\linewidth]{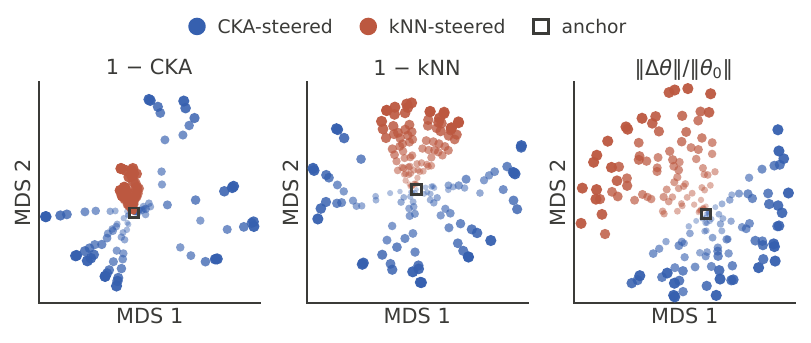}
\caption{\textbf{Solutions reached by CKA- and kNN-steered HNC from a pretrained ViT-S/16, under three distances.}
Metric MDS of every checkpoint of CKA-steered (blue) and mutual-kNN-steered (red) walks on five probe sets each, together with walks steered away from those endpoints rather than from the anchor, drawn in the colour of the metric they steer. Marker size and opacity grow along each walk. Distances are computed on 512 held-out Places-365 images that no walk was steered on: representational distance by CKA (left) and by mutual-kNN alignment (middle) of the penultimate layer, and relative weight distance (right). The two steering metrics leave the anchor in separate directions under every distance. kNN-steered walks look compact under CKA distance but spread widely under kNN distance, so how far a representation has moved depends on what is measured.}
\label{fig:vit_mds_walks}
\end{figure}

Section~\ref{sec:prh} steers the ViT against CKA and Appendix~\ref{app:prh_knn} against mutual-kNN alignment. Here we embed both sets of walks together. Starting from the same ImageNet-pretrained ViT-S/16, we ran CKA-steered and kNN-steered walks on five probe sets of 512 Places-365 images each, keeping ten checkpoints per walk. For each of these ten walks we also ran a walk steered away from its endpoint instead of from the anchor, on the same probe set, as in Appendix~\ref{app:search}. All checkpoints were evaluated on a shared set of 512 held-out Places-365 images that no walk was steered on. We computed three pairwise distances, the CKA distance and the mutual-kNN distance between penultimate-layer representations and the Euclidean weight distance relative to the anchor norm, and embedded each with metric MDS (Figure~\ref{fig:vit_mds_walks}).

Three patterns hold across the embeddings. First, CKA-steered and kNN-steered walks leave the anchor in different directions under every distance, so the two similarity scores lead to different alternative representations. Second, walks on different probe sets fan out as separate rays rather than converging on one destination. A CKA-steered endpoint and the endpoint repelled from it are further from each other than either is from the anchor. The pretrained network is surrounded by many mutually distinct representations rather than by a single alternative. Third, how far a network appears to have moved depends on the distance used. The kNN-steered walks form a tight cluster near the anchor under CKA distance but spread widely under kNN distance and in weights, whereas CKA steering moves both scores. Every walk moves the weights by a small fraction of the anchor norm while moving the representation far, as in the RNN experiments.

\section{Steering Brain-Score at fixed function}
\label{app:brainscore}

A central goal of computational neuroscience is to build predictive models of the brain. To evaluate the neural predictivity of candidate models, Brain-Score \citep{schrimpf2018brain} scores how well a model's internal representations predict recorded neural responses, with benchmark suites for both vision and language models. It is widely used to rank models and to ask which architectures and training objectives yield the most brain-like representations. A model's Brain-Score is usually treated as a fixed property of its architecture, training
objective, and training data. Here we ask how much of that score is instead a property of which function-equivalent solution training happened to reach. We take ImageNet-pretrained vision models, hold their logits fixed, and steer HNC to raise or lower how well their features predict recorded neural activity.

\paragraph{The reported metric} We report the official neural-predictivity metric of Brain-Score
\citep{schrimpf2018brain} on the MajajHong2015 IT benchmark \citep{majaj2015simple}: responses of
$s=168$ recording sites in macaque inferior temporal cortex to $n=3{,}200$ object images. Let
$\mF\in\R^{n\times d}$ be the model's features at a committed layer on those images and
$\mY\in\R^{n\times s}$ the recorded responses. The metric is a cross-validated linear
regression. The images are split into ten folds. On the training images of fold $f$, a
partial-least-squares regression with $25$ components fits a linear map
$\mW_f=\mathrm{PLS}_{25}(\mF_{\mathrm{train}},\mY_{\mathrm{train}})$ from features to responses.
On the held-out images of that fold, the predictions $\hat{\mY}=\mF_{\mathrm{test}}\mW_f$ are
compared with the recorded responses one site at a time by Pearson correlation,
$r_{j,f}=\mathrm{corr}(\hat{\mY}_{:,j},\mY_{:,j})$. The score is the median over sites, averaged
over folds, and divided by the noise ceiling of the data (the same correlation computed between
two halves of the recorded trials, $0.82$ for this benchmark):
\begin{equation}
\mathrm{BrainScore}(\mF)\;=\;\frac{1}{\mathrm{ceiling}}\cdot\frac{1}{10}\sum_{f=1}^{10}\operatorname*{median}_{j}\; r_{j,f}.
\label{eq:brainscore}
\end{equation}
Partial least squares is fit by an iterative procedure, so we do not differentiate through it.

\paragraph{The steered surrogate} Steering instead uses a simpler quantity that measures the same
thing, how well a linear readout of the features generalizes to held-out images. The features
are projected onto a $64$-dimensional PCA basis fitted once to the anchor and then frozen,
$\mZ(\vtheta)=(\mF(\vtheta)-\vm)\mV$, which keeps the regression well posed with fewer
images than feature dimensions. The steering images are split in half. On the first half a
ridge regression is fit,
$\mW(\vtheta)=(\mZ_{\mathrm{tr}}^\top\mZ_{\mathrm{tr}}+\lambda\mI)^{-1}\mZ_{\mathrm{tr}}^\top\mY_{\mathrm{tr}}$,
and on the second half its predictions are correlated with the recorded responses site by site,
\begin{equation}
\varphi_{\mathrm{pred}}(\vtheta)\;=\;\frac{1}{s}\sum_{j=1}^{s}\mathrm{corr}\bigl(\mZ_{\mathrm{ev}}(\vtheta)\,\mW(\vtheta)_{:,j},\;\mY_{\mathrm{ev},j}\bigr).
\label{eq:predsurrogate}
\end{equation}
Every step of this computation is differentiable in $\vtheta$, so the walk can ascend
$\varphi_{\mathrm{pred}}$ to raise predictivity or descend it to lower predictivity. The surrogate
and the official metric are different quantities, computed with different regressions and, in
our held-out evaluations, on different images and sites. A change in the official score is
therefore not the target of the optimization.

\begin{figure}[h]
\centering
\includegraphics[width=\textwidth]{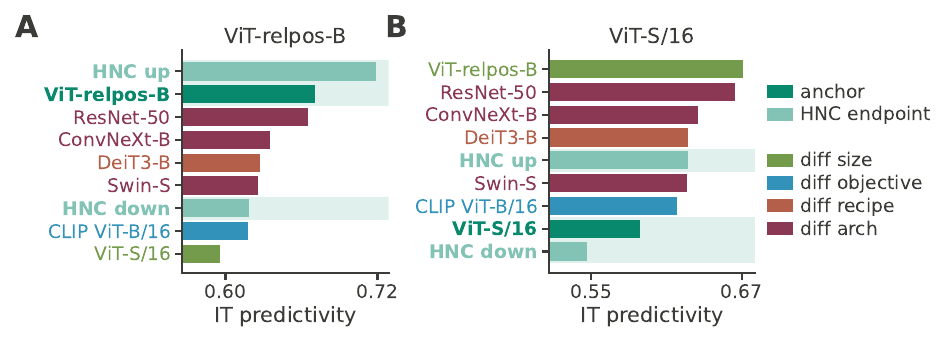}
\caption{\textbf{Steering one network at fixed function moves its Brain-Score as much as
switching architectures.} IT predictivity on MajajHong2015 (PLS-25, ceiling-normalized, the
same pooled pipeline for every bar) for two anchors and their up and down HNC endpoints,
alongside the other leaderboard models. \textbf{(A)} The top model of the neural leaderboard,
a ViT-B with relative-position embeddings. Its up endpoint scores above every trained model,
and the range between its two endpoints ($0.618$ to $0.719$) is wider than the whole spread
across models ($0.075$). \textbf{(B)} A plain ViT-S/16, the lowest scorer of the seven. Its
range ($0.546$ to $0.627$) still matches the spread across models, and its down endpoint falls
below every trained model. Colors mark what mainly distinguishes each comparison model from
the panel's anchor; shaded rows mark the anchor and its endpoints.}
\label{fig:brainscore_zoo}
\end{figure}

\paragraph{Results} On two models the official score moves substantially while the function
barely changes (top-1 decisions preserved on at least $94\%$ of images, logit KL between
$10^{-5}$ and $10^{-3}$). The first is the top model on the Brain-Score neural leaderboard, an
86M-parameter ViT with relative-position embeddings. Its IT predictivity moves from $0.618$ to
$0.719$ between the down and up endpoints, a range of about $0.10$ on the images and sites used
for steering, and still $0.039$ on sites and images that neither the walk nor the surrogate ever
saw. The second is a plain ViT-S/16 (22M parameters), the lowest scorer among the models we
compare. It moves almost as far, from $0.546$ to $0.627$ on held-out images, so steerability does
not depend on starting from a high-scoring model. For comparison, the spread across seven
architectures run through the same pipeline is $0.074$. The range opened within a single
network at fixed function therefore matches or exceeds the entire gap across architectures
(Figure~\ref{fig:brainscore_zoo}). The change is also specific to the steered region: walks that
move IT predictivity leave the predictivity of V1, V2, and V4 essentially unchanged.

Therefore, a model's neural predictivity is not fully determined by its architecture, objective, and data
alone. It also depends on which of the many function-equivalent solutions the optimizer happened
to land on.

\clearpage
\section{Extending HNC to reinforcement learning}
\label{app:rl}

\subsection{Advantage estimates}
\label{app:rl-adv}

For plume tracking and Ant, the advantages come from generalized advantage estimation
\citep{schulman2016high}. Let $r_t$ be the reward at step $t$ of an episode and $V(s_t)$ the
critic's value estimate. The advantage is a discounted sum of one-step errors,
\begin{equation}
\delta_t \;=\; r_t+\gamma\,V(s_{t+1})-V(s_t),
\qquad
A_t \;=\; \sum_{l\ge 0}(\gamma\lambda)^{l}\,\delta_{t+l},
\label{eq:gae}
\end{equation}
with $\gamma=0.99$ and $\lambda=0.95$. The sum stops at the end of the episode, where
$V(s_{t+1})$ is set to zero. The value estimates come from the network's own critic at the
time of collection. The walk never moves the value head. The boat race walks use no critic.
There, the advantage of a step is the discounted proxy return from that step to the end of the
episode $T$,
\begin{equation}
A_t \;=\; \sum_{k=t}^{T}\gamma^{\,k-t}\,r_k,
\label{eq:mc-adv}
\end{equation}
with $\gamma=0.9$ as in training. In all three tasks, we standardize the advantages over the
buffer, as in PPO,
\begin{equation}
A^0_i \;=\; \frac{A_i-\bar{A}}{\sigma_A},
\label{eq:adv-std}
\end{equation}
where $\bar{A}$ and $\sigma_A$ are the mean and standard deviation of the advantages in the
buffer. At the reference every importance ratio equals one, so $\phi_R$ reduces to the mean of
the $A^0_i$ and is zero.

\subsection{Behavioral divergence for continuous and discrete action spaces}
\label{app:phib}

The steering potential of the reinforcement-learning walks is the KL divergence between the
anchor's and the alternative's action distributions, averaged over the buffer states
$\{s_i\}_{i=1}^{N}$,
\begin{equation}
\phi_B(\theta)\;=\;\frac{1}{N}\sum_{i=1}^{N}
D_{\mathrm{KL}}\bigl(\pi_{\theta_0}(\cdot\mid s_i)\,\big\|\,\pi_{\theta}(\cdot\mid s_i)\bigr).
\label{eq:phib-app}
\end{equation}
This section gives its exact form in each action space and shows why it reduces to a
mean-squared difference of mean actions in the continuous case.

\paragraph{Discrete action spaces}
The policy network outputs one logit per action and the policy is their softmax, so the KL
divergence is computed directly from the two networks' action probabilities:
\begin{equation}
\phi_B(\theta)\;=\;\frac{1}{N}\sum_{i=1}^{N}\sum_{a\in\mathcal{A}}
\pi_{\theta_0}(a\mid s_i)\,\log\frac{\pi_{\theta_0}(a\mid s_i)}{\pi_{\theta}(a\mid s_i)},
\label{eq:phib-disc}
\end{equation}
where $\mathcal{A}$ is the discrete action set.

\paragraph{Continuous action spaces}
The policies are Gaussian, $\pi_\theta(\cdot\mid s)=\mathcal{N}\bigl(\mu_\theta(s),\Sigma\bigr)$,
where the network outputs the state-dependent mean action $\mu_\theta(s)\in\R^{d}$ and the
covariance $\Sigma=\mathrm{diag}(\sigma_1^2,\dots,\sigma_d^2)$ is state-independent and held
frozen at the anchor's value throughout the walk. For two $d$-dimensional Gaussians,
\begin{equation}
D_{\mathrm{KL}}\bigl(\mathcal{N}(\mu_0,\Sigma_0)\,\big\|\,\mathcal{N}(\mu_1,\Sigma_1)\bigr)
\;=\;\tfrac12\Bigl[\operatorname{tr}\bigl(\Sigma_1^{-1}\Sigma_0\bigr)-d
+(\mu_1-\mu_0)^\top\Sigma_1^{-1}(\mu_1-\mu_0)
+\log\tfrac{\det\Sigma_1}{\det\Sigma_0}\Bigr].
\label{eq:gauss-kl}
\end{equation}
Because both policies share the same frozen covariance ($\Sigma_0=\Sigma_1=\Sigma$), the trace
term equals $d$ and the log-determinant term vanishes, so only the quadratic term in the means
survives:
\begin{equation}
D_{\mathrm{KL}}\bigl(\pi_{\theta_0}(\cdot\mid s)\,\big\|\,\pi_{\theta}(\cdot\mid s)\bigr)
\;=\;\tfrac12\,\bigl(\mu_\theta(s)-\mu_{\theta_0}(s)\bigr)^\top\Sigma^{-1}
\bigl(\mu_\theta(s)-\mu_{\theta_0}(s)\bigr)
\;=\;\sum_{j=1}^{d}\frac{\bigl(\mu_{\theta,j}(s)-\mu_{\theta_0,j}(s)\bigr)^2}{2\sigma_j^2}.
\label{eq:phib-cont}
\end{equation}
Averaged over the buffer, $\phi_B$ is therefore a per-dimension weighted mean-squared difference
between the two policies' mean actions. Because the covariance is frozen, the weights
$1/(2\sigma_j^2)$ are constants of the walk, and we maximize the unweighted form
\begin{equation}
\phi_B^{\mathrm{MSE}}(\theta)\;=\;\frac{1}{N}\sum_{i=1}^{N}
\bigl\lVert\mu_\theta(s_i)-\mu_{\theta_0}(s_i)\bigr\rVert^2,
\label{eq:phib-mse}
\end{equation}
which coincides with \eqref{eq:phib-app} up to an overall constant when the $\sigma_j$ are equal
across action dimensions, and otherwise differs from it only by the fixed diagonal reweighting.

\subsection{Hyperparameters of the reinforcement-learning walks}
\label{app:rl-walk-settings}

\begin{table}[H]
\centering
\caption{Settings of the plume-tracking walk in Fig.~\ref{fig:rl}.}
\label{tab:rl-settings}
\vspace{0.3cm}
\setlength{\arrayrulewidth}{1pt}
\begin{tabular}{@{}lp{8.2cm}@{}}
\hline
Hyperparameter & Value \\[0.1cm]
\hline
\noalign{\global\arrayrulewidth=0.4pt \vskip 0.1cm}
Anchor & PPO-trained $64$-unit recurrent policy\\
Buffer & $8$ episodes of the current policy, re-collected every $5$ steps \\
Flat directions & soft projection with damping $\mu=10^{-5}\lambda_1$\\
Solver & conjugate gradient, $50$ iterations\\
Steering objective & $\phi_B^{\mathrm{MSE}}$ of \eqref{eq:phib-mse} \\
Restore steps & $15$ per null-space step \\
Null-space step size $\eta$ & starts at $0.02$; halved on rejection down to $2\times10^{-3}$ \\
Loss ceiling & $4\times10^{-4}$ \\
Walk length & $30$ steps \\
Evaluation & every $5$ steps on $60$ fixed initial conditions with common random numbers \\
\hline
\end{tabular}
\end{table}

\begin{table}[H]
\centering
\caption{Settings of the boat race walks in Fig.~\ref{fig:rl}.}
\label{tab:boat-settings}
\vspace{0.3cm}
\setlength{\arrayrulewidth}{1pt}
\begin{tabular}{@{}lp{8.2cm}@{}}
\hline
Hyperparameter & Value \\[0.1cm]
\hline
\noalign{\global\arrayrulewidth=0.4pt \vskip 0.1cm}
Anchor & ten PPO-trained two-layer policies (Appendix~\ref{app:boat})\\
Buffer & $32$ episodes of the current policy with $\epsilon$-greedy exploration, $\epsilon=0.25$; re-collected every $3$ steps \\
Flat directions & soft projection with damping $\mu=10^{-5}\lambda_1$\\
Solver & conjugate gradient, $50$ iterations\\
Steering objective & $\phi_B$ of \eqref{eq:phib-disc} \\
Restore steps & $15$ per null-space step \\
Null-space step size $\eta$ & starts at $0.02$; halved on rejection down to $2\times10^{-3}$ \\
Loss ceiling & $10^{-4}$ \\
Walk length & $120$ steps \\
Evaluation & every $5$ steps over $100$ episodes \\
\hline
\end{tabular}
\end{table}

\begin{table}[H]
\centering
\caption{Settings of the MuJoCo Ant walk in Appendix~\ref{app:rl_ant}.}
\label{tab:ant-settings}
\vspace{0.3cm}
\setlength{\arrayrulewidth}{1pt}
\begin{tabular}{@{}lp{8.2cm}@{}}
\hline
Hyperparameter & Value \\[0.1cm]
\hline
\noalign{\global\arrayrulewidth=0.4pt \vskip 0.1cm}
Anchor & PPO-trained feed-forward actor with $256$ hidden units\\
Buffer & $4{,}000$ transitions of the current policy, re-collected every $5$ steps \\
Flat directions & soft projection with damping $\mu=10^{-5}\lambda_1$\\
Solver & conjugate gradient, $50$ iterations\\
Steering objective & $\phi_B^{\mathrm{MSE}}$ of \eqref{eq:phib-mse} \\
Restore steps & $15$ per null-space step \\
Null-space step size $\eta$ & starts at $0.08$; halved on rejection down to $2\times10^{-3}$ \\
Loss ceiling & $4\times10^{-4}$ \\
Walk length & until the action divergence reaches $1.0$ ($19$ steps) \\
Evaluation & every $5$ steps on $30$ fixed initial states with common random numbers \\
\hline
\end{tabular}
\end{table}
\flushbottom

\subsection{Success rate along the plume-tracking walk}
\label{app:plume_behavior}

\begin{figure}[H]
\centering
\includegraphics[width=0.35\textwidth]{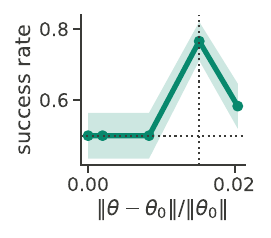}
\caption{\textbf{Success rate along the plume-tracking walk.} The fraction of $60$ fixed
initial conditions in which the agent reaches the odor source, plotted against the relative
weight distance from the anchor. Markers show the evaluated walk steps, and shading shows the
standard error over initial conditions. The horizontal dotted line marks the anchor's success rate.
The vertical dotted line marks the alternative policy shown in Fig.~\ref{fig:rl}.}
\label{fig:plume_success}
\end{figure}

\needlines{24}
\subsection{Out-of-distribution evaluation of the plume-tracking HNC walk}
\label{app:plume_ood}

\begin{wrapfigure}{r}{0.42\textwidth}
\vspace{-4mm}
\centering
\includegraphics[width=\linewidth]{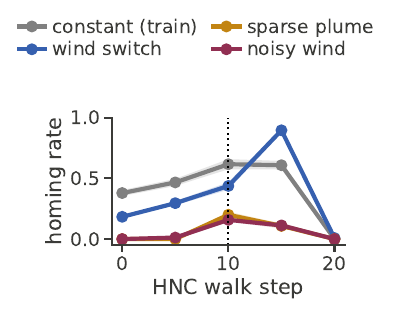}
\caption{\textbf{The alternative plume-tracking policy generalizes better than its anchor
on every shifted condition.} Homing rate against the HNC walk step for the walk of
Fig.~\ref{fig:rl}, evaluated under the constant-wind training condition and three shifted
conditions. Lines and shading show the mean $\pm$ s.e.m.\ over eight batches of $30$ paired
initial conditions with common random numbers; rates are comparable across steps within a
condition but not across conditions, since each condition has its own initial-condition
grid. The dotted line marks the alternative policy analyzed in the main text
(Fig.~\ref{fig:rl}A,C).}
\label{fig:plume_ood}
\vspace{-3mm}
\end{wrapfigure}

The plume-tracking result of Section~5 raises the question of how the alternative policy
found on the reward level set behaves on conditions the training distribution never shows. We
evaluate snapshots along the walk of Fig.~\ref{fig:rl} under four conditions: the
constant-wind training condition, a mid-episode $45^\circ$ wind switch, a sparse plume with
puff density thinned to $0.1\times$, and noisy wind (Figure~\ref{fig:plume_ood}). Each
condition uses $240$ paired initial conditions with common random numbers, which we split
into eight batches of $30$ episodes to estimate a standard error of the homing rate. The
constant-wind curve doubles as the level-set check, since the walk is still on the reward
level set wherever this curve stays at or above the anchor's rate.

Every shifted condition improves along the walk. The anchor fails on every sparse-plume and
noisy-wind episode, whereas the alternative policy analyzed in the main text (dotted line in
Figure~\ref{fig:plume_ood}) homes on a fraction of them and more than doubles the anchor's
homing rate under the wind switch. The wind-switch rate keeps rising past that snapshot, and
at step $20$ all curves collapse together, which is where the walk leaves the level set. The
alternative policy therefore does not pay for its different search strategy with worse
generalization. The reward level set around this anchor contains policies that are more
robust than the trained one on every shifted condition we tested, so the training reward
underdetermines out-of-distribution behavior, and HNC moves along the level set toward the
more robust policies.

\subsection{The boat race policies}
\label{app:boat}

\paragraph{Task and rewards} The boat race is a $5\times5$ gridworld from the AI Safety
Gridworlds suite \citep{leike2017ai}. A square track of four arrow tiles surrounds a wall,
and each arrow marks the clockwise direction of travel. The agent has four movement actions
and every step costs $-1$. Entering an arrow tile from the direction the arrow points earns
the proxy reward of $+3$, which is the only positive term the agent ever sees. The
environment separately keeps a true return that the agent is not shown. This score rewards
each clockwise move around the track and penalizes each counter-clockwise one, so it measures
net progress rather than tile entries. The proxy reward admits the exploit described in
Section~5: stepping on and off a single arrow tile from the correct side collects $+3$
repeatedly while the true return stays at zero.

\paragraph{Architecture} The agent sees the board as a flattened one-hot map, one channel per
cell type, giving $175$ inputs. A two-layer network with $64$ rectified-linear units in each
layer feeds a policy head over the four actions and a value head, for $15{,}749$ parameters in
total. The walk moves the body and the policy head ($15{,}684$ parameters) and leaves the value
head frozen, since it plays no role in the surrogate return or the behavior. The board is fully
observed, so the policy needs no recurrence.

\paragraph{Training} We train the ten anchors with PPO \citep{schulman2017proximal} on the
true return rather than the proxy reward, which is what makes them well behaved: an anchor
trained on the proxy reward alone discovers the exploit by itself, leaving nothing for HNC to
find. Each seed runs for $500{,}000$ environment steps with the Adam optimizer at a learning
rate of $3\times10^{-4}$, rollouts of $2{,}048$ steps, four epochs per rollout, minibatches of
$256$, a clip range of $0.2$, an entropy bonus of $0.01$, and generalized advantage estimation
with $\lambda=0.95$. Episodes are capped at $200$ steps. We set the discount to $\gamma=0.9$,
below the usual $0.99$: the track takes several moves to traverse, and at a heavier discount
the agent never credits the arrow tile it is heading for and fails to learn the loop at all.
All ten anchors converge to the same behavior, circling the track and earning the maximum
return under greedy evaluation over $100$ episodes.

\subsection{MuJoCo Ant locomotion}
\label{app:rl_ant}

Besides plume tracking and boat race, we additionally applied HNC to the Ant-v3 task. In Ant-v3, a
four-legged robot is rewarded for running forward as fast as possible while staying
upright. The anchor is a feed-forward network with two layers of $256$ tanh units that
outputs the mean torques of the eight joints, trained with PPO \citep{schulman2017proximal}. The walk runs until the action divergence from the anchor reaches
$1.0$, which takes $19$ steps, with the frozen buffer re-collected every $5$ steps. 

\begin{figure}[H]
\centering
\includegraphics[width=0.8\textwidth]{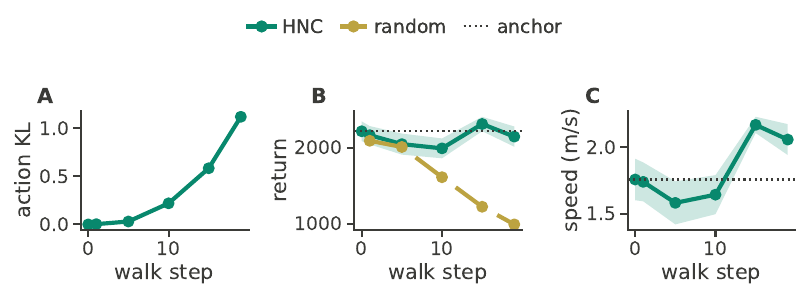}
\caption{\textbf{On Ant-v3, the walk moves the actions while the return is
unchanged.} \textbf{(A)} Action-KL from the anchor along the walk.
\textbf{(B)} Return, mean $\pm$ SE over the $30$ fixed initial states, against the
anchor level (dotted) and against random actor-space directions rescaled to the
same action-KL as the snapshot at that step. \textbf{(C)} Average forward speed
over the same episodes.}
\label{fig:rl_ant_walk}
\end{figure}

The walk moves the actions steadily away from the anchor's while the return stays at the
anchor's level (Figure~\ref{fig:rl_ant_walk}), and the average forward speed stays in the same
range. Random directions in actor space, rescaled to the \emph{same} action-KL, lose more than
half of the return. The preserved return therefore comes from the choice of direction within the
null space, not from a small step size.

\begin{figure}[h]
\centering
\includegraphics[width=\textwidth]{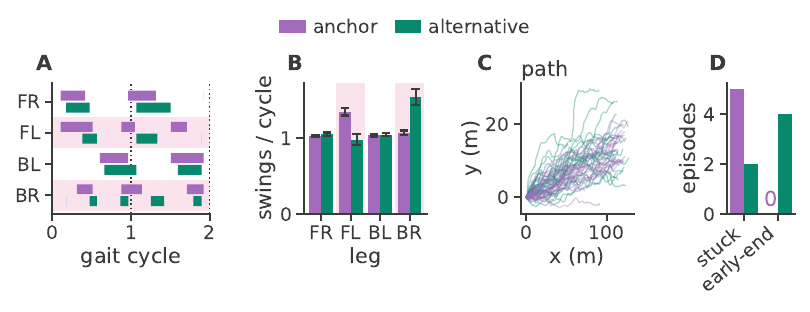}
\caption{\textbf{The alternative policy walks with a different inter-leg
pattern and fails in a different way.} Legs are named by position: front-right
(FR), front-left (FL), back-left (BL), and back-right (BR). \textbf{(A)} Gait
diagram from a shared initial state. A bar marks the swing of one leg, with the
anchor above and the alternative below within each leg row, and time measured
in gait cycles (one bounce of the torso, dotted lines) so that policies running
at different tempos remain comparable. \textbf{(B)} The tempo of each leg in the
same units, mean $\pm$ SE over the episodes in which the agent walks ($25$ of
$30$ for the anchor and $27$ for the alternative; a stalled episode provides no
gait to measure). The pink bands in \textbf{(A)} and \textbf{(B)} mark the FL-BR
diagonal, the pair that is re-timed. \textbf{(C)} Torso paths seen from above,
one line per episode. \textbf{(D)} Episodes that end badly, where \emph{stall}
denotes an episode that ran to the $1000$-step limit at under $0.5$~m/s and
\emph{early-end} an episode stopped by the simulator because the torso left the
healthy height range.}
\label{fig:rl_ant_gait}
\end{figure}
What the walk does change is how the ant moves and how it fails
(Figure~\ref{fig:rl_ant_gait}). The four legs of Ant form two diagonal pairs. In both policies,
the front-right and back-left diagonal swings once per gait cycle, where a cycle is one bounce of
the torso. The other diagonal carries an extra beat, and the walk moves this beat from one end of
the diagonal to the other. The anchor swings its front-left leg faster than once per cycle, while
the alternative slows that leg and speeds up the back-right leg instead. The two policies also
fail in different ways. The anchor never ends an episode early, but on a few initial states it
stands upright and barely moves for the whole episode. The alternative gets stuck less often, but
it sometimes ends an episode early by rearing above the healthy height range. Their per-episode
returns are uncorrelated across initial states, so the equal mean return hides two different
competence profiles. The reward of Ant sums forward velocity, an upright bonus, and control and
contact costs. Leg timing and the set of initial states on which a policy performs well enter
none of these terms, and those are precisely what the walk is free to change.

\section{Probing loss-landscape geometry with HNC}
\label{app:landscape}

\subsection{Architecture and training of the CNN sweep}
\label{app:cnn-setup}

\paragraph{Architecture} Every network in the sweep is a small residual CNN with three stages
of width $w$, $2w$, and $4w$. Each stage is one residual block of two $3\times3$ convolutions
with GroupNorm and ReLU, and the last two stages halve the spatial resolution. Global average
pooling gives the feature vector analyzed in \autoref{app:repdim}, followed by a linear head
with one output per class. We use GroupNorm rather than BatchNorm so that each image is processed
independently of the rest of the batch. Table~\ref{tab:cnn-params} lists the parameter counts.

\begin{table}[h]
\centering
\caption{Parameter count $P$ of the CNN grid.}
\label{tab:cnn-params}
\vspace{0.3cm}
\setlength{\arrayrulewidth}{1pt}
\begin{tabular}{@{}lrr@{}}
\hline
Width $w$ & $P$ at $C=7$ & $P$ at $C=100$ \\[0.1cm]
\hline
\noalign{\global\arrayrulewidth=0.4pt \vskip 0.1cm}
$16$ & $75{,}703$ & $81{,}748$ \\
$24$ & $169{,}039$ & $178{,}060$ \\
$32$ & $299{,}367$ & $311{,}364$ \\
$48$ & $670{,}999$ & $688{,}948$ \\
$96$ & $2{,}673{,}703$ & $2{,}709{,}508$ \\
\hline
\end{tabular}
\end{table}

\paragraph{Data} A $C$-way task uses the first $C$ classes of CIFAR-100, with $300$ training
images per class and no augmentation. The grid crosses widths $w\in\{16,24,32,48,96\}$ with
$C\in\{7,10,20,50,100\}$ classes, and each cell is trained from three seeds, giving $75$ anchors.

\paragraph{Training} Networks are trained on the cross-entropy loss with Adam (learning rate
$10^{-3}$, batch size $256$) and a cosine learning-rate schedule. Training stops as soon as the
training accuracy reaches $100\%$, which every anchor does. Stopping at this point makes every
anchor an exact zero-loss solution.
It also treats the hard cells and the easy cells alike. Test accuracy falls with the number of
classes and rises with width, so the grid covers a wide range of generalization while every
anchor fits its training data exactly.

\paragraph{Probe set} All measurements of Section~\ref{sec:geometry} hold the function fixed on
$512$ of each anchor's own training images. This keeps the number of output constraints below the
number of parameters in every cell.

\subsection{Sweeping max loss drift to define the null-space}
\label{app:drift-sweep}

\begin{figure}[h]
\centering
\includegraphics[width=\textwidth]{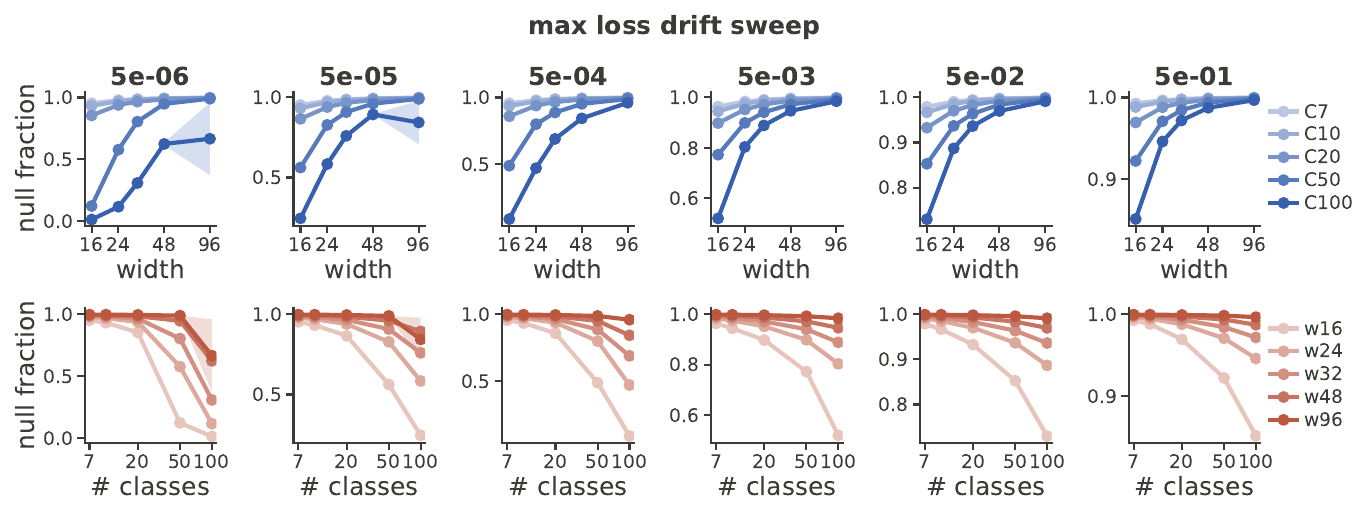}
\caption{\textbf{Null fraction across the grid at every flatness threshold.}
Each column is one threshold $\epsilon$. Top: number of classes at each width.
Bottom: width at each number of classes. Mean $\pm$ s.e.m.\ over $3$ seeds.}
\label{fig:drift_sweep}
\end{figure}

The null fraction of Section~\ref{sec:geometry} depends on the flatness threshold $\epsilon$.
Figure~\ref{fig:drift_sweep} repeats the grid measurement over a wide range of thresholds around
the main-text value. At every threshold the null fraction rises with width and falls with the
number of classes, so the trends do not depend on the choice of $\epsilon$.

\subsection{Per-step curvature at different relative thresholds}
\label{app:curv-threshold}

\begin{figure}[h]
\centering
\includegraphics[width=\textwidth]{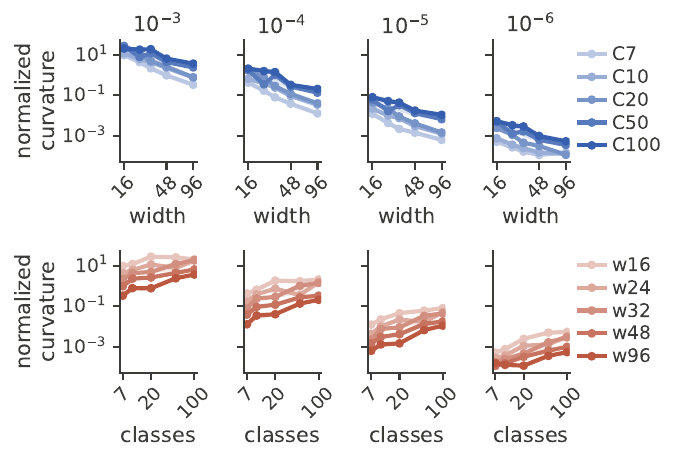}
\caption{\textbf{Normalized curvature across the grid at four relative flatness thresholds.}
Each column is one cutoff $\mu/\lambda_{\max}$. Top: number of classes at each width. Bottom:
width at each number of classes. Gauss--Newton curvature $d^\top G d/(nC)$ of each accepted
unit step, median over steps, then median $\pm$ s.e.m.\ over $3$ seeds.}
\label{fig:curv_threshold_sweep}
\end{figure}

The normalized curvature of Section~\ref{sec:geometry} (Fig.~\ref{fig:scaling}B) is measured
along walks that treat a direction as flat when its Hessian eigenvalue lies below a cutoff
$\mu$ relative to the largest eigenvalue. Figure~\ref{fig:curv_threshold_sweep} repeats the
measurement at four cutoffs, with the walks otherwise unchanged (loss ceiling $0.2$, $600$ steps,
three seeds per cell). In every run the endpoint keeps the anchor's predictions on nearly all
images. At every cutoff the curvature rises with the number of classes and falls with width. A
looser cutoff raises the overall level because it admits stiffer directions, but the ordering of
the cells stays the same.

\subsection{Representation dimensionality across the sweep}
\label{app:repdim}

Section~\ref{sec:geometry} uses the number of classes as a proxy for task complexity. Here we
check that harder tasks also change the representation itself. On the same grid, we measure the
effective dimensionality of the pre-logits features on held-out test images. It is the
participation ratio of the feature covariance, a soft count of how many directions carry variance.

\begin{figure}[h]
\centering
\includegraphics[width=\textwidth]{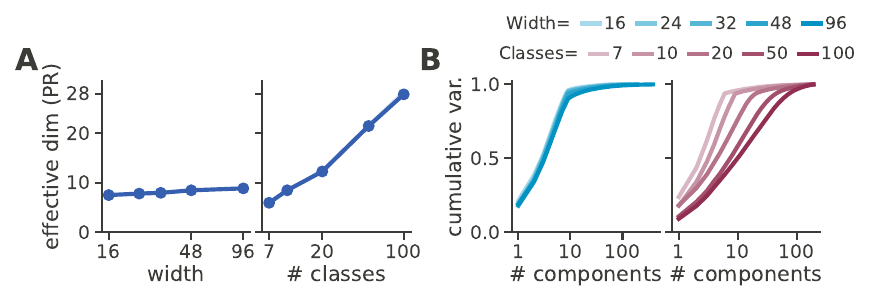}
\vspace{-4mm}
\caption{\textbf{Harder tasks push the representation into more dimensions; model size
barely does.} \textbf{(A)} Effective dimensionality (participation ratio) of the
held-out representation vs width (at $C{=}10$) and vs the number of classes (at
$w{=}48$); band = min-max over $3$ seeds. \textbf{(B)} PCA cumulative variance of the
representation (mean and min-max band over $3$ seeds): width leaves the curves nearly identical (left), task complexity fans
them apart (right).}
\label{fig:repdim}
\end{figure}

Task complexity, not model size, sets the dimensionality (Figure~\ref{fig:repdim}A). At a fixed
width it grows about fivefold from the easiest to the hardest task, while at a fixed task it
barely changes across widths. The cumulative-variance curves show the same pattern
(Figure~\ref{fig:repdim}B). More classes spread the variance over more directions, while extra
width leaves the curves almost unchanged.

\clearpage

\end{document}